\documentclass[preprint,12pt]{clear2025} 

\usepackage{amsfonts}
\usepackage[normalem]{ulem}
\usepackage{array}
\PassOptionsToPackage{hyphens}{url}\usepackage{hyperref}
\usepackage[shortlabels]{enumitem}
\setlist[enumerate]{noitemsep, topsep=0pt}
\usepackage{centernot}
\usepackage{comment}
\usepackage{booktabs}
\usepackage{multirow}
\usepackage{subcaption}

\usepackage{xcolor} 
\definecolor{matplotlib_blue}{rgb}{0.12156863, 0.46666667, 0.70588235}
\definecolor{yellow}{HTML}{FEC000}
\definecolor{green}{HTML}{70AD47}
\definecolor{new}{HTML}{000000}

\usepackage{tikz}
\usetikzlibrary{arrows,shapes,plotmarks,positioning, arrows.meta, decorations.markings}
\tikzset{>=stealth'} 
\tikzset{every node/.style={inner sep=2pt}}
\tikzset{every path/.style={line width=.7pt}}

\newcommand{\Exp}{\mathbb{E}} 
\newcommand{\R}{{\mathbb R}}
\newcommand{\N}{{\mathbb N}}
\newcommand{\cX}{{\cal X}}
\newcommand{\cY}{{\cal Y}}
\newcommand{\cZ}{{\cal Z}}
\newcommand{\bZ}{{\boldsymbol{Z}}}
\newcommand{\bz}{{\boldsymbol{z}}}
\newcommand{\bW}{{\boldsymbol{W}}}
\newcommand{\cF}{{\cal F}}
\newcommand{\bN}{\boldsymbol{N}}

\newcommand{\cov}{{\rm Cov}}
\newcommand{\ch}{\text{ch}}
\newcommand{\pa}{\text{pa}}
\newcommand{\sib}{\text{sib}}

\newcommand\independent{\protect\mathpalette{\protect\independenT}{\perp}}
\def\independenT#1#2{\mathrel{\rlap{$#1#2$}\mkern2mu{#1#2}}}

\newcommand{\new}[1]{{\color{new}{#1}}}

\graphicspath{{./Graphics/}}

\title[Cross-validating causal discovery]{Cross-validating causal discovery via Leave-One-Variable-Out}
\usepackage{times}

\clearauthor{%
 \Name{Daniela Schkoda} \Email{daniela.schkoda@tum.de}\\
 \addr Technical University of Munich, Germany
 \AND
 \Name{Philipp M. Faller} \Email{philipp.faller@partner.kit.edu}\\
 \addr Karlsruhe Institute of Technology, Germany\\Amazon Research T\"ubingen, Germany\\
 \AND
 \Name{Patrick Bl\"obaum} \Email{bloebp@amazon.com}\\
 \addr Amazon Research T\"ubingen, Germany
 \AND
 \Name{Dominik Janzing} \Email{janzind@amazon.com}\\
 \addr Amazon Research T\"ubingen, Germany
}

\begin{document}

\maketitle

\begin{abstract}
We propose a new approach to falsify causal discovery algorithms without ground truth, which is based on testing the causal model on a variable pair excluded during learning the causal model. Specifically, given data on $X, Y, \boldsymbol{Z}=X, Y, Z_1,\dots,Z_k$, we apply the causal discovery algorithm separately to the 'leave-one-out' data sets $X, \boldsymbol{Z}$ and $Y, \boldsymbol{Z}$. We demonstrate that the two resulting causal models, in the form DAGs, ADMGs, CPDAGs or PAGs, often entail conclusions on the dependencies between $X$ and $Y$ and allow to estimate $\mathbb{E}(Y\mid X=x)$ without any joint observations of $X$ and $Y$, given only the leave-one-out datasets. This estimation is called 
 "Leave-One-Variable-Out (LOVO)" prediction. Its error can be  estimated since the joint distribution $P(X, Y)$ is available, and $X$ and $Y$ have only been omitted for the purpose of falsification.
 
We present two variants of LOVO prediction: One graphical method, which is applicable to general causal discovery algorithms, and one version tailored towards algorithms relying on specific a priori assumptions, such as linear additive noise models. 
Simulations indicate that the LOVO prediction error is indeed correlated with the accuracy of the causal outputs, affirming the method's effectiveness.
\end{abstract}

\section{Introduction}
Causal discovery \citep{Spirtes1993}, the inference of (typically acyclic) causal graphs from observational data, has attained substantial research interest since the development of the PC algorithm \citep{Spirtes1993,Glymour2019}, which leverages the causal Markov condition and faithfulness assumption.
Research gained further momentum after it was observed that additional assumptions render identification solvable also within Markov equivalence classes, see, e.g., \cite{Kano2003, Shimizu2006, SunLauderdale, Hoyer,Zhang_UAI, UAI_identifiability,Kocaoglu2017, Gnecco2019, Roland2022, Montagna2022}. 
These approaches are meanwhile complemented by supervised learning methods: 
 \cite{Lopez2015} treats cause-effect inference as a binary classification problem, 
\cite{Nauta2019,lachapelle2020,zheng20a,ke2022learning} use techniques from deep learning to learn multivariate causal graphs using architectures tailored for learning properties of probability distributions. 
Further, it has been shown that data from changing environments  helps in identification of causal models \citep{Tian2001CausalDF,Peters15,Zhang2017CausalDF,Mooij2020,Rothenhaeuser2021}.  %
However, even after decades of creative contributions, it is fair to say that causal discovery did not experience any widely celebrated breakthroughs in practical applications despite interesting success stories, e.g., \cite{Xinpeng2020,Lagemann2023}. 
One reason, if not the main one, is that researchers working on practical applications find it hard to judge which method works best or if any works reasonably well for their use case. 
Extensive evaluations are mainly performed on simulated data, a practice about which serious doubts are in order \citep{Reisach2021}. 
Despite the existence of instructive examples for interventional data \citep{Lagemann2023}, such as knockout
experiments in genetics \citep{Hamilton1989}, it is a serious limitation to entirely rely on interventions. 
This is not only because interventional experiments are expensive but also because interventions can not necessarily be attributed to single nodes \citep{zhang2023identifiability}. Thus, some datasets may require a separate discussion about what 
node is intervened on, which motivated research on learning causal graphs from unknown intervention targets \citep{Jaber2020}. 
Further, "ground truth" reported in the literature \citep{sachs2005causal} has later been challenged elsewhere 
\cite[Section 5.8]{Mooij2020}.  
 In other words, despite all the interesting experimental data sets, automatic retrieval of a vast number of interventional data for trustworthy benchmarking seems currently out of reach.
In practical applications, researchers often solve causal inference tasks like treatment effect estimation in directed acyclic graphs (DAGs) with comparable low complexity and raise doubts about whether more complex DAGs can be trusted \citep{Imbens2020}.

Motivated by the lack of benchmarking data, \cite{Faller2024} suggests a "self-compatibility" check, which applies causal discovery algorithms to subsets of variables and quantifies to what extent the algorithm's outputs on subsets contradict the output on the entire 
 set of variables. 
The paper discusses different measures of disagreement. However,
\textit{some} disagreement is not unlikely,  and the paper provides no guidance on which level of disagreement is large enough to reject the algorithm's outcome. 
\new{Given data on variables $\bW = (X,Y,\bZ)=(X, Y, Z_1, \dots, Z_k)$, similar to \cite{Faller2024}, our method begins with applying the causal discovery algorithm in question to subsets of variables, namely $X,\bZ$ and $Y,\bZ$. The novelty lies in introducing a new measurement: We demonstrate that the causal models on the two subsets often allow us to infer $\Exp[Y|X=x]$ only based on the two datasets  $X,\bZ$ and $Y,\bZ$. Comparing this to an estimate obtained from the joint distribution $P(X, Y)$, we obtain the LOVO prediction error. Unlike the discrete metrics in \cite{Faller2024}), this prediction error is continuous and, therefore, constitutes an optimizable loss function. Moreover, by focusing on a bivariate relation, we ensure a statistically well-behaved target. Most importantly, we establish a rigorous criterion to decide when to reject the algorithm's output by developing a baseline to which we compare the prediction error.} 

While our task can be seen as a missing data problem (for which causal structure is known to enable better imputation, see e.g. \cite{Mohan2021}),
 here, missingness  comes from dropping on purpose 
 for testing causal hypotheses.  We call our  scenario "Leave-One-Variable-Out (LOVO)" cross-validation in 
analogy to leave-one-out (LOO) cross-validation in statistical learning \citep{Stone1974}. While LOO cross-validation
evaluates a model prediction for a {\it datapoint} that has not been used for learning, we test the prediction of the causal model 
at a {\it variable pair} $(X,Y)$ \new{whose relation} has not been used for learning. We will discuss conditions under which 
causal models render this task solvable. In other words, while statistical learning is based on an inductive bias that
allows to interpolate a function at a point that has not been seen before, causal learning may "interpolate"
dependencies between variable pairs that have not been seen together. 
This way, we further elaborate on the view of previous works  \citep{Tsamardinos,Dhir2019,Gresele2022,Guo2024} 
that causal models help for generalization across variables,
called "out-of-variable generalization" in \cite{Guo2024}, with the difference that we predict a statistical relation that is actually known and only temporarily dropped for the purpose
of testing. One of the early works that observed that causal models enable predicting relations between variables
not observed together is \cite{Tsamardinos}, where  Maximal Ancestral Graphs (MAGs)
on overlapping subsets of variables are used to infer dependencies between variables, each of which only occurs in one of the subsets. 

\paragraph{Structure of the paper:} After Section \ref{sec:building_blocks} formally defines LOVO prediction and lays out the general structure of LOVO cross-validation, in Section \ref{sec:causally_informed_LOVO}, we examine under which conditions causal models enable LOVO prediction and propose a practical estimation method. Section \ref{sec:baseline} defines the "non-causal baseline," i.e., a simple LOVO prediction rule to be used when nothing is known about the causal structure.  
Section \ref{sec:exp} reports experiments where we apply LOVO prediction to the causal discovery algorithms DirectLiNGAM \citep{DirectLiNGAM} and Repetitive Causal Discovery \citep{maeda2020rcd}. All proofs are given in the appendix.

\paragraph{Notation and technical assumptions:} To simplify mathematical discussions and notation, we will always assume that the joint distribution
$P(X,Y,\bZ)$ has a density (the probability mass function for the case of discrete variables) with respect to a product measure.  Further, except for results that explicitly refer to continuous variables, we use discrete sum over probabilities 
without being explicit about replacing them with integrals over densities otherwise. For standard concepts of
causal discovery like the causal Markov condition, 
d-separation, and Markov equivalence, we refer to the literature, e.g., \cite{Spirtes1993,Pearl:00}.

\section{Building blocks for LOVO cross-validation}\label{sec:building_blocks}
We interpret LOVO prediction as the task of inferring $P(Y|X)$, or the reduced problem of inferring the regression function $f(x) = \Exp[Y|X=x]$, or the correlation $\rho_{XY}$, from $P(X, \bZ)$ and $P(Y, \bZ)$. \new{This prediction is used to assess the reliability of a causal discovery algorithm as follows: Given data on the whole set of variables $\bW$, we select one pair of variables $(X, Y)$ and create the two leave-one-out datasets $(X, \bZ)$ and $(Y, \bZ)$. Next, we run the causal discovery method separately on both sets. As we show later, the outputs $G_X, G_Y$ often provide information about the relation between $X$ and $Y$, aiding in LOVO prediction. Correspondingly, we call this specific way of LOVO prediction based on $G_X, G_Y$ \textit{causally informed LOVO prediction}. Its prediction error is then estimated by comparing it to an estimate obtained from the joint distribution $P(Y, X)$.} To obtain the overall LOVO cross-validation error, we repeat the procedure for all choices of pairs $(X, Y)$ from $\bW$ and take the average. Finally, to decide whether the error is still acceptable or so large that we should reject the outcome of the causal discovery algorithm, we compare the cross-validation error to the error of a baseline LOVO predictor, which estimates $P(Y|X)$ from $P(X, Z)$ and $P(Y, Z)$ without using any causal information. 

Since we perform causal inference on subsets of the whole dataset, we need a class of causal graphs that enables marginalizations. Following \cite{Faller2024}, we use acyclic directed mixed graphs (ADMGs) \citep{richardson2003markov} 
which contain the usual causal edges $\rightarrow$ as well as edges $\leftrightarrow$ (indicating a confounding path that can not be blocked
by any observed variable). There exist slightly different definitions of ADMGs across the literature, varying in whether to allow co-occurrence of both types of edges as  confounded causal links $A \stackrel{\leftrightarrow }{\rightarrow} B$. We allow these confounded causal links\footnote{Nonetheless, our approach can be adapted to the other definition, see Supplement \ref{subsec:LOVO_ADMGs_without_confounded_links}.} and say that a child $B$ of $A$ is a confounded child, if $A \stackrel{\leftrightarrow }{\rightarrow} B$, and an unconfounded child if $A \to B$ is the sole edge between the two nodes, similarly for parents. If $A \leftrightarrow B$, we say that $A$ and $B$ are siblings. We denote the children, parents, and siblings of a node $A$ by $\ch(A), \pa(A),$ and $\sib(A)$, respectively. Moreover, we use the symbol $-$ to denote any type of directed or bidirected edge, and $A\centernot{-}B$ to indicate that $A$ and $B$ are not connected by an  edge.
For details of marginalization in ADMGs, we refer
to \cite{richardson2003markov}, but it is rather intuitive: directed paths $A \rightarrow * \rightarrow B$ turn into edges $A \rightarrow B$ when marginalizing over the mediators, confounding paths $A\leftarrow * \rightarrow B$ or $A\leftarrow * \leftrightarrow B$ into bidirected edges $A \leftrightarrow B$ when marginalizing over the common cause. By $G$, we refer to the DAG or ADMG for the entire set of nodes $\bW$\footnote{This implicitly assumes that the joint data is Markov to some ADMG.}, and $G_X, G_Y$ are its marginalizations when leaving out $Y, X$.

\section{Constructing LOVO predictors via causal discovery \label{sec:causally_informed_LOVO}}
\subsection{\new{Connection of causality and LOVO prediction}}\label{subsec:3_variables}
We first consider a small toy scenario on three variables $(X, Y, Z)$, whose joint graph is assumed to be a DAG, to provide an intuition under which conditions causally informed LOVO prediction is feasible. Alongside, we illustrate that it makes sense to proceed by the following steps:\\
1. Infer the joint DAG (or later ADMG) $G$ from the two marginal graphs $G_X, G_Y.$ \\
2. Use the joint graph $G$ to reconstruct $P(X,Y)$ from $P(X,Z)$ and $P(Y,Z)$.\\
Suppose a causal discovery algorithm yields the outputs 
$$G_X = (X \to Z), \,\, \ G_Y = (Z \to Y)$$ when applied to the leave-one-out subsets $(X, Z), (Y, Z)$. First, we want to explore if these outputs allow us to draw conclusions about the entire graph.
Employing the marginalization rules mentioned earlier, we find that the edge $X \to Z \in G_X$ can arise if and only if $X \to Z$, $X \to Y \to Z$, or both structures are contained in $G$. Since the second option contradicts $G_Y = (Z \to Y)$, we conclude $X \to Z \in G$. Similarly, we obtain that $Z \to Y \in G$. Lastly, we need to check if $G$ could contain other edges: By acyclicity, the only potential additional edge is $X \to Y$. However, if $X \to Y \in G$, then $G_Y$ would be $Z \stackrel{\rightarrow}{\leftrightarrow} Y$. Knowing $G$, we can conclude $X \independent Y \mid Z$, and therefore,
\begin{equation*}
    P(X,Z,Y) = P(X,Z)P(Y|Z),
\end{equation*}
which determines $\Exp[Y|X=x]$. Thus, if the true underlying graph is $X \to Z \to Y$, we are able to  construct a LOVO predictor. 
While we use a conditional independence statement entailed by $G$ here, we want to stress that not every LOVO predictor 
is based on conditional independencies. 
Theorem \ref{thm:LOVO_via_lingam} or Table \ref{tab:dags} in the supplement 
contain cases where 
it can be inferred despite 
$X\not\independent Y\mid\bZ$, using other statistical properties entailed by the joint causal model. 
If, for instance, the joint DAG is $X\to Y\to Z$, linear models enable the identification of $\Exp[Y|X=x]$. 
Moreover, in the preceding step to infer $G$ from $G_X, G_Y$, we heavily make use of the arrows' directions. If, in the example above, we had only gained knowledge on the skeletons of $G_X, G_Y$, the joint graph could be any graph in which $Z$ is not an isolated node.  This is not special to this example; instead 
inferring $\Exp[Y|X=x]$  can not follow from the conditional independencies observed in $P(X,\bZ)$ and $P(Y,\bZ)$ via any {\it mathematical laws} 
(except for degenerate cases, e.g, when $\bZ$ uniquely determines $X,Y$), as formalized in the following lemma:
\begin{lemma}[No probabilistic law enables LOVO prediction]\label{lem:needbias}
Let $X,Y$ be real-valued variables whose conditional distributions $P(\new{X}|\bZ=\bz)$, $P(Y|\bZ=\bz)$
have densities $p(x|\bz)$ and $p(y|\bz)$ with respect to the \new{Lebesgue} measure. Let $\bZ=\{Z_1,\dots,Z_k\}$ be variables
with arbitrary range. Then $P(X,\bZ)$ and $P(Y,\bZ)$ can never uniquely determine $P(X,Y)$. 
In particular, even the sign of their correlation is ambiguous. 
\end{lemma}
The proof is quite explicit about the remaining ambiguity: when generating $P(X,\bZ)$ and $P(Y,\bZ)$ via structural equation models with noise variables $N_\bZ^X$ and $N_\bZ^Y$, respectively, the dependencies between $N_\bZ^X$ and $N_\bZ^Y$ only influence the joint distribution, but not the marginals.
Note that $\bZ$ can consist of multiple variables here; thus, the lemma is a general result.

However, also causal models do not always enable LOVO prediction. For example, $G=(Z \to X \to Y)$ can not be uniquely reconstructed from its marginal graphs $G_X = (Z \to X)$ and $G_Y = (Z \to Y)$ since, e.g., the graphs $X \leftarrow Z \to Y$ and  $Z \to Y \to X$ have the same marginal graphs. While the graph $G = (X \to Y \leftarrow Z)$ is uniquely determined from its marginal graphs, $G_X = (X \centernot{-} Z)$ and $G_Y = (Z \to Y)$, 
here, the second step fails. Because $X \independent Z$, we can not combine $P(X, Z)$ and $P(Y, Z)$ to extract information on the connection strength from $X$ to $Y$. Appendix \ref{subsec:12dags} presents an overview of the realizability of LOVO for all possible graphs consisting of three nodes and two edges, indicating that in many cases, one of the two steps fails. However, in practice, it suffices if we can construct a LOVO predictor for a few pairs of nodes in the graph; then, we simply compute the cross-validation error as an average over those pairs for which we can construct the LOVO predictor. Furthermore, small graphs are particularly challenging for LOVO prediction because the overlap between $(X, Z)$ and $(Y, Z)$ is small. The next section and the simulations reveal that for larger graphs, we typically discover at least one (and often several) pairs that can be handled.

\subsection{\new{LOVO prediction via parent adjustment}}\label{sec:lovofromdags}
This section discusses the general case in which $\bZ$ may contain multiple variables, and the joint graph may be a DAG or an ADMG. 
Analogously to our first example, whenever there is a set $\bZ_S$ that renders $X$ and $Y$ conditionally independent, we can define a LOVO predictor via
the equation 
\[
P(y|x) = \sum_{\bz_S} P(y|\bz_S) P(\bz_S|x).
\]
Since graphs with more than three nodes often contain at least some pairs of conditionally independent nodes, we mainly rely on this LOVO predictor. Specifically, in a DAG, $X$ and $Y$ are conditionally independent 
with respect to an appropriate conditioning set
if and only if they are not connected by an edge. In this case, the union of parents of $X$ and $Y$ is a d-separating set. In contrast, in ADMGs, the absence of a direct link does not guarantee the existence of an m-separating set (e.g., for $ X \rightarrow Z \stackrel{\leftrightarrow}{\rightarrow} Y$). Here, the union of parents is m-separating if there is no link and all the parents are unconfounded.
    \begin{figure}
    \centering
    $G$ = \begin{tikzpicture}[baseline={(0,.6)}]
    \node (3) at (-1.4, 0) {$W_3$};
    \node (5) at (1.4, 0) {$W_5$};
    \node (2) at  (0,1.2) {$W_2$};
    \node (4) at (0,0){$W_4$};
    \node (1) at (-1.4,1.2) {$W_1$};
    \draw[->, matplotlib_blue,transform canvas={yshift=-2pt}] (3) to (4); 
    \draw[<->, orange,transform canvas={yshift=2pt}] (3) to (4);
    \draw[->] (1) -- (4);
    \draw[->] (1) -- (3);
    \draw[->] (2) -- (3);
    \draw[->] (4) -- (5);
    \draw[->] (2) -- (4);
\end{tikzpicture} \hspace{0.3cm}
 $G_{W_3}$ =  \begin{tikzpicture}[baseline={(0,.6)}]
    \node (3) at (-1.4, 0) {$W_3$};
    \node (5) at (1.4, 0) {$W_5$};
    \node (2) at  (0,1.2) {$W_2$};
    \node (1) at (-1.4,1.2) {$W_1$};
    \draw[->] (1) -- (5);
    \draw[->] (1) -- (3);
    \draw[->] (2) -- (3);
    \draw[->] (2) -- (5);
    \draw[<->,orange,transform canvas={yshift=2pt}] (3) -- (5);
    \draw[->,matplotlib_blue,transform canvas={yshift=-2pt}] (3) -- (5);
\end{tikzpicture}\hspace{0.3cm}
    $G_{W_4}$ = \begin{tikzpicture}[baseline={(0,.6)}]
    \node (5) at (1.4, 0) {$W_5$};
    \node (2) at  (0,1.2) {$W_2$};
    \node (4) at (0,0){$W_4$};
    \node (1) at (-1.4,1.2) {$W_1$};
    \draw[->] (1) -- (4);
    \draw[->] (4) -- (5);
    \draw[->] (2) -- (4);
\end{tikzpicture} \hspace{0.3cm}
\caption{\color{new}Adding edges between $X=W_3$ and $Y=W_4$ introduces changes in the marginal graphs.}\label{fig:example}
    \end{figure}
Hence, the question arises of how to identify these unlinked pairs 
with only the marginal graphs available. 
\new{For example, Figure \ref{fig:example} shows how if we would add 
the edge $W_3 \to W_4$ or $W_3 \leftrightarrow W_4$ to $G$,  this introduces additional edges in the marginal graphs, represented by the blue and orange edges. Therefore, the marginal graphs allow us to conclude that these edges are not in $G$.  Similarly, one can find that $W_3 \leftarrow W_4 \notin G$. In general, even though  $X$ does not appear in $G_Y$ as a node, it indirectly shapes the graph. As a confounder, that is $C \leftarrow X \rightarrow D \in G$, it creates the edge $C \leftrightarrow D \in G_{Y}$, and as a mediator $P \to X \rightarrow C \in G$, it leads to $P \to C \in G_Y$.} 
These relations allow us to deduce the absence of an edge, as formalized in the following lemma.
\begin{lemma}[excluding links in ADMGs]\label{lem:nolinks_ADMG} Let $G$ an ADMG whose marginalizations are $G_X$ and $G_Y$.
If $X$ has a child in $G_X$ that is neither a sibling nor a child of $Y$ in $G_Y$, or the same holds with  reversed the roles of $X$ and $Y$, then $X\centernot{-}Y$ in $G$.
\end{lemma} 
\new{For the graph $G$ depicted in Figure \ref{fig:example}, the lemma allows to deduce $W_3 \centernot{-} W_4$ as $W_4$ has the child $W_5$ in $G_{W_4}$, which is not a child or sibling of $W_3$ in $G_{W_3}$. Similarly, one can conclude $W_1 \centernot{-} W_5$ and $W_2 \centernot{-} W_5$. In contrast, the lemma can neither certify $W_1 \centernot{-} W_2$ nor $W_3 \centernot{-} W_5$.}

\new{If we know that $G$ is a DAG, we can decide whether an edge exists and, if so, determine its type in even more cases. Knowing that $G$ contains no bidirected edges, we can trace each bidirected edge $C \leftrightarrow D \in G_X$ back to $C$ and $D$ being children of $Y$ in the joint model.  Continuing the previous example, but now looking at the pair $(W_1, W_2)$, the bidirected edge $W_3 \leftrightarrow W_4 \in G_{W_2}$ implies that $W_3, W_4$ are both children of  $W_1$, compare Figure \ref{fig:example_causal_sufficiency}. Therefore, if there would be an edge $W_1\to W_2 \in G$, the edges $W_3\leftrightarrow W_2, W_4\leftrightarrow W_2$ would appear in $G_{W_2}$, which they do not.  Thus, $W_1 \not\to W_2 \in G$. The next lemma summarizes all conclusions we can make.} 

    \begin{figure}
    \centering
    $G$ = \begin{tikzpicture}[baseline={(0,.6)}]
    \node (3) at (-1.4, 0) {$W_3$};
    \node (5) at (1.4, 0) {$W_5$};
    \node (2) at  (0,1.2) {$W_2$};
    \node (4) at (0,0){$W_4$};
    \node (1) at (-1.4,1.2) {$W_1$};
    \draw[->, matplotlib_blue] (1) to (2);
    \draw[->] (1) -- (4);
    \draw[->] (1) -- (3);
    \draw[->] (2) -- (3);
    \draw[->] (4) -- (5);
    \draw[->] (2) -- (4);
\end{tikzpicture} \hspace{0.3cm}
    $G_{W_2}$ = \begin{tikzpicture}[baseline={(0,.6)}]
    \node (3) at (-1.4, 0) {$W_3$};
    \node (5) at (1.4, 0) {$W_5$};
    \node (2) at  (0,1.2) {$W_2$};
    \node (4) at (0,0){$W_4$};
    \draw[<->] (3) -- (4);
    \draw[-> ,transform canvas={xshift=1.5pt, yshift=-1.5pt}] (2) -- (3);
    \draw[<-> ,matplotlib_blue, transform canvas={xshift=-1.5pt, yshift=1.5pt}] (2) -- (3);
    \draw[->] (4) -- (5);
    \draw[->,transform canvas={xshift=2pt}] (2) -- (4);
    \draw[<->,matplotlib_blue,transform canvas={xshift=-2pt}] (2) -- (4);
\end{tikzpicture} \hspace{0.3cm}
\caption{\color{new}When $G$ is known to be causally sufficient, $W_1\not\to W_2$ can be inferred based on just $G_{W_2}$.}\label{fig:example_causal_sufficiency}
    \end{figure}
\begin{lemma}[determining edge types in DAGs]\label{lem:determine_edge_type_DAG} Assume $G$ is a DAG with marginalizations $G_X, G_Y$. Then $X$  has at least two children in $G$ if and only if $G_Y$ contains a bidirected edge. Moreover,
\begin{enumerate}[(1)]
    \item if $X$ has at least two children \new{in $G$}, then  $X \to Y \in G$ if and only if $Y$ has a sibling in $G_{\new{Y}}$.
    \item If $Y$ has at least two children \new{in $G$}, and $X$ has fewer, then $X \to Y \new{\in G}$ if and only if $X$ has multiple children in $G_{\new{X}}$.
    \item In the case that both have fewer than two children \new{in $G$}, 
    \begin{enumerate}[(a)]
        \item if $ch^{G_X}(X) \neq ch^{G_Y}(Y)$, then $X \centernot{-} Y \new{\in G}$;
        \item if neither $\pa^{G_X}(X) \subseteq \pa^{G_Y}(Y)$, nor vice versa, then $X \centernot{-} Y \new{\in G}$;
    \item if $ch^{G_X}(X) = ch^{G_Y}(Y) = \{C\}$, and neither $\pa^{G_Y}(Y)\subseteq \pa^{G_X}(X) \subseteq  \pa^{G_Y}(Y)\cup \pa^{G_Y}(C)$
    nor vice versa, then $X \centernot{-} Y \new{\in G}$;
        \item if $ch^{G_X}(X) = ch^{G_Y}(Y) = \{C\}$, and neither $\pa^{G_Y}(Y)\subseteq \pa^{G_X}(X) \subseteq  \pa^{G_Y}(Y)\cup \pa^{G_Y}(C)$
    nor at the same time $\pa^{G_X}(X) \subseteq \pa^{G_Y}(C)$ and $ \pa^{G_Y}(Y)\subseteq \pa^{G_X}(C)$, then $X \to Y \new{\in G}$.
    \end{enumerate}
    \item All the above criteria hold for reversed roles of $X$ and $Y$.
\end{enumerate}
\end{lemma} \new{This lemma leaves only a few exceptions where it does not make a statement about the edge type; in particular, it determines the edge type whenever $X$ or $Y$ has at least two children in $G$ or when they do not have precisely the same children in the marginal graphs. Moreover, t}he lemma is exhaustive in the sense that if none of the conditions apply, it is impossible to determine whether $X$ and $Y$ are linked. \new{In the example, it finds all five unlinked pairs.}

\new{One limitation of both lemmas is that they are restricted to causal discovery algorithms outputting ADMGs.  In contrast, many  algorithms learn DAGs or produce partially oriented graphs such as CPDAGs or PAGs. However, given the straightforward nature of the proof of the lemmas, it is unsurprising that analogous results can readily be derived for other graph types, see Appendix \ref{sec:LOVO_further_graph_types}.}

Once a pair is identified as unlinked, the next step is to derive a d- or m-separating set by inferring the union of parents and verifying their unconfoundedness with $X, Y$. If there is no edge $X - Y \in G$, the parents and the siblings of $X$ are the same in both $G_X$ and $G$, as are those of $Y$. This allows us to directly assess the parents and their unconfoundedness  directly from $G_X, G_Y,$ and leads to the following LOVO predictor.
\begin{theorem}[LOVO via parent adjustment]\label{thm:main}
Let all parents of $X$ be unconfounded. 
Likewise, let all parents of $Y$ be unconfounded. If $\bZ_S$ denotes the union of the parents of $X$ and $Y$,
then we have 
\begin{equation}\label{eq:lovod}
P(y|x) = \sum_{\bz_S} P(y|\bz_S) P(\bz_S|x).   
\end{equation}
\end{theorem}
\new{Overall, our method first employs Lemma \ref{lem:nolinks_ADMG} or \ref{lem:determine_edge_type_DAG} to determine whether $X$ and $Y$ are linked. 
Lemma \ref{lem:determine_edge_type_DAG} is used when assuming causal sufficiency in the joint model; otherwise, Lemma \ref{lem:nolinks_ADMG} is used. If the variables are unlinked, their parents are determined using $\pa_G(X)=\pa_{G_X}(X)$ and $\pa_G(Y)=\pa_{G_Y}(Y)$. To implement Theorem \ref{thm:main} in practice, we resort to the following simple procedure:}

\noindent{\bf Three-step LOVO predictor} 
\begin{enumerate}[(1)]
    \item Learn a predictor $\hat{P}(Y|\bZ)$ or a regression function  $\hat{f}$ with $\hat{f}(\bz):=\hat{\Exp}[Y|\bZ=\bz]$.
    \item Apply the predictor to the $\bz$-values of the pairs $(x, \bz)$ sampled from $P(X,\bZ)$ to generate 
artificial pairs $(x,\hat{y})$, with $\hat{y}$ sampled from $\hat{P}(y|\bz)$ or chosen as $\hat{y}:= \hat{f}(\bz)$, respectively. 
\item Use these pairs to learn the predictor $\hat{P}(Y|X)$, $\hat{\Exp}[Y|X]$ or $\hat{\rho}_{XY}$.
\end{enumerate}
As in the example on three nodes, Theorem \ref{thm:main} relies on conditional independence statements postulated by $G$. In contrast, the preceding steps to infer the non-existence of the edge and the joint parents from the marginal distributions employ the built-in inductive bias of causal models, particularly the faithfulness of the joint model. 
Although we do not claim that LOVO prediction necessitates causal models, the results in this section suggest that they are a natural way to solve this task  \new{and conditional independence information alone is insufficient for  LOVO prediction. This is a key distinction to the falsification approach in \cite{Biza2020}, where the estimated Markov blanket of a target variable $Y$ is used for predicting $Y$ from as few variables as possible, and the predictions are good whenever $Y$ is conditionally  independent of the remaining variables with respect to the estimated Markov blanket.}

\subsection{LOVO tailored to Linear non-Gaussian Acyclic Models (LiNGAM)}
Some causal discovery algorithms are based on structural equation models, such as the linear additive noise model (LiNGAM). The LiNGAM for the DAG $G$ postulates that  
    \begin{align}\label{eq:lingam}
W_i = \sum_{W_j\in \pa({i})} \lambda_{ij}W_j + N_j , \quad i=1,\ldots,k+2,
\end{align} where the $\lambda_{ij}$ are real coefficients 
and the $N_j$ are  independent centered non-Gaussian variables. Since the structure  matrix $\Lambda=(\lambda_{ij})$ collecting all coefficients is sparse according to the acyclic graph $G$, it can be transformed into a strictly lower triangular matrix through  simultaneous row and column permutations. 
We assume faithfulness, that is, for all edges $(i,j)\in G$, the total causal effect
\begin{equation}\label{eq:total_causal_effects}
  m_{ij} = \sum_{\substack{\pi \text{ directed path} \\ \text{from $j$ to $i$}}} \ \ \prod_{k \to l \text{ edge on }\pi} \lambda_{lk}
\end{equation}
is not zero. This assumption is fulfilled for Lebesgue almost all structure matrices $\Lambda$ compatible with a fixed graph $G$.
Reflecting the model, the corresponding algorithms commonly output not only a DAG but also the matrix $\Lambda$. Thus, when developing a LOVO predictor, it appears natural to incorporate the learned matrix in order to falsify the algorithm's entire output. Additionally, this enables LOVO prediction even if a direct link exists, as long as we can determine its type using Lemma \ref{lem:determine_edge_type_DAG}. 
\begin{theorem}[LOVO via LiNGAM]\label{thm:LOVO_via_lingam} Suppose $P^\bW$ follows a linear additive noise model for some DAG $G$. If Lemma \ref{lem:determine_edge_type_DAG} allows us to identify
the edge type between $X$ and $Y$, ($X \to Y, Y\to X,$ or $X \centernot{-} Y$), and the children sets are not $\text{ch}(Y)=\{X, Z_j\}$ and $\text{ch}(X)=\{Z_j\}$ or vice versa, 
then\footnote{Lemma \ref{lem:determine_edge_type_DAG} assumes $G_X, G_Y$ to be ADMGs, whereas most LiNGAM-based causal discovery algorithms produce DAGs entailing explicit latent nodes. However, such a DAG can be easily transformed into an equivalent ADMG by replacing each structure $W_1 \leftarrow L \to W_2$, with $L$ a latent node, by $W_1 \leftrightarrow W_2$.}
\begin{enumerate}[(1)]
    \item the structure matrix $\Lambda$ can be
 uniquely identified from $P (X, \bZ)$ and $P (Y, \bZ)$. 
 \item If, in addition, all second and higher order moments of $\bN$ are finite,  $P(X, \boldsymbol{Z})$ and $P(Y, \boldsymbol{Z})$ uniquely determines $P(X, Y, \boldsymbol{Z})$, except for a measure zero set of moments of $\bN$.
\end{enumerate}
\end{theorem}
As in the previous theorem, all graphical assumptions in the theorem can be verified from the marginal graphs. Knowing the structure matrix, we can construct the LOVO predictor as follows. Denoting $\bZ_S = \pa^{G}(Y)\cap \bZ$, and using that under the model assumptions $Y = \lambda_{Y\bZ_S}\bZ_S + \lambda_{YX} X + N_Y$, and $N_Y$ is centered and independent of $(X, \bZ_S)$, we obtain
\begin{equation*}
    \mathbb{E}(Y \mid X = x) = \lambda_{Y\bZ_S }\mathbb{E}(\bZ_S\mid X=x)+\lambda_{YX}x, 
\end{equation*}
which can be estimated from $P(X, \bZ)$.

\section{Baseline: LOVO prediction in absence of causal information \label{sec:baseline}}
Although the two proposed LOVO predictors provide a reasonable approximation of $\Exp(Y \mid X=x)$ whenever the marginal graphs $G_X, G_Y$ are accurate, some error will persist. To decide which level of deviation is still acceptable, we ask whether the causal information helped the prediction by comparing the prediction error to the error of the best LOVO predictor {\it without causal information}, also called baseline predictor.  
Note that one may consider $P(Y|X) = P(Y)$ (that is, assuming independence of $X$ and $Y$) as the best predictor in the absence of any causal knowledge. We reject this idea for two reasons:
First, the dependencies between $X,\bZ$ and between $Y,\bZ$ may be so strong that it is impossible that $X$ and $Y$ are independent, see No 4. in Table \ref{tab:dags} in the supplement, last column.
Second, the predictor  $P(Y|X) = P(Y)$ is unlikely to be the right one in graphs with several nodes unless one 
assume relatively sparse graphs. 

Instead of assuming independent $X,Y$ as the best "causally agnostic" predictor, we use the "MaxEnt prediction" \citep{Jaynes2003}, which is  the joint distribution that maximizes entropy subject to the given marginal distributions
$P(X,\bZ)$ and $P(Y,\bZ)$ \citep{Mejia2022}.
It is given by the unique joint distribution with $X\independent Y\,|Z$. To see this, note that 
the joint entropy reads \citep{cover}
\[
H(X,Y,Z) = H(X,Z) + H(Y|X,Z) = H(X,Z) + H(Y|Z) - I(Y:X\,|Z),
\]
which is maximal when the conditional mutual information  $I(Y:X\,|Z)$ vanishes. 

To justify MaxEnt as a reasonable approach for our purpose, we remind the reader of the intuition that
the MaxEnt distribution is the "most mixed" distribution within the set of distributions satisfying the given bivariate
constraints, which seems like a better compromise rather than choosing distributions closer to the boundary. 
 \cite{Gruenwald2004} provides a game-theoretic view on MaxEnt and 
 shows that it minimizes the worst-case 
 logarithmic cross-entropy loss. 

\begin{definition}[MaxEnt Baseline predictor]\label{def:maxent}
Given $P(X,\bZ)$ and $P(Y,\bZ)$, the MaxEnt baseline predictor is defined by
\begin{equation}\label{eq:baseline}
P^{\rm MaxEnt}(y|x) = \sum_\bz P(y|\bz) P(\bz|x).
\end{equation}
\end{definition} 
One can easily show, see Lemma \ref{lem:base} in the appendix, that the MaxEnt predictor is correct for all DAGs whose "moral graph" \footnote{{The moral graph  \citep{Lauritzen} is the undirected graph obtained by removing orientations and connecting parents of a common child.}} 
does not have an edge $X-Y$.  
Since the overall shape of the MaxEnt predictor aligns with the one of the parent adjustment LOVO predictor, we can again use the three-step procedure to estimate it from finite data. An important difference between the two predictors is that the MaxEnt baseline generally adjusts for more variables. However, comparison of regression models with an equal number of features is "fairer" with respect to statistical inaccuracies entailed by finite data. Therefore, when using the MaxEnt predictor as a baseline against the parent adjustment predictor, we recommend comparing against {\it random adjustment sets} $\bZ_R$  of equal size.  In this slightly modified version, the baseline is generically worse than parent adjustment whenever $P(X,Y,\bZ)$ is Markov to a graph $G$, in which $\bZ_R$ does not d-separate $X$ and $Y$.

\section{Experiments   \label{sec:exp}}
 The code to reproduce our experiments is available at GitHub\footnote{\url{https://github.com/DanielaSchkoda/LOVO}.}.
\subsection{LOVO prediction given the true marginal graphs $G_X, G_Y$ \label{subsec:experiment_parents}} 
\begin{figure}
\centering
\includegraphics[width=0.32\textwidth]{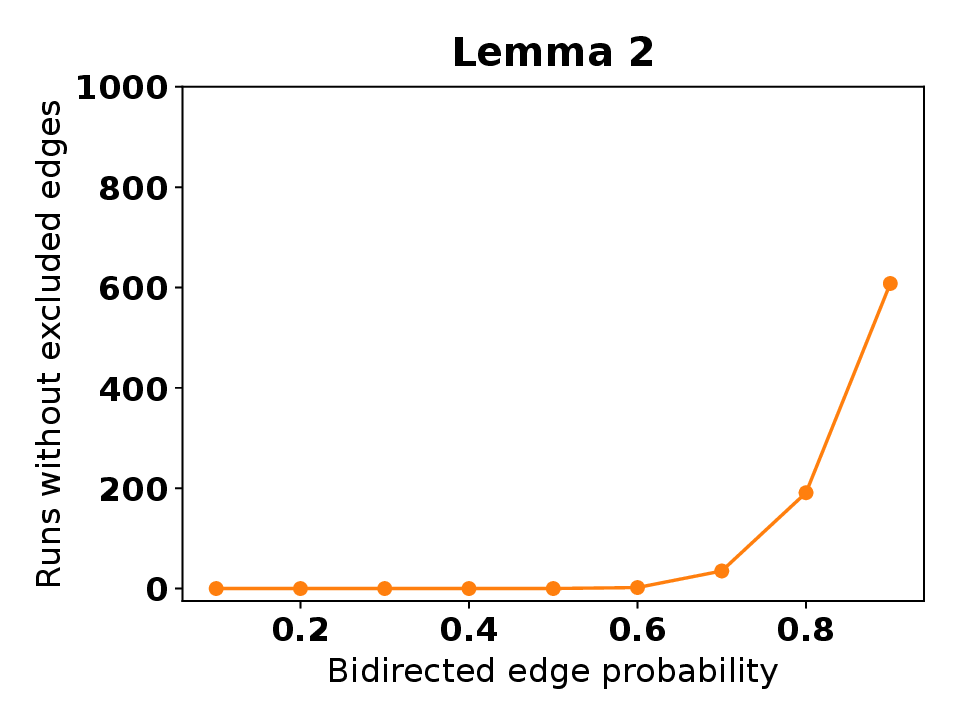}
\includegraphics[width=0.32\textwidth]{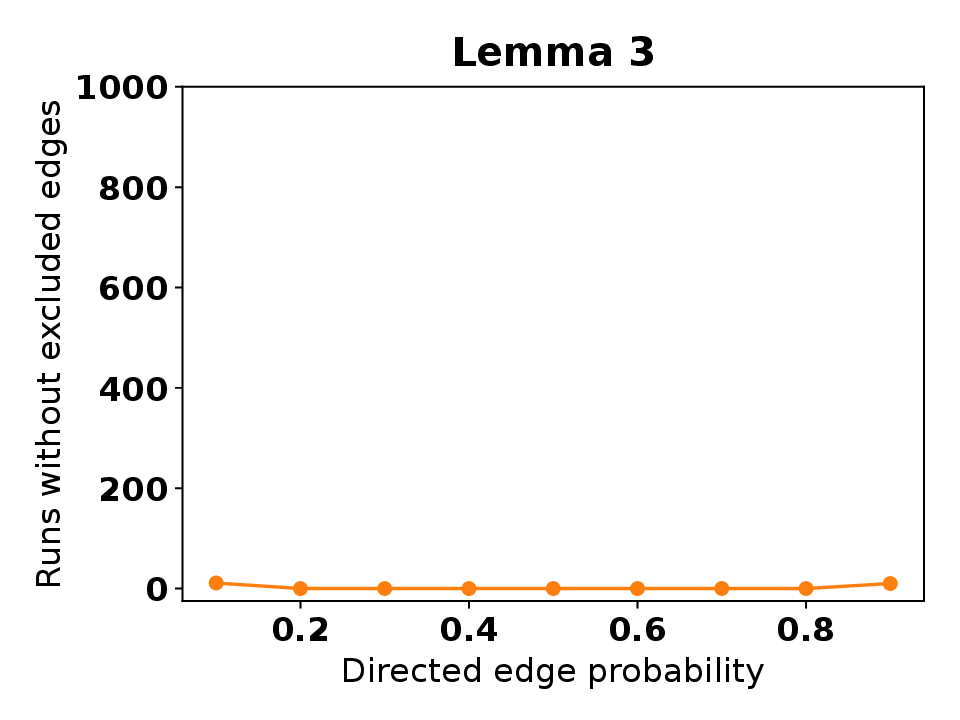}
\caption{For Lemma \ref{lem:nolinks_ADMG} and small values of $q$, 
and Lemma \ref{lem:determine_edge_type_DAG} regardless of $p$, only in a few graphs no single unlinked pair  can be detected, so LOVO is realizable in most cases.}\label{fig:no_excluded_edges}
\end{figure}

First, we want to shed light on the question of how frequently Lemmas \ref{lem:nolinks_ADMG} - \ref{lem:determine_edge_type_DAG} succeed in excluding edges. 
To this end, we randomly generate $1000$ Erdős–Rényi DAGs on $10$ nodes by choosing a random ordering and then inserting each edge with probability $p$ varying between $0.1$ and $0.9$. For each generated graph, we check if, for at least one pair of nodes $(X, Y)$, we can rule out that they are linked based on the marginal graphs $G_X, G_Y$. For Lemma \ref{lem:nolinks_ADMG}, we use ADMGs instead of DAGs; they are generated following the same procedure, except that we fix $p=0.3$ and additionally include bidirected edges with a probability $q \in [0.1, 0.9]$. Figure \ref{fig:no_excluded_edges} shows in how many runs no single pair without edge can be found, and therefore, LOVO prediction would not be possible. 
In Appendix \ref{subsec:add_details_performance_lemmas}, we illustrate the average number of pairs 
identified  as unlinked in each graph.

\begin{figure}
	\centering
\includegraphics[width=0.32\textwidth]{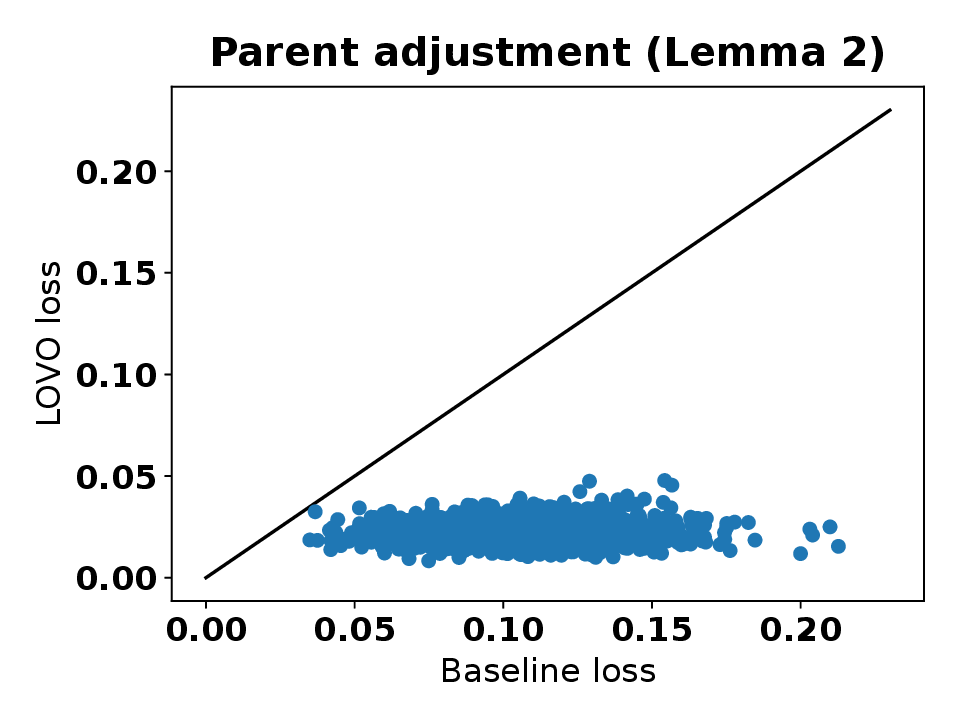}
\includegraphics[width=0.32\textwidth]{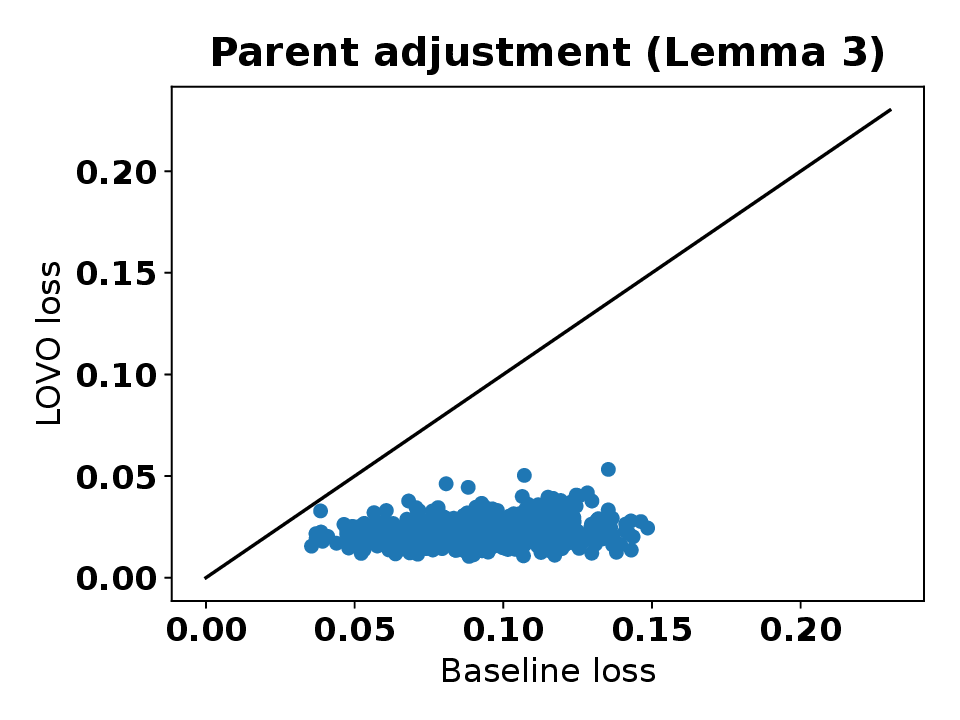}
\includegraphics[width=0.32\textwidth]{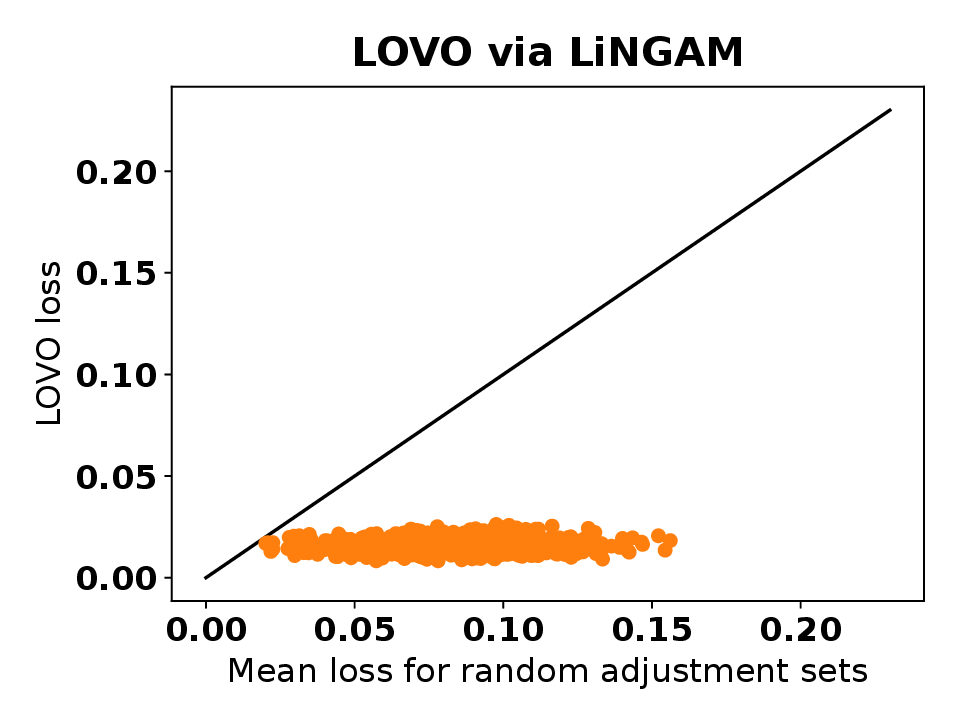}
\caption{\label{fig:parent_adjustment} 
When provided with the true marginal graphs $G_X$ and $G_Y$, the parent adjustment LOVO predictor and the LiNGAM LOVO predictor outperform the baseline.}
\end{figure}

Next, we assess the parent adjustment LOVO predictor, where we use the correlation $\rho$ between $X$ and $Y$ as the estimation target. We generate graphs as above with $p=0.3$, and, for ADMGs, $q=0.1$. To obtain data in accordance with the graphs,  we employ a linear additive noise model, with noise uniformly distributed on $[-1,1]$ and coefficients drawn uniformly from $[-1,-0.5]\cup [0.5, 1]$. We set the sample size to $n=5000$. Again, based on the true marginal graphs $G_X, G_Y$, for each pair $(X,Y)$, we evaluate whether they might be linked. If not, we compute the three-step LOVO predictor $\hat{\rho}^{\text{LOVO}}$ according to
Theorem \ref{thm:main}, as well as the baseline predictor $\hat{\rho}^{\text{Base}}$, for which we calculate a MaxEnt predictor with a random adjustment set of the same size as the union of parents multiple times, and then take the average. Moreover, we directly calculate the sample correlation coefficient $\hat{\rho}$ from $P(X, Y)$ in order to estimate the prediction errors $\hat{\rho}^{\text{Base}} - \hat{\rho}$, $\hat{\rho}^{\text{LOVO}} - \hat{\rho}$. For a more accurate error assessment, in the above steps, we never use all samples; instead, we split the data into three parts of sample size $n/3$ each. The first two parts are used for $P(X, \bZ)$ and $P(Y, \bZ)$, respectively, required in the three-step procedure, while the third part is reserved to estimate $\hat{\rho}$.
Finally, we average the results across all pairs to derive the cross-validation errors $CV^{\text{LOVO}}, CV^{\text{Base}}$, which are compared in Figure \ref{fig:parent_adjustment} (left and middle). As before, we use ADMGs combined with Lemma \ref{lem:nolinks_ADMG} and DAGs with Lemma \ref{lem:determine_edge_type_DAG}. In the ADMG setting, the LOVO predictor abstains in $0.5\%$ of the replications, and in the DAG setting, never.
\new{The supplement includes similar experiments for other graph types.}

To analyze the LOVO via LiNGAM predictor, we sample DAGs and data as before and again use the correlation as the estimation target. The right plot in Figure \ref{fig:parent_adjustment} compares the prediction error of LOVO to the MaxEnt baseline predictor with all variables $\bZ$ as the adjustment set. Again, the LOVO predictor never abstains.
 
\subsection{LOVO applied to DirectLiNGAM and RCD}\label{subsec:experiment_RCD_DirectLiNGAM}
\begin{figure}
    \centering
   \includegraphics[width=0.32\linewidth]{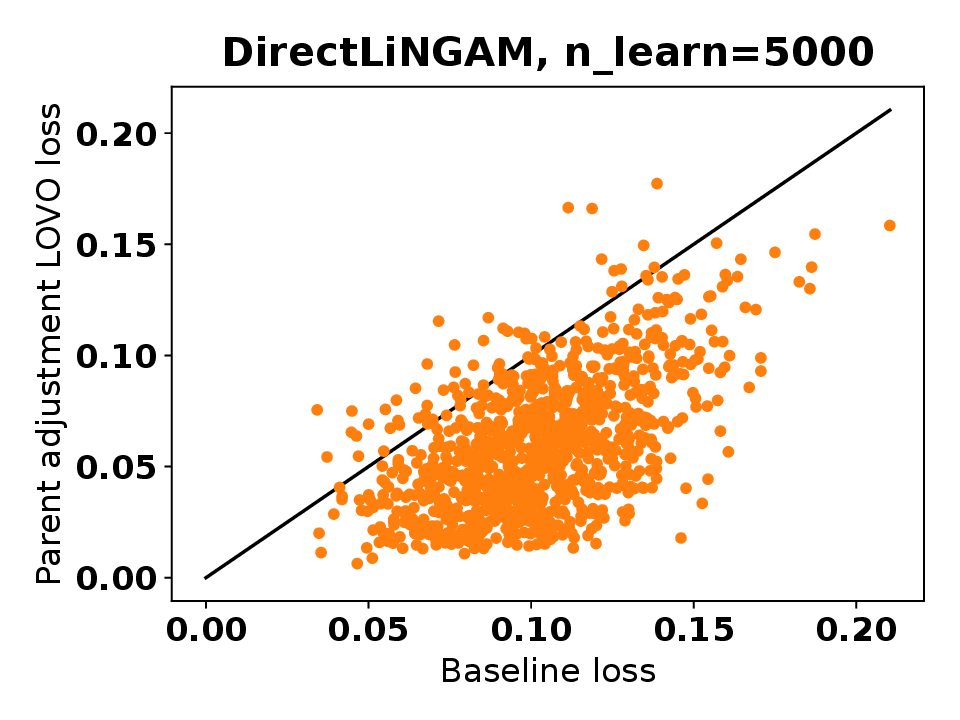}
    \includegraphics[width=0.32\linewidth]{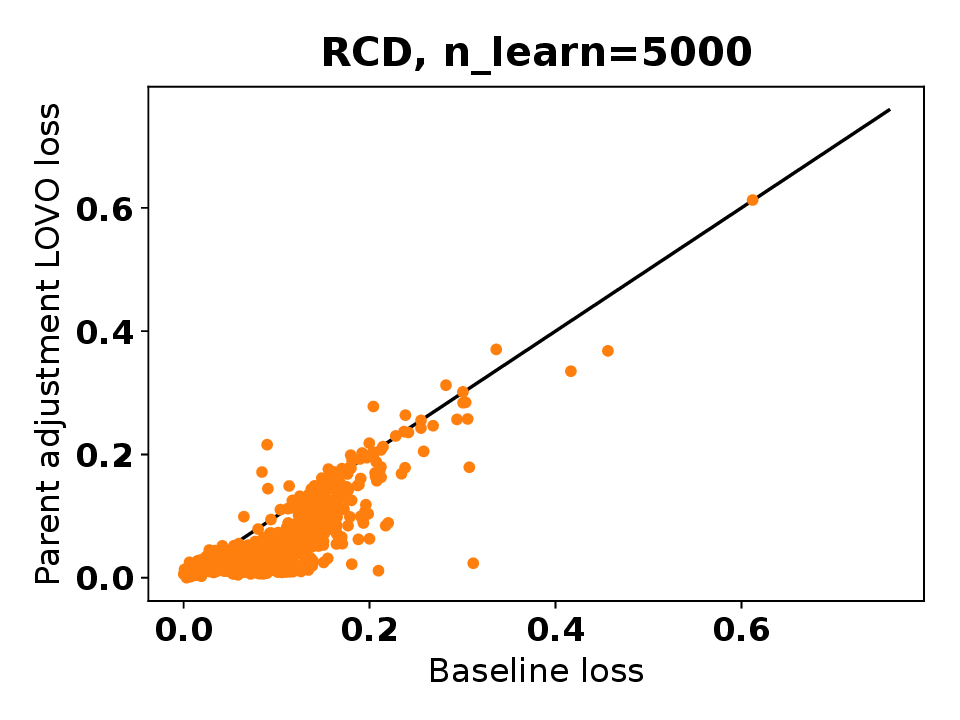}
    \includegraphics[width=0.32\linewidth]{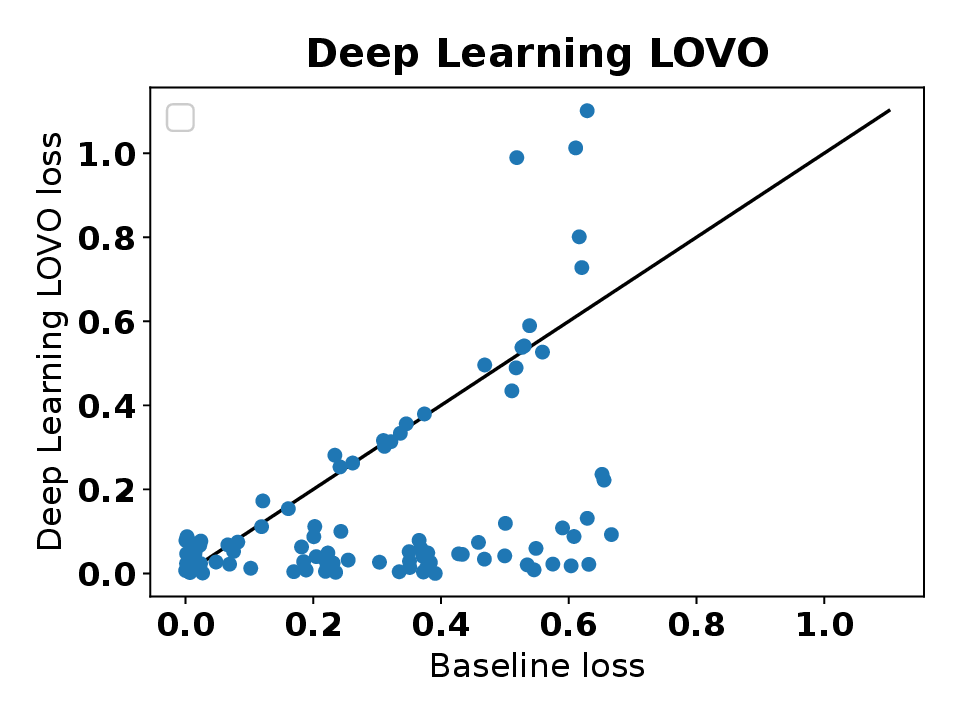}
    \caption{The scatter plots show LOVO versus baseline loss for parent adjustment LOVO applied to graphs estimated with DirectLiNGAM and RCD; and for DL LOVO prediction.}
    \label{fig:applied_experiments}
\end{figure}

Next, we apply the LOVO predictor to two causal discovery algorithms, namely DirectLiNGAM \citep{DirectLiNGAM} and Repetitive Causal Discovery (RCD) \citep{maeda2020rcd}. The first method produces DAGs, and the second one uses the alternative  definition of ADMGs, which forbids the co-occurrence of a directed and a bidirected edge. Correspondingly, we use the slightly adjusted LOVO predictors described in Appendix \ref{subsec:LOVO_ADMGs_without_confounded_links}-\ref{subsec:DAGs}. We sample DAGs and the data as before, but double the sample size and use the first half to learn $G_X, G_Y$. For DirectLiNGAM, we use $p=0.5$ to ensure that Lemma \ref{lem:exclude_links_directed_part} often applies. For RCD, we adhere to $p=0.3$ but decrease the number of nodes to 5 because of its slower execution time. Figure \ref{fig:applied_experiments}  (left and middle) compares the LOVO cross-validation error to the baseline. For DirectLiNGAM, LOVO abstained in 23\% of the cases, and for RCD in 3\%. To examine whether the LOVO cross-validation error indeed increases with the number of mistakes in the learned graphs, we repeat the above experiment with varying sample sizes for learning the graphs, specifically, $n_{\text{learn}}=100, 500, 1000, 5000$. In Appendix \ref{subsec:varying_n}, we plot the LOVO loss for each value of $n_{\text{learn}}$. Moreover, we concatenate all the results to  calculate the Spearman correlation coefficient of the LOVO cross-validation error and
\begin{enumerate}[1.]
    \item whether an edge $X-Y$ exists in $G$, averaged over all $(X, Y)$ used in the cross-validation.
    \item the sum of the Structural Hamming Distances (SHDs) of $\hat{G}_X$ to $G_X$ and of $\hat{G}_Y$ to $G_Y$, averaged over all $(X, Y)$ used in the cross-validation.
\end{enumerate}
\begin{figure}
    \centering
\includegraphics[width=0.24\linewidth]{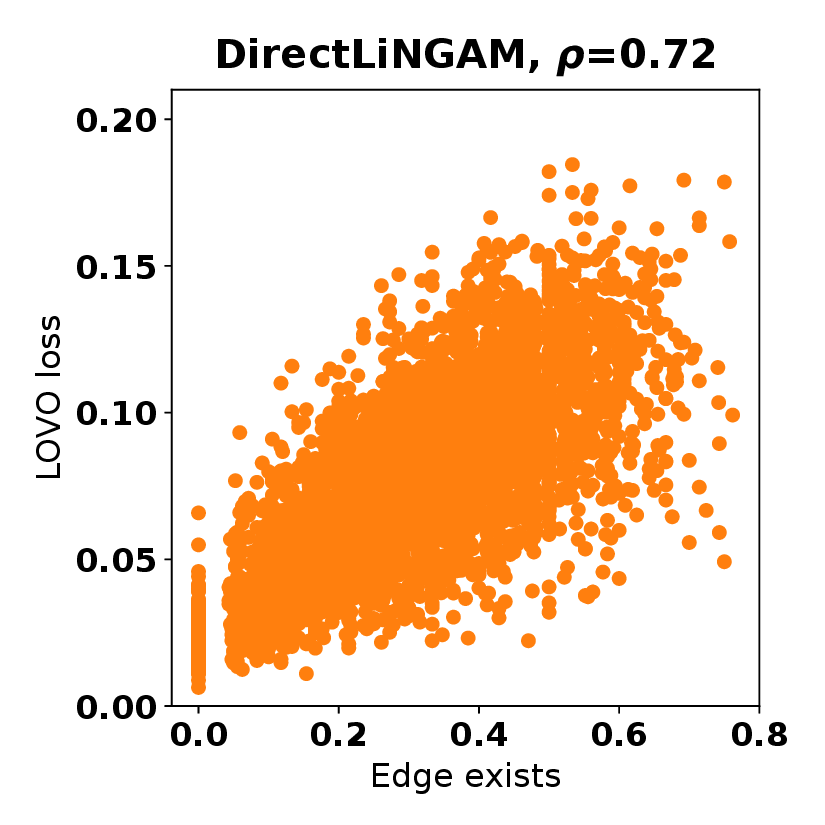}
\includegraphics[width=0.24\linewidth]{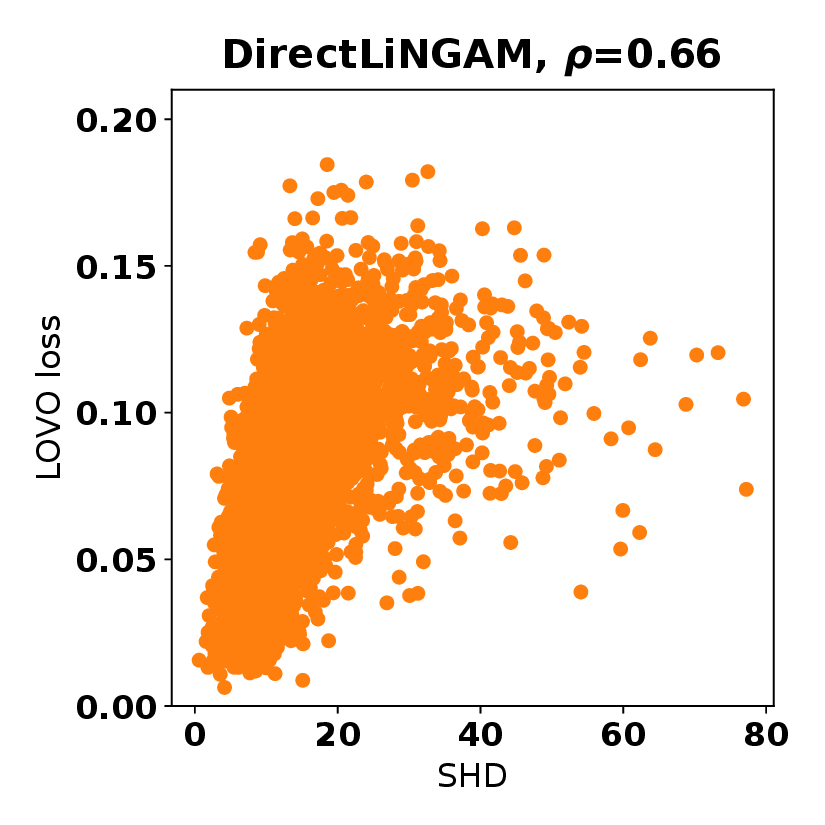}
\includegraphics[width=0.24\linewidth]{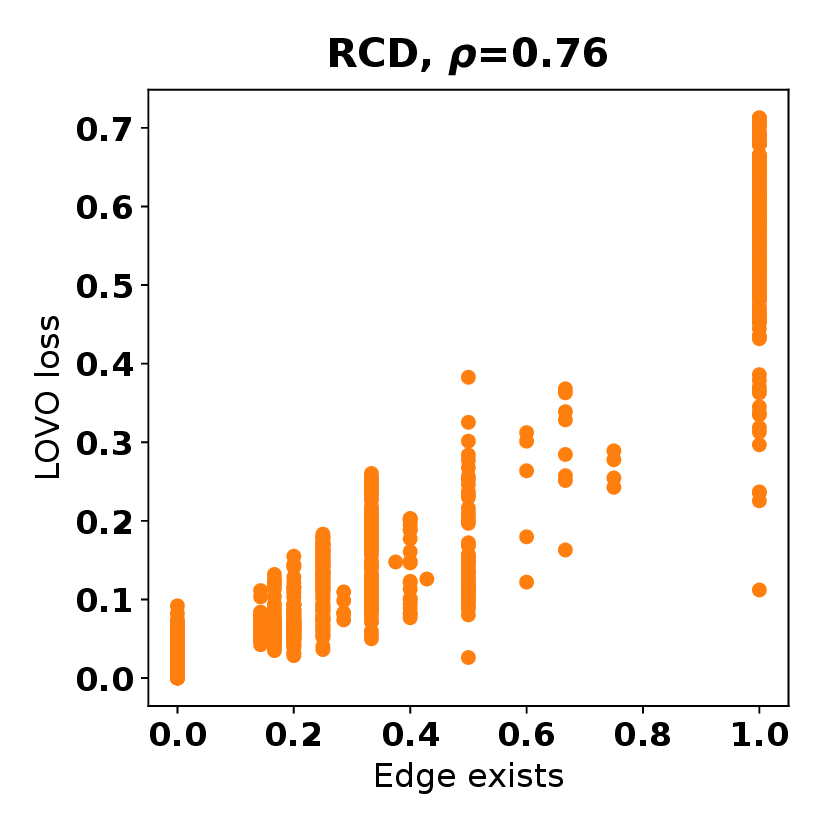}
\includegraphics[width=0.24\linewidth]{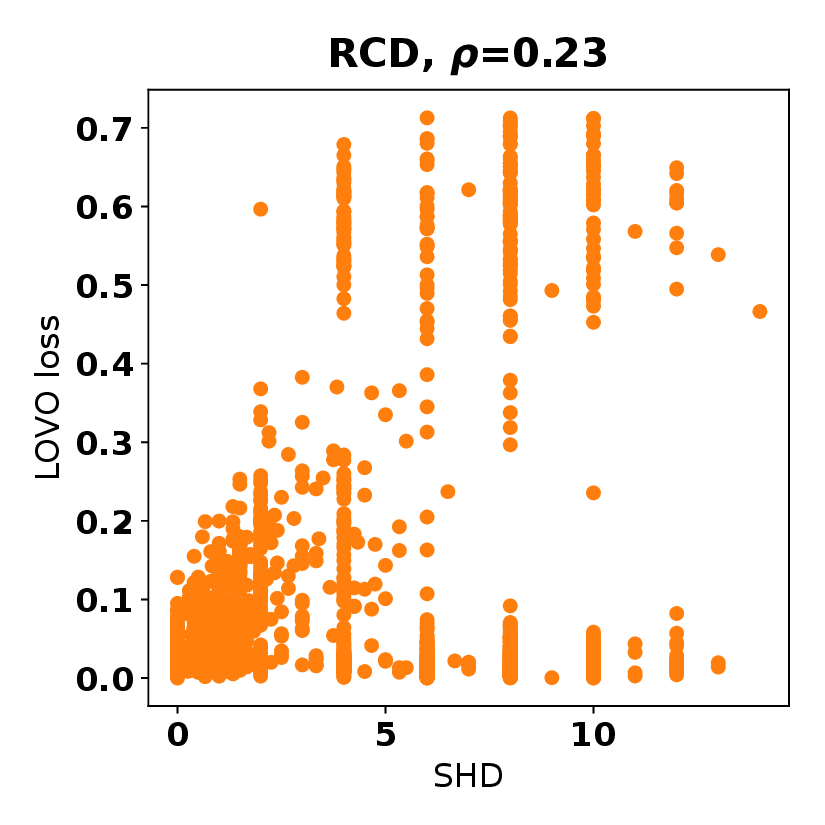}
    \caption{The scatter plots show how LOVO performance correlates with causal discovery performance. The LOVO error increases with the number of pairs misidentified as unlinked and with the SHD. The corresponding Spearman correlation coefficients included in the titles all significantly deviate from zero, with $p$-values $0.0,0.0,0.0,$ and $ 4\cdot10^{-44}$.}
    \label{fig:correlations}
\end{figure}
Including the first measurement is motivated by the fact that the parent adjustment LOVO predictor relies on 
the absence of an edge and, therefore, mistakes here are expected to lead to inaccurate LOVO prediction. The second measurement more straightforwardly evaluates the accuracy of the learned graphs. We obtain significant positive correlations in all cases, as presented in 
Figure \ref{fig:correlations}.

Appendix \ref{subsec:real-world}  includes  experiments on real-world data.
\subsection{Training DL for trivariate LOVO} 
\label{subsec:dl}
\begin{figure}
        \centering
\includegraphics[width=.7\linewidth]{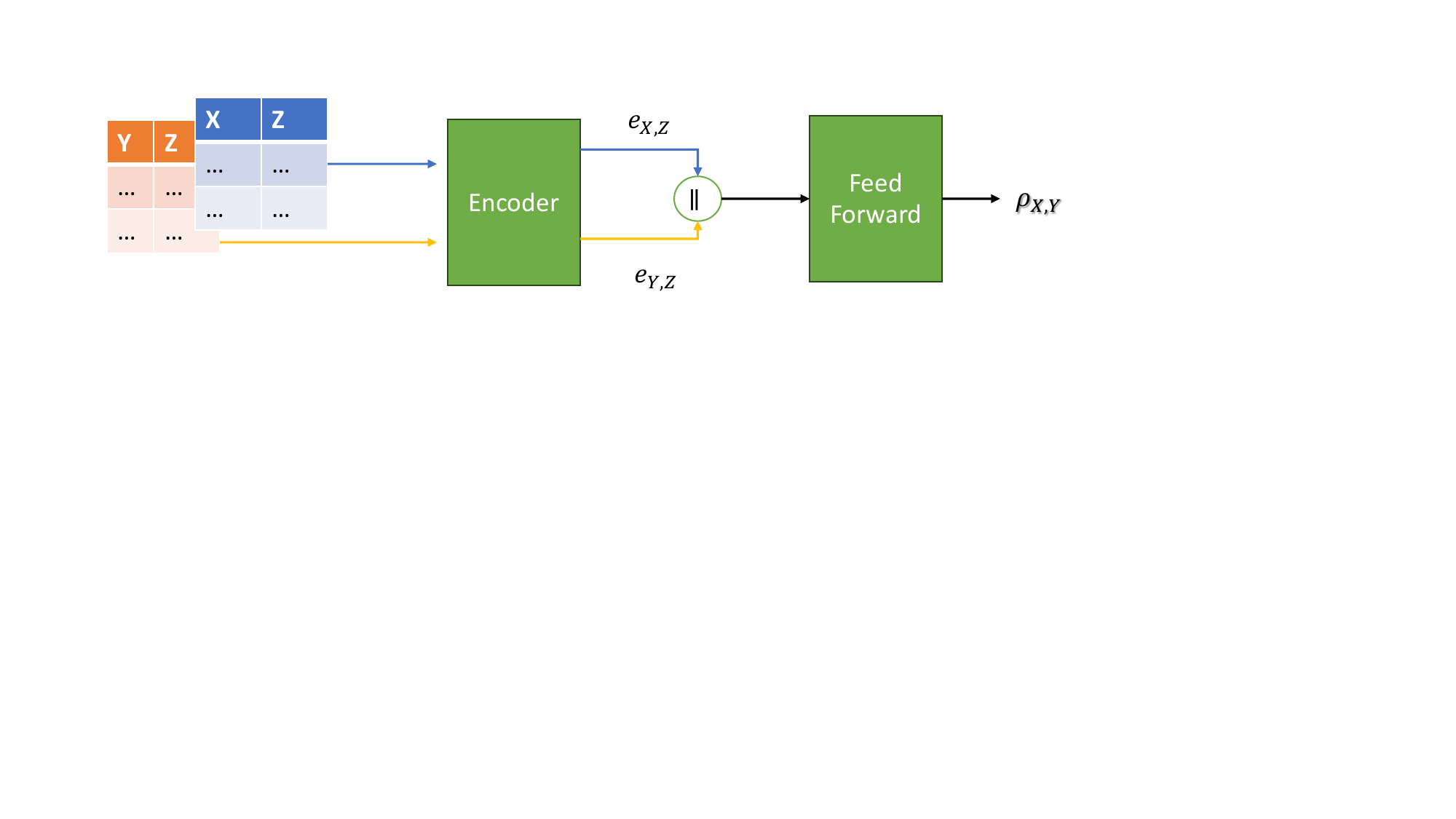}
\caption{Architecture of our DL LOVO predictor: the encoder learns appropriate features of the two marginal distributions from which the correlation of $X$ and $Y$ is inferred.} 
\label{fig:architecture}
\end{figure} So far, we have constructed LOVO predictors either by restricting to unlinked pairs or by
 assuming LiNGAM. 
 To support the hypothesis that also causal models without such restrictive assumptions help for 
LOVO predictions, we now show that a deep learning architecture that has been proposed for causal discovery can be modified to a LOVO predictor. 
To this end, we built on 
\cite{ke2022learning}, who use a transformer-based architecture to directly infer the adjacency matrix of the causal graph from a given dataset. 
We apply the encoder part from their architecture to each marginal dataset to get representation vectors $e_{X, Z}$ and $e_{Y, Z}$.
These representations are concatenated and used as input to a feed-forward layer (see Figure \ref{fig:architecture} for an overview and Section \ref{sec:additional_details} in the appendix for more details). 
This way, the network can be trained to predict the correlation $\rho_{X, Y}$ from given marginal datasets. \new{We want to stress that we do not input any parametric model assumptions in the architecture and, therefore, also do not input any preconceptions about how to construct a LOVO predictor.}
Figure \ref{fig:applied_experiments} (right) shows that it outperforms our  baseline in most cases.
To shed light on the challenging question of whether our DL LOVO predictor implicitly 
learns a causal representation, we try to predict the causal structure between $X$ and $Z$ from the learned representation $e_{X, Z}$.
If a second model could learn to map the representation to the causal structure, 
this suggests that the learned features are suitable for both tasks. 
To this end, we train a simple feed-forward network to predict the underlying causal structure, encoded as categories $\{\to, \gets, \leftrightarrow,  \centernot{-}\}$ (see again Section \ref{sec:additional_details} for more details).
As a na\ "ive baseline, we consider the average training label (where the categories are represented via one-hot-encoding).
Indeed, we can predict the causal structure better than our baseline (see Figure \ref{fig:structure_cross_entropy} in the appendix).

\section{Conclusions}
We have shown that causal hypotheses built via applying causal discovery to 
two Leave-One-Variable-Out datasets 
can, in principle, enable the prediction of the statistical relations between the two variables $X,Y$ that were dropped. As a concrete LOVO predictor, we first propose prediction via adjusting for parents, which relies on the absence of edges and is applicable to general causal discovery algorithms that produce DAGs or ADMGs. We further demonstrate how LOVO prediction can be customized for specific structural equation models, such as LiNGAM, enabling prediction even when a direct link is present. In simulation experiments, we observe a significant correlation between the LOVO prediction error and the accuracy of the estimated causal graphs. This reinforces our conjecture that the goodness of LOVO predictions can be utilized to evaluate (inferred) causal relationships. 
\acks{This project has received funding from the European Research Council (ERC) under the European Union's Horizon 2020 research and innovation programme (grant agreement No 883818). 
DS acknowledges support by the DAAD program Konrad Zuse Schools of Excellence in Artificial Intelligence, sponsored by the Federal Ministry of Education and Research. 
PMF was supported by a doctoral scholarship of the
Studienstiftung
des deutschen Volkes (German Academic Scholarship
Foundation).}

\bibliography{references}

@book{Spirtes1993,
  author = {Peter Spirtes and Clark Glymour and Richard Scheines},
  title = {Causation, Prediction, and Search},
  year = {1993},
  address = {New York, NY},
  publisher = {Springer-Verlag},
}

@article{Glymour2019,
  author = {Clark Glymour and Kun Zhang and Peter Spirtes},
  title = {Review of causal discovery methods based on graphical models},
  year = {2019},
  journal = {Frontiers in Genetics},
  volume = {10},
}

@article{Shimizu2006,
  author = {Shimizu, Shohei and Hoyer, Patrik O. and Hyv\"{a}rinen, Aapo and Kerminen, Antti},
  title = {A linear {non-Gaussian} acyclic model for causal discovery},
  year = {2006},
  journal = {Journal of Machine Learning Research},
  volume = {7},
  pages = {2003-2030},
}

@article{Grace_2016,
 author = {Grace, J. B. and Anderson, T. M. and Seabloom, E. W. and Borer, E. T. and Adler, Peter B. and Harpole, W. Stanley and Hautier, Yann and Hillebrand, Helmut and Lind, Eric M. and P{\"a}rtel, Meelis and Bakker, Jonathan D. and Buckley, Yvonne M. and Crawley, Michael J. and Damschen, Ellen I. and Davies, Kendi F. and Fay, Philip A. and Firn, Jennifer and Gruner, Daniel S. and Hector, Andy and Knops, Johannes M. H. and MacDougall, Andrew S. and Melbourne, Brett A. and Morgan, John W. and Orrock, John L. and Prober, Suzanne M. and Smith, Melinda D.},
 year = {2016},
 title = {{Integrative modelling reveals mechanisms linking productivity and plant species richness}},
 pages = {390--393},
 volume = {529},
 number = {7586},
 journal = {{Nature}}
}

@article{Gnecco2019,
    AUTHOR = {Gnecco, Nicola and Meinshausen, Nicolai and Peters, Jonas and Engelke, Sebastian},
     TITLE = {Causal discovery in heavy-tailed models},
   JOURNAL = {The Annals of Statistics},
  FJOURNAL = {The Annals of Statistics},
    VOLUME = {49},
      YEAR = {2021},
    NUMBER = {3},
     PAGES = {1755--1778},
}

@article{Nauta2019,
  author = {Meike Nauta and Doina Bucur and Christin Seifert},
  title = {Causal discovery with attention-based convolutional neural networks},
  year = {2019},
  journal = {Machine Learning and Knowledge Extraction},
  volume = {1},
  number = {1},
  pages = {312-340},
}

@article{Peters15,
  author = {Peters, Jonas and Bühlmann, Peter and Meinshausen, Nicolai},
    title = {Causal Inference by using Invariant Prediction: identification and Confidence Intervals},
    journal = {Journal of the Royal Statistical Society Series B: Statistical Methodology},
    volume = {78},
    number = {5},
    pages = {947-1012},
    year = {2016}
}

@article{Mooij2020,
  author  = {Joris M. Mooij and Sara Magliacane and Tom Claassen},
  title   = {Joint Causal Inference from Multiple Contexts},
  journal = {Journal of Machine Learning Research},
  year    = {2020},
  volume  = {21},
  number  = {99},
  pages   = {1-108},
}

@article{Rothenhaeuser2021,
  author = {Rothenh\"ausler, Dominik and Meinshausen, Nicolai and B\"uhlmann, Peter and Peters, Jonas},
  title = {Anchor regression: heterogeneous data meets causality},
  year = {2021},
  journal = {Journal Royal Statistical Society Series B},
  pages = {215-246},
  volume = {83},
}

@article{Lagemann2023,
  author = {Kai Lagemann and Christian Lagemann and Bernd Taschler and Sach Mukherjee},
  title = {{D}eep learning of causal structures in high dimensions under data limitations},
  year = {2023},
  journal = {Nature Machine Intelligence},
  volume = {5},
  number = {11},
  pages = {1306-1316},
}

@article{Hamilton1989,
  title={New method for generating deletions and gene replacements in Escherichia coli},
  author={Carol M. Hamilton and Mart{\'i} Aldea and Brian K. Washburn and Paul Babitzke and Sidney R Kushner},
  journal={Journal of Bacteriology},
  year={1989},
  volume={171},
  pages={4617 - 4622},
}

@article{sachs2005causal,
  author = {Karen Sachs and Omar Perez and Dana Pe'er and Douglas A. Lauffenburger and Garry P. Nolan},
  title = {Causal protein-signaling networks derived from multiparameter single-cell data},
  year = {2005},
  journal = {Science},
  volume = {308},
  number = {5721},
  pages = {523-529},
}

@article{Imbens2020,
  author = {Guido W. Imbens},
  title = {Potential outcome and directed acyclic graph approaches to causality: relevance for empirical practice in economics},
  year = {2020},
  monthopt = {December},
  journal = {Journal of Economic Literature},
  volume = {58},
  number = {4},
  pages = {1129-79},
}

@inproceedings{Faller2024,
  author = {Philipp M. Faller and Leena Chennuru Vankadara and Atalanti A. Mastakouri and Francesco Locatello and Dominik Janzing},
  title = {Self-compatibility: evaluating causal discovery without  Ground Truth},
  booktitle = 	 {Proc. 27th International Conference on Artificial Intelligence and Statistics},
  pages = 	 {4132--4140},
  year = 	 {2024},
  editoropt = 	 {Dasgupta, Sanjoy and Mandt, Stephan and Li, Yingzhen},
  volume = 	 {238},
  monthopt = 	 {02--04 May},
  publisher =    {PMLR}
}

@article{Mohan2021,
  author = {Karthika Mohan and Judea Pearl},
title = {Graphical Models for Processing Missing Data},
journal = {Journal of the American Statistical Association},
volume = {116},
number = {534},
pages = {1023--1037},
year = {2021},
}

@article{Stone1974,
  author = {Mervyn Stone},
  title = {Cross-Validatory Choice and Assessment of Statistical Predictions},
  year = {2018},
  journal = {Journal of the Royal Statistical Society: Series B (Methodological)},
  volume = {36},
  number = {2},
  pages = {111-133},
}

@article{Tsamardinos,
  title={Towards Integrative Causal Analysis of Heterogeneous Data Sets and Studies},
  author={Tsamardinos, Ioannis and Triantafillou, Sofia and Lagani, Vincenzo},
  journal={Journal of Machine Learning Research},
  year={2012},
  volume={13},
  pages={1097-1157},
}

@InProceedings{Gresele2022,
  title = 	 {Causal Inference Through the Structural Causal Marginal Problem},
  author =       {Gresele, Luigi and K{\"u}gelgen, Julius Von and K{\"u}bler, Jonas and Kirschbaum, Elke and Sch{\"o}lkopf, Bernhard and Janzing, Dominik},
  booktitle = 	 {Proc. 39th International Conference on Machine Learning},
  pages = 	 {7793--7824},
  year = 	 {2022},
  editoropt = 	 {Chaudhuri, Kamalika and Jegelka, Stefanie and Song, Le and Szepesvari, Csaba and Niu, Gang and Sabato, Sivan},
  volume = 	 {162},
  monthopt = 	 {17--23 Jul},
  publisher =    {PMLR}
}

@inproceedings{Guo2024,
  author = {Siyuan Guo and Jonas Wildberger and Bernhard Schölkopf},
    booktitle = {Proc. 12th International Conference on Learning Representations},
  title = {Out-of-Variable Generalization for Discriminative Models},
  year = {2024},
}

@book{Pearl:00,
  author = {Judea Pearl},
  title = {Causality},
  year = {2000},
  publisher = {Cambridge University Press},
}

@article{richardson2003markov,
  author = {Thomas Richardson},
  title = {{M}arkov properties for acyclic directed mixed graphs},
  year = {2003},
  journal = {Scandinavian Journal of Statistics},
  volume = {30},
  number = {1},
  pages = {145-157},
}

@article{Xinpeng2020,
author = {Shen, Xinpeng and Ma, Sisi and Vemuri, Prashanthi and Simon, Gyorgy and Weiner, Michael and Aisen, Paul and Petersen, Ronald and Jack, Clifford and Saykin, Andrew and Jagust, William and Trojanowki, John and Toga, Arthur and Beckett, Laurel and Green, Robert and Morris, John and Shaw, Leslie and Khachaturian, Zaven and Sorensen, Greg and Carroll, Maria and Fargher, Kristin},
year = {2020},
month = {02},
pages = {2975},
title = {Challenges and Opportunities with Causal Discovery Algorithms: application to {A}lzheimer’s Pathophysiology},
volume = {10},
journal = {Scientific Reports},
doiopt = {10.1038/s41598-020-59669-x}
}

@article{DirectLiNGAM,
  title={{DirectLiNGAM}: A direct method for learning a linear non-{G}aussian structural equation model},
  author={Shimizu, Shohei and Inazumi, Takanori and Sogawa, Yasuhiro and Hyv\"{a}rinen, Aapo and Kawahara, Yoshinobu and Washio, Takashi and Hoyer, Patrik O. and Bollen, Kenneth},
  journal={Journal of Machine Learning Research},
  volume={12},
  pages={1225--1248},
  year={2011}
}

@book{Jaynes2003,
  author = {Edwin T. Jaynes},
  title = {Probability Theory: The Logic of Science},
  year = {2003},
  address = {Cambridge, MA},
  publisher = {Cambridge University Press},
}

@book{cover,
  author = {Thomes Cover and Joy Thomas},
  title = {Elements of Information Theory},
  year = {1991},
  publisher = {Wileys Series in Telecommunications},
  address = {New York},
}

@article{Gruenwald2004,
  author = {Peter Grünwald and Philip Dawid},
  title = {Game theory, maximum entropy, minimum discrepancy and robust bayesian decision theory},
  year = {2004},
  journal = {The Annals of Statistics},
  volume = {32},
  number = {4},
  pages = {1367-433},
}

@book{Lauritzen,
  author = {Steffen Lauritzen},
  title = {Graphical Models},
  year = {1996},
  publisher = {Clarendon Press}}

@misc{Schkoda2024,
      title={Causal Discovery of Linear Non-{G}aussian Causal Models with Unobserved Confounding}, 
      author={Daniela Schkoda and Elina Robeva and Mathias Drton},
      year={2024},
      eprint={2408.04907},
      archivePrefix={arXiv},
      primaryClass={stat.ML},
      note={arXiv preprint}, 
}

@article{Salehkaleybar2020,
    AUTHOR = {Salehkaleybar, Saber and Ghassami, AmirEmad and Kiyavash,
              Negar and Zhang, Kun},
     TITLE = {Learning linear non-{G}aussian causal models in the presence
              of latent variables},
   JOURNAL = {Journal of Machine Learning Research},
  FJOURNAL = {Journal of Machine Learning Research (JMLR)},
      YEAR = {2020},
  volume  = {21},
  number  = {39},
  pages   = {1--24},
      issnopt = {1532-4435,1533-7928},
   MRCLASS = {62D20},
  MRNUMBER = {4073772},
}

@inproceedings{Reisach2021,
  booktitle = {Advances in Neural Information Processing Systems},
 editorppt = {M. Ranzato and A. Beygelzimer and Y. Dauphin and P.S. Liang and J. Wortman Vaughan},
 pages = {27772--27784},
 publisher = {Curran Associates, Inc.},
 title = {Beware of the Simulated {DAG}! {C}ausal Discovery Benchmarks May Be Easy to Game},
 volume = {34},
author = {Alexander G. Reisach and Christof Seiler and Sebastian Weichwald},
 year = {2021}
}

@inproceedings{Jaber2020,
  author = {Amin Jaber and Murat Kocaoglu and Karthikeyan Shanmugam and Elias Bareinboim},
  title = {Causal discovery from soft interventions with unknown targets: characterization and learning},
 publisher = {Curran Associates, Inc.},
 pages = {9551--9561},
  year = {2020},
  booktitle = {Advances in Neural Information Processing Systems},
 volume = {33},
}

@inproceedings{Hoyer,
   author = {Hoyer, Patrik and Janzing, Dominik and Mooij, Joris M. and Peters, Jonas and Sch\"{o}lkopf, Bernhard},
 booktitle = {Advances in Neural Information Processing Systems},
 editoropt = {D. Koller and D. Schuurmans and Y. Bengio and L. Bottou},
 pages = {},
 publisher = {Curran Associates, Inc.},
 title = {Nonlinear causal discovery with additive noise models},
 volume = {21},
 year = {2008},
}

@inproceedings{zhang2023identifiability,
  author = {Zhang, Jiaqi and Greenewald, Kristjan and Squires, Chandler and Srivastava, Akash and Shanmugam, Karthikeyan and Uhler, Caroline},
 booktitle = {Advances in Neural Information Processing Systems},
 editoropt = {A. Oh and T. Naumann and A. Globerson and K. Saenko and M. Hardt and S. Levine},
 pages = {50254--50292},
 publisher = {Curran Associates, Inc.},
 title = {Identifiability Guarantees for Causal Disentanglement from Soft Interventions},
 volume = {36},
 year = {2023}}

@inproceedings{Roland2022,
  author = {Paul Rolland and Volkan Cevher and Matth{\"a}us Kleindessner and Chris Russell and Dominik Janzing and Bernhard Sch{\"o}lkopf and Francesco Locatello},
  booktitle = 	 {Proc. 39th International Conference on Machine Learning},
  title = {Score matching enables causal discovery of nonlinear additive noise models},
  volume = 	 {162},
  year = {2022},
  publisher = {PMLR},
}

@inproceedings{Lopez2015,
author = {Lopez-Paz, David and Muandet, Krikamol and Sch\"{o}lkopf, Bernhard and Tolstikhin, Ilya},
title = {Towards a learning theory of cause-effect inference},
year = {2015},
publisher = {PMLR},
booktitle = {Proc. 32nd International Conference on International Conference on Machine Learning},
pages = {1452--1461},
numpages = {10},
volume = {37}
}

@inproceedings{anticausal,
author = {Sch\"{o}lkopf, Bernhard and Janzing, Dominik and Peters, Jonas and Sgouritsa, Eleni and Zhang, Kun and Mooij, Joris M.},
title = {On causal and anticausal learning},
year = {2012},
publisher = {Omnipress},
booktitle = {Proc. 29th International Coference on International Conference on Machine Learning},
pages = {459–466},
numpages = {8},
locationopt = {Edinburgh, Scotland},
seriesopt = {ICML'12}
}

@InProceedings{zheng20a,
  title = 	 {Learning Sparse Nonparametric {DAG}s},
  author =       {Zheng, Xun and Dan, Chen and Aragam, Bryon and Ravikumar, Pradeep and Xing, Eric},
  booktitle = 	 {Proc. 23rd International Conference on Artificial Intelligence and Statistics},
  pages = 	 {3414--3425},
  year = 	 {2020},
  editoropt = 	 {Chiappa, Silvia and Calandra, Roberto},
  volume = 	 {108},
  publisher =    {PMLR}
}

@inproceedings{Biza2020,
  author = {Konstantina Biza and Ioannis Tsamardinos and Sofia Triantafillou},
  title = {Tuning causal discovery algorithms},
  year = {2020},
  pages = {23--25},
  booktitle = {Proc. 10th International Conference on Probabilistic Graphical Models, volume 138},
  editoropt = {Manfred Jaeger and Thomas Dyhre Nielsen},
  publisher = {PMLR}
}

@inproceedings{Montagna2022,
  title = 	 {Scalable Causal Discovery with Score Matching},
  author =       {Montagna, Francesco and Noceti, Nicoletta and Rosasco, Lorenzo and Zhang, Kun and Locatello, Francesco},
  booktitle = 	 {Proc. 2nd Conference on Causal Learning and Reasoning},
  pages = 	 {752--771},
  year = 	 {2023},
  editoropt = 	 {van der Schaar, Mihaela and Zhang, Cheng and Janzing, Dominik},
  volume = 	 {213},
  publisher =    {PMLR},

}

@InProceedings{Mejia2022,
  title = 	 { Obtaining Causal Information by Merging Datasets with {MAXENT} },
  author =       {Garrido Mejia, Sergio H. and Kirschbaum, Elke and Janzing, Dominik},
  booktitle = 	 {Proc. 25th International Conference on Artificial Intelligence and Statistics},
  pages = 	 {581--603},
  year = 	 {2022},
  editoropt = 	 {Camps-Valls, Gustau and Ruiz, Francisco J. R. and Valera, Isabel},
  volume = 	 {151},
  seriesopt = 	 {Proceedings of Machine Learning Research},
  monthopt = 	 {28--30 Mar},
  publisher =    {PMLR},
}

@inproceedings{maeda2020rcd,
  author = {Takashi Nicholas Maeda and Shohei Shimizu},
  title = {{RCD}: repetitive causal discovery of linear non-{G}aussian acyclic models with latent confounders},
  year = {2020},
  pages = {735-745},
  booktitle = {International Conference on Artificial Intelligence and Statistics},
  publisher = {PMLR},
}

@inproceedings{Tian2001CausalDF,
  author = {Jin Tian and Judea Pearl},
  title = {Causal discovery from changes},
  year = {2001},
publisher = {AUAI Press},
booktitle = {Proc. 17th Conference on Uncertainty in Artificial Intelligence},
}

@inproceedings{UAI_identifiability,
  author = {Peters, Jonas and Mooij, Joris M. and Janzing, Dominik and Sch\"{o}lkopf, Bernhard},
  title = {Identifiability of causal graphs using functional models},
  year = {2011},
publisher = {AUAI Press},
    pages = {589–598},
  booktitle = {Proc. 27th Conference on Uncertainty in Artificial Intelligence},
}

@inproceedings{Zhang_UAI,
  title={On the Identifiability of the Post-Nonlinear Causal Model},
  author={Kun Zhang and Aapo Hyv{\"a}rinen},
  booktitle={Proc. 25th Conference on Uncertainty in Artificial Intelligence},
  year={2009}
}

@inproceedings{Kano2003,
  author = {Yutaka Kano and Shohei Shimizu},
  title = {Causal inference using nonnormality},
  year = {2003},
  pages = {261-270},
  booktitle = {Proc. International Symposium on Science of Modeling, the 30th Anniversary of the Information Criterion},
  publisheropt = {Tokyo},
  addressopt = {Japan},
}

@inproceedings{SunLauderdale,
  author =     {Xiaohai Sun and Dominik Janzing and Bernhard Sch\"{o}lkopf},
  title =      "{Causal inference by choosing graphs with most plausible Markov kernels}",
  booktitle =  {Proc. 9th International Symposium on Artificial Intelligence and Mathematics},
  year = 			 {2006},
  addressopt =		 {Fort Lauderdale, FL},
  pages =			 {1--11},
}

@inproceedings{Kocaoglu2017,
  author = {Murat Kocaoglu and Alexandros G. Dimakis and Sriram Vishwanath and Babak Hassibi},
  title = {Entropic causal inference},
  year = {2017},
  pages = {1156-1162},
  booktitle = {Proc. 31st AAAI Conference on Artificial Intelligence},
  publisher = {AAAI Press},
}

@inproceedings{Dhir2019,
  title={Integrating overlapping datasets using bivariate causal discovery},
  author={Anish Dhir and Ciar{\'a}n M. Lee},
  booktitle={Proc. 34th AAAI Conference on Artificial Intelligence},
  year={2019},
  publisher = {AAAI Press},
}

@inproceedings{ke2022learning,
  author = {Nan Rosemary Ke and Silvia Chiappa and Jane X. Wang and Jorg Bornschein and Anirudh Goyal and Melanie Rey and Matthew Botvinick and Theophane Weber and Michael Curtis Mozer and Danilo Jimenez Rezende},
  title = {Learning to induce causal structure},
  year = {2023},
  booktitle = {International Conference on Learning Representations},
}

@inproceedings{Zhang2017CausalDF,
  author = {Kun Zhang and Biwei Huang and Jiji Zhang and Clark Glymour and Bernhard Schölkopf},
  title = {Causal discovery from nonstationary/heterogeneous data: Skeleton estimation and orientation determination},
  year = {2017},
  pages = {1347-1353},
  booktitle = {Proc. 26th International Joint Conference on Artificial Intelligence},
}

@inproceedings{lachapelle2020,
  author = {Sébastien Lachapelle and Philippe Brouillard and Tristan Deleu and Simon Lacoste-Julien},
  title = {Gradient-based neural {DAG} learning},
  year = {2020},
  booktitle = {International Conference on Learning Representations},
}

\appendix

\section{LOVO predictors for DAGs with two arrows \label{subsec:12dags}}
As mentioned earlier, the case of three nodes is particularly challenging, and our approach presented in Section \ref{sec:lovofromdags} may not always succeed. Therefore, we present alternative LOVO predictors that can be beneficial in these cases. Specifically, we consider the "promise"-scenario of  
three variable $(X, Y, Z)$, where
we are given the information that the joint distribution $P(X,Y,Z)$ has been generated by a causal directed
acyclic graph (DAG) with {\it two arrows} only. In Table \ref{tab:dags}, we group the $12$ possible DAGs 
according to the $3$ possible 
skeletons $X -Z - Y$, $X - Y - Z$, $Y - X - Z$, with each skeleton allowing for $4$ different DAGs. 
 We will see that in each of these groups, the collider is special, but the three other Markov equivalent DAGs entail the same LOVO predictor.

\paragraph{No. 1-3: DAGs with $X\independent Y\,|Z$ \label{subsec:maxent}}
This is the simplest case where the conditional independence directly entails 
the solution
\begin{equation}\label{eq:chain}
P(X,Z,Y) = P(X,Z)P(Y|Z),
\end{equation} 
without any parametric assumptions. 
The solution is most intuitive for the DAGs  $X\rightarrow Z \rightarrow Y$  (No.1)
and  $X \leftarrow Z \rightarrow Y$ (No.2), where the algebraic structure of \eqref{eq:chain}
resembles the data generating process by applying the stochastic map $P(Y|Z)$ to the joint distribution of $X,Z$. 
While \ref{eq:chain} is certainly also valid for $X \leftarrow Z \leftarrow Y$, now $P(Y|Z)$ turns into an "anticausal" \citep{anticausal} 
conditional. When parametric assumptions are imposed for causal conditionals (e.g. linear non-Gaussian models \citep{Kano2003} or non-linear additive noise models \citep{Hoyer}), $P(Y|Z)$ now results from Bayesian inversion of those models.  
For linear models, the Pearson correlation between $X$ and $Y$ is easily obtained via\footnote{This follows from zero partial correlation, which is defined by $\rho_{X,Y|Z} = \frac{\rho_{XY} - \rho_{X,Z}\rho_{Z,Y}}{\sqrt{1- \rho_{X,Z}^2 }   \sqrt{1- \rho_{Y,Z}^2 }}$.}
\begin{equation}\label{eq:corrmaxent} 
\rho_{XY} = \rho_{XZ} \cdot \rho_{ZY}.
\end{equation} 
If $X,Y$ have zero mean and unit variance, the best linear predictor for $Y$ from $X$ then reads
$\Exp[Y|X=x] = \rho_{XY}\cdot x$. 
By slight abuse of terminology, we therefore call $\rho_{XY}$ the "linear LOVO predictor," which implicitly refers to this convention.

Note that this LOVO predictor coincides with the MaxEnt baseline predictor, and therefore, the cases where \eqref{eq:corrmaxent} does {\it not} hold are the interesting ones for us. 

\paragraph{No. 4: variable $Z$ as collider}
Due to $X\independent Y$, we ignore $X$ and take $P(Y)$ as the correct LOVO predictor for $Y$. We will later see, however, that this case is hard to recognize from the bivariate distributions because 
the bivariate causal models $X\rightarrow Z$ and $Y\rightarrow Z$ can also originate from the joint models
$X\rightarrow Y \rightarrow Z$ and $Y\rightarrow X \rightarrow Z$.

\begin{table}
\begin{tabular}{|r|l|l|l|l|l|}
 \hline 
No. &DAG & linear predictor & stochastic matrix & bivariate   & necessary\\
 & & & predictor & causality&  conditions\\
 \hline 
1& $X \rightarrow  Z \rightarrow Y$ &$\rho_{XY} = \rho_{XZ} \cdot \rho_{YZ}$ & $P_{Y|X} = P_{Y|Z} P_{Z|X} $ &  $X\rightarrow  Z$  & \label{eq:chainXZY}   \\
 &&&& $Y \leftarrow Z$ &\\
 \hline
2 & $X \leftarrow Z \rightarrow Y$   &  $\rho_{XY} = \rho_{XZ} \cdot \rho_{YZ}$   &   $P_{Y|X} = P_{Y|Z} P_{Z|X} $  & $X\leftarrow  Z$ & \\ 
  & &&& $Y \leftarrow Z$  & \\
 \hline
3 &$ X \leftarrow Z \leftarrow Y$  &   $\rho_{XY} = \rho_{XZ} \cdot \rho_{YZ}$  &  $P_{Y|X} = P_{Y|Z} P_{Z|X} $ & $X\leftarrow  Z$ & \\
&&& &$Y \rightarrow Z$ &  \\
\hline
4 & $X \rightarrow Z \leftarrow Y$   & $\rho_{XY}=0$ & $P_{Y|X} = P_Y$ & $X\rightarrow  Z$ & $\rho^2_{XZ} + \rho_{YZ}^2 \leq 1$ \\
&&&& $Y \rightarrow Z$ &   \\
\hline
\hline
5 & $X \rightarrow Y \rightarrow Z$  &   $\rho_{XY} = \rho_{XZ} / \rho_{YZ}$  &  $P_{Y|X} = P_{Z|Y}^{-1} P_{Z|X}$ &  $X\rightarrow  Z$  & $I(X:Z) $ \\ 
&&& & $Y \rightarrow Z$  &  $ \leq I(Y:Z)$ \\
\hline
6 & $X \leftarrow Y \rightarrow Z$  & $\rho_{XY} = \rho_{XZ} / \rho_{YZ}$  &  $P_{Y|X} = P_{Z|Y}^{-1} P_{Z|X}$ & $X\leftrightarrow  Z$ & $I(X:Z)  $ \\
&&&& $Y \rightarrow Z$ & $\leq I(Y:Z)$ \\
\hline
7 & $X \leftarrow Y \leftarrow Z$ & $\rho_{XY} = \rho_{XZ} / \rho_{YZ}$ &   $P_{Y|X} = P_{Z|Y}^{-1} P_{Z|X}$   & $X\leftarrow  Z$ &  $I(X:Z) $ \\
&&&& $Y \leftarrow Z$ & $ \leq I(Y:Z)$ \\
\hline
8 & $X \rightarrow Y \leftarrow Z$  & ? & ?  & $X \centernot{-} Z$  & \\
&&& &  $Y \leftarrow Z$  & \\
\hline
\hline
9 & $Y \rightarrow X \rightarrow Z$ & $\rho_{XY} = \rho_{YZ} / \rho_{XZ}$ & $P_{Y|X} =  P_{Y|Z} P_{X|Z}^{-1}$  & $X\rightarrow  Z$ & $I(X:Z)  $ \\
&&&& $Y \rightarrow Z$  & $\geq I(Y:Z)$  \\
\hline
10 & $Y \leftarrow X \rightarrow Z$ & $\rho_{XY} = \rho_{YZ} / \rho_{XZ}$ & $P_{Y|X} =  P_{Y|Z} P_{X|Z}^{-1}$ & $X\rightarrow  Z$ & $I(X:Z)  $ \\
 &&&& $Y \leftrightarrow Z$ & $\geq I(Y:Z)$  \\
 \hline
11 & $Y \leftarrow X \leftarrow Z$ &  $\rho_{XY} = \rho_{YZ} / \rho_{XZ}$ & $P_{Y|X} =  P_{Y|Z} P_{X|Z}^{-1}$ & $X\leftarrow  Z$ &   $I(X:Z)  $ \\
&&&& $Y \leftarrow Z$ &  $\geq I(Y:Z)$  \\
\hline
12 & $Y \rightarrow X \leftarrow Z$ & ? & ?   &$X\leftarrow  Z$ & \\
&&&&  $Y  \centernot{-} Z$ & \\
\hline
\end{tabular}
\caption{\label{tab:dags} All possible DAGs on $X,Y,Z$ with two arrows, together with their LOVO predictors.}   
\end{table}

\paragraph{No. 5-7: DAGs with $X \independent Z\,|Y$}
Now, the conditional distribution of $X$ given $Z$ is a concatenation of Markov kernels
\begin{equation}\label{eq:ymediator} 
P(Z|X) =P(Z|Y) \cdot P(Y|X).
\end{equation} 
In linear models we conclude $\rho_{XY} \cdot \rho_{YZ} = \rho_{XZ}$, from which we can directly construct
the linear LOVO predictor. 

In the case where $X,Y,Z$ are variables with finite ranges $\cX,\cY,\cZ$,  we introduce the stochastic matrices 
$P_{X|Z} :=(p(x|z))_{x\in \cX,z\in \cZ}$ and obtain the matrix equation 
$P_{Z|X} = P_{Z|Y} \cdot P_{Y|X}$. 
Whenever the matrix $P_{Z|Y}$ is invertible\footnote{Note that the inverse is not a stochastic matrix except for the trivial case of determinism.}, we thus obtain
\begin{equation}\label{eq:Ymedi}
P^{\rm mediator \,\,Y}_{Y|X} := P_{Z|Y}^{-1} \cdot P_{Z|X}.
\end{equation} 

\paragraph{No. 8: variable $Y$ as collider}
This is a negative case: as explained in Section \ref{subsec:3_variables}, it is unclear how to construct a good LOVO predictor. 

\paragraph{No. 9-11: DAGs with $Y \independent Z\,|X$} 
Here we obtain
\begin{equation}\label{eq:xmediator} 
P_{Y|Z} = P_{Y|X} P_{X|Z},
\end{equation}  
which implies the predictor 
\begin{equation}\label{eq:Xmedi} 
P^{\rm mediator \,\,X}_{Y|X} :=  P_{Y|Z} P_{X|Z}^{-1},
\end{equation} 
if $P_{X|Z}$ is invertible. 

\paragraph{No. 12: variable $X$ as collider} 
As for No. 8, we can not infer the strength of the influence of $Y$ on $X$ and abstain from constructing a LOVO predictor. 

Inspecting the column with the bivariate causal graphs in Table \ref{tab:dags}, we find only $7$ cases where the joint DAG can be uniquely identified from the bivariate graphs, namely the numbers No. 1, 2, 3, 6, 8, 10, 12.
Unfortunately, recognizing 8 and 12 is not helpful for our purpose because we can not offer a LOVO predictor there. Moreover, the predictor of No. 1-3 coincides with our baseline. Overall, we are left with only two cases where a LOVO predictor is realizable and simultaneously beats the baseline. We can improve upon that by considering additional conditions on the marginals from which the respective DAG can be excluded, as listed in the last column of the table.

For instance, $X\independent Z \, |Y$ implies that the dependence between $X$ and $Z$  can not be larger than the dependence between $Y$ and $Z$. In the non-parametric case, this can be formalized via
the Shannon mutual information, for which we have the data processing inequality $I(X:Z)\leq I(Y:Z)$.
For linear models, we have $\rho_{XZ} \leq \rho_{YZ}$. Further, the collider $X\rightarrow Z \leftarrow Y$ 
is only possible if $\rho_{XZ}^2 + \rho_{YZ}^1 \leq 1$, otherwise the correlation matrix
\[
\left(\begin{array}{ccc} 1 & \rho_{XZ} & 0 \\
\rho_{XZ} & 1 & \rho_{YZ} \\
\rho_{YZ} & 0 & 1\end{array} \right),  
\]  
would not be positive semi-definite.

\new{\section{LOVO extended to further graph types}\label{sec:LOVO_further_graph_types}
Three graph types, which have so far not been considered and yet are the outputs of many causal discovery algorithms, are ADMGs without confounded causal links, Directed Acyclic Graphs (DAGs), Completed Partially Directed Acyclic Graphs (CPDAGs) and Partial Ancestral Graphs (PAGs). This section extends parent adjustment to these graph types. As before, we identify unlinked pairs and then find valid adjustment sets. To highlight the differences between the graph types, we revisit the example used in Subsection \ref{sec:lovofromdags}, but this time with the edge $W_1\to W_3$ omitted, resulting in the graph shown in Figure \ref{fig:example_lemma_7}. For this slightly modified graph, Lemma \ref{lem:nolinks_ADMG} identifies all unlinked pairs except $(W_3,W_5)$ and Lemma \ref{lem:determine_edge_type_DAG} finds all six unlinked pairs. When referring to this example, we may use the shorthand $i$ to denote the node $W_i$.
}
\subsection{ADMGs without confounded causal links}\label{subsec:LOVO_ADMGs_without_confounded_links}
The alternative ADMG definition prohibiting confounded causal links uses single bidirected edges in place of our confounded causal links $\stackrel{\leftrightarrow }{\rightarrow}$ or $\stackrel{\leftrightarrow }{\leftarrow}$. 
This substitution leads to a loss of information. For instance, when leaving out $X$ in each of the following three graphs,
\begin{center}
    \begin{tikzpicture}[baseline=(X.base)]
    \node (X) at (1, 0) {$X$};
    \node (Y) at (0, 0) {$Y$};
    \node (Z) at (2, 0) {$Z$};
    \draw[->] (X) -- (Y);
    \draw[->] (X) -- (Z);
\end{tikzpicture}, \hspace{0.5cm}
    \begin{tikzpicture}[baseline=(X.base)]
    \node (X) at (1, 0) {$X$};
    \node (Y) at (0, 0) {$Y$};
    \node (Z) at (2, 0) {$Z$};
    \draw[->] (X) -- (Y);
    \draw[->] (X) -- (Z);
    \draw[->, bend left=20] (Y) to (Z);
\end{tikzpicture},\hspace{0.5cm}and
\begin{tikzpicture}[baseline=(X.base)]
    \node (X) at (1, 0) {$X$};
    \node (Y) at (0, 0) {$Y$};
    \node (Z) at (2, 0) {$Z$,};
    \draw[->] (X) -- (Y);
    \draw[->] (X) -- (Z);
    \draw[->, bend left=20] (Z) to (Y);
\end{tikzpicture}
\end{center}
 according to the ADMG definition used so far, one obtains three distinct marginal graphs, namely
$$Y \leftrightarrow Z, \hspace{0.5cm}Y\stackrel{\leftrightarrow }{\rightarrow}Z, \hspace{0.5cm}\text{and }Y\stackrel{\leftrightarrow }{\leftarrow}Z.$$ In contrast, the other definition always yields $Y \leftrightarrow Z$. 
Despite these differences, Lemma \ref{lem:nolinks_ADMG} remains valid. Recall that it asserts that whenever $X$ has a child in $G_X$ that is not a sibling of $Y$ in $G_Y$, then $X$ and $Y$ are not linked in $G$. We write $G_X, G_Y$ for the ADMGs with the previously used definition, and $\overline{G}_X, \overline{G}_Y$ for the ADMGs where confounded causal links are substituted for bidirected edges. Because every child of $X$ in $\overline{G}_X$  is also a child of $X$ in $G_X$, and every sibling of $Y$ in $G_Y$ is also a sibling of $Y$ in $\overline{G}_Y$, lemma remains true when considering $\overline{G}_X$ and $\overline{G}_Y$.
Similarly, one can show that  Lemma \ref{lem:determine_edge_type_DAG} still holds. 
However, the next step, that is, reading off the parents of $X, Y$  from the marginal graphs, becomes more involved. Specifically,  a bidirected edge between $Y \leftrightarrow Z_i \in \overline{G}_Y$ precludes their parent-child relationship: In $G$, we could have $Y \to Z_i, Y \leftarrow Z_i,$ or $Y \centernot{-}Z_i$. So, each sibling of $Y$ in $\overline{G}_Y$ could be a parent in $G$ or not, and therefore, we can not infer the parents whenever $Y$ has siblings in $\overline{G}_Y$. %
Accordingly, whenever $\sib^{\overline{G}_X}(X)\neq \varnothing$ or $\sib^{\overline{G}_Y}(Y)\neq \varnothing$, we refrain from a prediction for all pairs with siblings. Note that when Lemma \ref{lem:determine_edge_type_DAG} is employed, we anyways exclude all these pairs since an edge might exist according to condition (1). However, this does not necessarily apply when Lemma \ref{lem:nolinks_ADMG} is used. As stated above, 
in the running example,  Lemma \ref{lem:nolinks_ADMG} detects five pairs as unlinked. For all of them, $X$ has no siblings in $\overline{G}_X$ and likewise $Y$ none in $\overline{G}_Y$, so we do not need to abstain.
\new{
\subsection{Directed Acyclic Graphs}\label{subsec:DAGs}
Turning to DAGs, it is evident that they are not able to model latent confounding, which, however, is typically present in the marginal graphs $G_X$ and $G_Y$. To address this mismatch, we assume to be given DAGs $\tilde{G}_X, \tilde{G}_Y$, which are derived from $G_X, G_Y$ by orienting each bidirected edge into a directed one. Often, only one direction avoids the creation of cycles. Otherwise, we opt for a random direction. We are aware that it might be unrealistic to expect an algorithm assuming causal sufficiency to generate precisely such a DAG when latent confounding is actually present. Nonetheless, we argue that this assumption is still reasonable since we do not necessitate the LOVO predictor based on this assumption to yield a perfect prediction but only require it to outperform the non-causal predictor based on a random adjustment set.
\begin{figure}
    \centering
    $G$ = \begin{tikzpicture}[baseline={(0,.6)}]
    \node (3) at (-1.4, 0) {$W_3$};
    \node (5) at (1.4, 0) {$W_5$};
    \node (2) at  (0,1.2) {$W_2$};
    \node (4) at (0,0){$W_4$};
    \node (1) at (-1.4,1.2) {$W_1$};
    \draw[->, orange, bend left=60] (1) to (5);
    \draw[->] (1) -- (4);
    \draw[->] (2) -- (3);
    \draw[->] (4) -- (5);
    \draw[->] (2) -- (4);
\end{tikzpicture} \hspace{0.3cm}
    $G_{W_1}$ = \begin{tikzpicture}[baseline={(0,.6)}]
    \node (3) at (-1.4, 0) {$W_3$};
    \node (2) at  (0,1.2) {$W_2$};
    \node (4) at (0,0){$W_4$};
    \node[white] (5) at (1.4, 0) {$W_5$};
    \node (1) at (-1.4,1.2) {$W_1$};
    \draw[->] (1) -- (4);
    \draw[->] (2) -- (3);
    \draw[->] (2) -- (4);
\end{tikzpicture} \hspace{0.3cm}
   $G_{{W_5}}$ =  
   \begin{tikzpicture}[baseline={(0,.6)}]
    \node (3) at (-1.4, 0) {$W_3$};
    \node (5) at (1.4, 0) {$W_5$};
    \node (2) at  (0,1.2) {$W_2$};
    \node (4) at (0,0){$W_4$};
    \draw[->] (2) -- (3);
    \draw[<-> ,orange, transform canvas={yshift=2pt}] (4) -- (5);
    \draw[->,transform canvas={yshift=-2pt}] (4) -- (5);
    \draw[->] (2) -- (4);
\end{tikzpicture}\\[.3cm]
$G$ = 
\begin{tikzpicture}[baseline={(0,.6)}]
    \node (3) at (-1.4, 0) {$W_3$};
    \node (5) at (1.4, 0) {$W_5$};
    \node (2) at  (0,1.2) {$W_2$};
    \node (4) at (0,0){$W_4$};
    \node (1) at (-1.4,1.2) {$W_1$};
    \draw[->, matplotlib_blue, transform canvas={yshift=2pt}] (1) to (2);
    \draw[->, yellow, transform canvas={yshift=-2pt}] (2) to (1);
    \draw[->] (1) -- (4);
    \draw[->] (2) -- (3);
    \draw[->] (4) -- (5);
    \draw[->] (2) -- (4);
\end{tikzpicture} \hspace{0.3cm}
    $G_{{W_1}}$ =\begin{tikzpicture}[baseline={(0,.6)}]
    \node (3) at (-1.4, 0) {$W_3$};
    \node (5) at (1.4, 0) {$W_5$};
    \node (4) at (0,0) {$W_4$};
    \node (1) at (-1.4,1.2) {$W_1$};
    \draw[->, matplotlib_blue,transform canvas={xshift=2pt}] (1) to (3);
    \draw[->,transform canvas={yshift=-1pt,xshift=-1pt}] (1) -- (4);
    \draw[<->, yellow,transform canvas={xshift=-2pt}] (1) -- (3);
    \draw[<->, yellow,transform canvas={yshift=1pt,xshift=1pt}] (1) -- (4);
    \draw[->] (4) -- (5);
    \draw[<->] (3) -- (4);
\end{tikzpicture} \hspace{0.3cm}
   $G_{{W_2}}$ =  
\begin{tikzpicture}[baseline={(0,.6)}]
    \node (3) at (-1.4, 0) {$W_3$};
    \node (5) at (1.4, 0) {$W_5$};
    \node (2) at  (0,1.2) {$W_2$};
    \node (4) at (0,0){$W_4$};
    \draw[->] (2) -- (3);
    \draw[->] (4) -- (5);
    \draw[->,transform canvas={xshift=2pt}] (2) -- (4);
    \draw[<->,matplotlib_blue,transform canvas={xshift=-2pt}] (2) -- (4);
\end{tikzpicture}
\caption{\color{new}Adding the edge $1\to 5$ to $G$, introduces a bidirected edge in $G_{W_5}$. However, this additional edge is not visible in $\tilde{G}_{W_5}$ since it turns into $4\to 5$, which was already present before. Thus, $(1,5)$ can not be confirmed as unlinked. In contrast, we can conclude $1\centernot{-}2$. Lemma \ref{lem:exclude_links_directed_part} states that if $2$ and $1$ were connected in $G$, then $1$ would be linked to $3$ in $G_{{W_1}}$ since $3$ is a child of $2$ in $G_{W_2}$. Indeed, adding either $1\to 2$ or $2\to 1$ would yield a connection between $1$ and $3$ as visualized by the blue and yellow edges.}\label{fig:example_lemma_7}
    \end{figure}

Within this scenario, inferring the edge type is possible in fewer cases. Figure \ref{fig:example_lemma_7} illustrates why in the example, the pair $(1,5)$ can not be certified as unlinked anymore, but $(1,2)$ still can.

\begin{lemma}[excluding links based on DAGs]\label{lem:exclude_links_directed_part}
Let $G$ be a DAG, and $\tilde{G}_X, \tilde{G}_Y$ DAGs obtained from $G_X, G_Y$ by exchanging each bidirected edge for a directed one. If $X$ has a child in $\tilde{G}_X$ that is not a neighbor of $Y$ in $\tilde{G}_Y$, then $G$ does not contain any edge $X -Y$. 
\end{lemma}
\begin{proof}
Since $X$ has a child $C$ in $\tilde{G}_X$, either   $X \leftrightarrow C$ or $X \to C$ in $G_X$. In the first case, $X \leftarrow Y \to C \in G$, and thus $Y \to C \in G_Y$ contrary to the assumption of the lemma. In the second case, if $G$ would feature the edge $X \to Y$, then combining both edges yields $Y \leftarrow X \to C$ and therefore $Y$ and $C$ are connected in $G_Y$, contrary to the assumption of the lemma. Similarly, if the reverse edge $Y \to X$ would be part of $G$, then $Y \leftarrow X \to C \in G$, which results in $Y\to C \in G_Y$ and again contradicts the assumption.
\end{proof}
In the example, the lemma detects four of the six unlinked pairs, namely $(1, 2), (1, 3), (2, 5), $ and $ (3, 4)$.

\subsection{Partially Directed Acyclic Graphs and Partial Ancestral Graphs}\label{subsec:CPDAGs_PAGs}
\begin{figure}
    \centering
$G$ = 
\begin{tikzpicture}[baseline={(0,.6)}]
    \node (3) at (-1.4, 0) {$W_3$};
    \node (5) at (1.4, 0) {$W_5$};
    \node (2) at  (0,1.2) {$W_2$};
    \node (4) at (0,0){$W_4$};
    \node (1) at (-1.4,1.2) {$W_1$};
    \draw[->, matplotlib_blue] (1) to (2);
    \draw[->] (1) -- (4);
    \draw[->] (2) -- (3);
    \draw[->] (4) -- (5);
    \draw[->] (2) -- (4);
\end{tikzpicture}\hspace{0.2cm}
   $\mathcal{C}(\tilde{G}_{W_1}$) =  \begin{tikzpicture}[baseline={(0,.6)}]
    \node (3) at (-1.4, 0) {$W_3$};
    \node (5) at (1.4, 0) {$W_5$};
    \node (4) at (0,0) {$W_4$};
    \node (1) at (-1.4,1.2) {$W_1$};
    \draw[-] (1) -- (4);
    \draw[-, matplotlib_blue] (1) -- (3);
    \draw[-] (4) -- (5);
    \draw[-] (3) -- (4);
\end{tikzpicture} \hspace{0.2cm}
    $\mathcal{C}(\tilde{G}_{W_2}$) = \begin{tikzpicture}[baseline={(0,.6)}]
    \node (3) at (-1.4, 0) {$W_3$};
    \node (5) at (1.4, 0) {$W_5$};
    \node (2) at  (0,1.2) {$W_2$};
    \node (4) at (0,0){$W_4$};
    \draw[-] (2) -- (3);
    \draw[-] (4) -- (5);
    \draw[-] (2) -- (4);
\end{tikzpicture}\\[.3cm]
$H$ = 
\begin{tikzpicture}[baseline={(0,.6)}]
    \node (3) at (-1.4, 0) {$W_3$};
    \node (5) at (1.4, 0) {$W_5$};
    \node (2) at  (0,1.2) {$W_2$};
    \node (4) at (0,0){$W_4$};
    \node (1) at (-1.4,1.2) {$W_1$};
    \draw[->] (1) to (2);
    \draw[->] (1) -- (4);
    \draw[<-] (2) -- (3);
    \draw[->] (4) -- (5);
    \draw[->] (2) -- (4);
\end{tikzpicture} \hspace{0.1cm}
   $\mathcal{C}(\tilde{H}_{W_1})$ =  \begin{tikzpicture}[baseline={(0,.6)}]
    \node (3) at (-1.4, 0) {$W_3$};
    \node (5) at (1.4, 0) {$W_5$};
    \node (4) at (0,0) {$W_4$};
    \node (1) at (-1.4,1.2) {$W_1$};
    \draw[-] (1) -- (4);
    \draw[-] (4) -- (5);
    \draw[-] (3) -- (4);
\end{tikzpicture}\hspace{0.1cm}
    $\mathcal{C}(\tilde{H}_{W_2})$ = \begin{tikzpicture}[baseline={(0,.6)}]
    \node (3) at (-1.4, 0) {$W_3$};
    \node (5) at (1.4, 0) {$W_5$};
    \node (2) at  (0,1.2) {$W_2$};
    \node (4) at (0,0){$W_4$};
    \draw[-] (2) -- (3);
    \draw[-] (4) -- (5);
    \draw[-] (2) -- (4);
\end{tikzpicture}
\caption{\color{new}Altough adding the edge $1\to 2$ to $G$ introduces changes in the marginal CPDAG $\mathcal{C}(\tilde{G}_{W_1})$, we can not identify this pair as unlinked. In contrast to the scenario in Figure \ref{fig:example_lemma_7}, we can not be certain that $3$ is a child of $2$ and do not find any child to utilize in Lemma \ref{lem:exclude_links_directed_part}. Indeed, there is another graph $H$, with the edge between $2$ and $3$ reversed, an edge $1\to 2$, and the same marginal CPDAGs as $G$.}\label{fig:example_cpdags_neg}
    \end{figure}
    \begin{figure}
        \centering
  $G$ = \begin{tikzpicture}[baseline={(0,.6)}]
    \node (3) at (-1.4, 0) {$W_3$};
    \node (5) at (1.4, 0) {$W_5$};
    \node (2) at  (0,1.2) {$W_2$};
    \node (4) at (0,0){$W_4$};
    \node (1) at (-1.4,1.2) {$W_1$};
    \draw[->, orange] (1) to (3);
    \draw[->] (1) -- (4);
    \draw[->] (2) -- (3);
    \draw[->] (4) -- (5);
    \draw[->] (2) -- (4);
\end{tikzpicture} \hspace{0.2cm}
   $\mathcal{C}(\tilde{G}_{W_1})$ =  
   \begin{tikzpicture}[baseline={(0,.6)}]
    \node[white] (3) at (-1.4, 0) {$W_3$};
    \node (2) at  (0,1.2) {$W_2$};
    \node (4) at (0,0){$W_4$};
    \node (5) at (1.4, 0) {$W_5$};
    \node (1) at (-1.4,1.2) {$W_1$};
    \draw[->] (1) -- (4);
    \draw[->] (4) -- (5);
    \draw[->] (2) -- (4);
\end{tikzpicture}\hspace{0.2cm}
    $\mathcal{C}(\tilde{G}_{W_3})$ = \begin{tikzpicture}[baseline={(0,.6)}]
    \node (3) at (-1.4, 0) {$W_3$};
    \node (5) at (1.4, 0) {$W_5$};
    \node (2) at  (0,1.2) {$W_2$};
    \node (4) at (0,0){$W_4$};
    \draw[-] (2) -- (3);
    \draw[- ,orange] (3) -- (4);
    \draw[-] (4) -- (5);
    \draw[-] (2) -- (4);
\end{tikzpicture} 
\caption{\color{new}Unlike the pair $(1,2)$, the pair 
 $(1,3)$ can be confirmed to not be connected. The graph $\mathcal{C}(\tilde{G}_{W_3})$ features directed edges, which allow to conclude that $4$ is definitively a child of $1$ in $\tilde{G}_{W_1}$, while not being a neighbor of $3$ in $\tilde{G}_{W_3}$. As before, the orange edges show that, if $1\to 3 \in G$, nodes $3$ and $4$ were neighbors in $\tilde{G}_{W_3}$.}\label{fig:example_cpdags_pos}
    \end{figure}

A CPDAG is an equivalence class of DAGs, which collects all DAGs entailing the same conditional independence statements. Consequently, CPDAGs also do not capture latent confounding. Because algorithms learning CPDAGs often rely on conditional independence testing, we again expect latent confounding to manifest as directed or undirected edge in the learned CPDAG. More precisely, we assume we are given the CDPAGs $\mathcal{C}(\tilde{G}_X)$ and $\mathcal{C}(\tilde{G}_Y)$ corresponding to $\tilde{G}_X$ and $\tilde{G}_Y$ as defined in the previous subsection. Since a CPDAG represents an equivalence class of DAGs, or in other words, a DAG except for uncertainty regarding some of the edge directions, Lemma \ref{lem:exclude_links_directed_part} remains valid. Unsurprisingly, we can exclude fewer pairs because edges might be unoriented and, therefore, we can less frequently find a node that is \textit{definitely} a child of  $X$ in $\tilde{G}_X$ and at the same time not a neighbor of $Y$ in $\tilde{G}_Y$. Figure \ref{fig:example_cpdags_neg} and \ref{fig:example_cpdags_pos} demonstrate this for our running example. Overall, the lemma can certify two pairs as unlinked, namely $(1, 3)$ and $ (3, 4)$.

\begin{figure}
    \centering
  $G$ = \begin{tikzpicture}[baseline={(0,.6)}]
    \node (3) at (-1.4, 0) {$W_3$};
    \node (5) at (1.4, 0) {$W_5$};
    \node (2) at  (0,1.2) {$W_2$};
    \node (4) at (0,0){$W_4$};
    \node (1) at (-1.4,1.2) {$W_1$};
    \draw[->, orange] (1) to (3);
    \draw[->] (1) -- (4);
    \draw[->] (2) -- (3);
    \draw[->] (4) -- (5);
    \draw[->] (2) -- (4);
\end{tikzpicture} \hspace{0.1cm}
   $\mathcal{P}(G_{W_1})$ =  
   \begin{tikzpicture}[baseline={(0,.6)}]
    \node[white] (3) at (-1.4, 0) {$W_3$};
    \node (2) at  (0,1.2) {$W_2$};
    \node (4) at (0,0){$W_4$};
    \node (5) at (1.4, 0) {$W_5$};
    \node (1) at (-1.4,1.2) {$W_1$};
    \draw[o->] (1) -- (4);
    \draw[->] (4) -- (5);
    \draw[o->] (2) -- (4);
\end{tikzpicture}\hspace{0.1cm}
    $\mathcal{P}(G_{W_3})$ = \begin{tikzpicture}[baseline={(0,.6)}]
    \node (3) at (-1.4, 0) {$W_3$};
    \node (5) at (1.4, 0) {$W_5$};
    \node (2) at  (0,1.2) {$W_2$};
    \node (4) at (0,0){$W_4$};
    \draw[-] (2) -- (3);
    \draw[- ,orange] (3) -- (4);
    \draw[-] (4) -- (5);
    \draw[-] (2) -- (4);
\end{tikzpicture} \\[.3cm]
$G$ = 
\begin{tikzpicture}[baseline={(0,.6)}]
    \node (3) at (-1.4, 0) {$W_3$};
    \node (5) at (1.4, 0) {$W_5$};
    \node (2) at  (0,1.2) {$W_2$};
    \node (4) at (0,0){$W_4$};
    \node (1) at (-1.4,1.2) {$W_1$};
    \draw[->, matplotlib_blue] (3) to (4);
    \draw[->] (1) -- (4);
    \draw[->] (2) -- (3);
    \draw[->] (4) -- (5);
    \draw[->] (2) -- (4);
\end{tikzpicture} \hspace{0.1cm}
   $\mathcal{P}(G_{W_3})$ =  \begin{tikzpicture}[baseline={(0,.6)}]
    \node (3) at (-1.4, 0) {$W_3$};
    \node (5) at (1.4, 0) {$W_5$};
    \node (2) at  (0,1.2) {$W_2$};
    \node (1) at (-1.4,1.2) {$W_1$};
    \draw[o->, bend left=20] (1) -- (5);
    \draw[o->] (2) -- (5);
    \draw[o->, matplotlib_blue] (3) -- (5);
    \draw[o-o] (3) -- (2);
\end{tikzpicture}\hspace{0.1cm}
    $\mathcal{P}(G_{W_4})$ =  
   \begin{tikzpicture}[baseline={(0,.6)}]
    \node[white] (3) at (-1.4, 0) {$W_3$};
    \node (2) at  (0,1.2) {$W_2$};
    \node (4) at (0,0){$W_4$};
    \node (5) at (1.4, 0) {$W_5$};
    \node (1) at (-1.4,1.2) {$W_1$};
    \draw[o->] (1) -- (4);
    \draw[->] (4) -- (5);
    \draw[o->] (2) -- (4);
\end{tikzpicture}
\caption{\color{new}The edge $1\to 3$ can not be excluded since neither $1$ nor $3$ has definite children in their marginal PAGs . In contrast, $3 \to 4$ can be excluded due to the child $5$ of $4$ in $\mathcal{P}(G_{W_4})$.}\label{fig:example_pags}
    \end{figure}
PAGs are similar to CPDAGs insofar as they both represent an equivalence class of graphs encoding the same conditional independence statements. The key difference is that PAGs can represent latent confounding. Specifically, PAGs feature directed edges ($X \to Y$) and bidirected edges ($X \leftrightarrow Y$) with the same meaning as in ADMGs. As in the alternative definition, confounded causal links are not allowed and are replaced by single bidirected edges. Additionally, circle edges (\begin{tikzpicture}[baseline=(X.base), line width=.7pt]
    \node (X) at (0, 0) {$X$};
    \node (Y) at (1.2, 0) {$Y$};
    \draw[o-o] (X) -- (Y);
\end{tikzpicture}, \begin{tikzpicture}[baseline=(X.base), >=To, line width=.7pt]
    \node (X) at (0, 0) {$X$};
    \node (Y) at (1, 0) {$Y$};
    \draw[o->] (X) -- (Y);
\end{tikzpicture}) can be present, where a circle at an endpoint indicates uncertainty about the presence of an arrowhead or tail, similar to the undirected edges in CPDAGs. Sometimes, undirected edges ($X - Y$) are also included to represent selection bias; however, we do not consider them here. Since the circles represent uncertainty, we can directly read off the \textit{definite} as well as the \textit{possible} children, parents, neighbors, and siblings of each node. Therefore,  Lemma \ref{lem:nolinks_ADMG} is still applicable if only the PAGs $\mathcal{P}(G_X), \mathcal{P}(G_Y)$  corresponding to $G_X, G_Y$ are available.\footnote{Note that these PAGs may not exist as explained above. In such cases, the lemma is not applicable.} Again, due to the uncertainty, we can find fewer unlinked pairs. In the running example, only the pair $(3,4)$ is found to be unlinked, as illustrated in  Figure \ref{fig:example_pags}.

For CPDAGs as well as PAGs, having found unlinked pairs, the next step is determining a valid adjustment set. With many edges being unoriented, it is often impossible to identify the parents of $X, Y$. Therefore, we rely on a different adjustment set: Assuming a causally sufficient joint model, no path between $X$ and $Y$ with more than two edges can consist solely of colliders. Thus, these paths can be blocked by including all interior nodes in the adjustment set. For paths of length two ($X - Z -Y $), a case distinction must be made: 
\begin{itemize}
    \item If $Z$ could be either a collider or a non-collider, we can not determine a valid adjustment set and must abstain.
    \item If $Z$  is definitely a non-collider, it is included in the adjustment set.
    \item If $Z$ is a definite collider and was not already added to the the adjustment set in any previous step, the adjustment set remains valid. Otherwise, we abstain.
\end{itemize}
If this procedure succeeds in finding a valid set, the LOVO predictor can be formed via the three-step predictor defined in Section \ref{sec:lovofromdags}.
}

\section{When is MaxEnt LOVO correct?}
The following simple criterion tells us when the MaxEnt predictor is right: 
\begin{lemma}[MaxEnt baseline]\label{lem:base}
Let $G$ be a causal DAG connecting $\bZ,X,Y$ and $G^m$ be the corresponding moral graph.\footnote{Following \cite{Lauritzen}, page 7, the moral graph corresponding to a DAG $G$ is the undirected graph that contains an edge $a - b$ if and only if $a$ and $b$ are directly connected in $G$ or if they have a common child.} 
If $G_m$ does not contain the edge $X-Y$, then the MaxEnt predictor is correct.
\end{lemma}

\proof{Due to the Markov condition for undirected graphs \citep{Lauritzen}, $X\independent Y\,|\bZ$ if there is no link $X-Y$ in $G^m$, which implies $P(y|x,\bz) = P(y|\bz)$. $\square$
}

If $\bZ$ consists of just one variable $Z$, the number of DAGs for which Lemma \ref{lem:base} holds can be counted as follows: obviously, it only holds for DAGs with less than $3$ arrows. For the one with $2$ arrows, the skeleton must read $X - Z - Y$. To ensure that $G^m$ does not contain $X-Y$, there can not be a collider at $Z$, thus only the Markov equivalence class  of $X\to Z \to Y$ is remaining (with $3$ elements). 
For the $6$ DAGs with one arrow, $X\independent Y\,|Z$ is always satisfied. Hence, we obtain $9$ DAGs for which
our MaxEnt LOVO is optimal, and the total number of DAGs with $3$ nodes reads $25$.

 \section{Proofs}
 \subsection{Proof of Lemma \ref{lem:needbias}} 
Define the conditional cumulative distribution functions  $F_X(x|\bz) := P( X \leq x |\bZ=\bz)$  and
 $F_Y(y|\bz) := P( Y \leq y |\bZ=\bz)$.
We then define structural equation models for $P(Y|\bZ=\bz)$ $P(Y|\bZ=\bz)$ with uniformly distributed noise variables:
$X= f_X(\bZ,N_X)$ and $Y=f_Y(\bZ,N_Y)$, where $f_X(\bz,N_X) = F^{-1}_X(N_X |\bz)$ and 
 $f_Y(\bz,N_Y) = F^{-1}_Y(N_Y |\bz)$. Whenever 
 we generate $\bz$-values with distribution $P(\bZ)$, we obtain the right marginal distributions $P(X,\bZ)$ and
 $P(Y,\bZ)$. 
 Note that this holds even for dependent noise with arbitrary  $P(N_X,N_Y)$ with the only constraint
 that their marginals need to be uniform (in other words, $P(N_X,N_Y)$ is a copula) since
 the dependences between $N_X$ and $N_Y$ do not affect the marginals. 
 When we choose $P(N_X,N_Y)=P(N_X)P(N_Y)$, we obtain the MaxEnt solution $P^{\rm MaxEnt}(X,Y,\bZ)=P(X,\bZ)P(Y|\bZ)$.
 However, when we choose $N_Y=N_X$, the variables $X$ and $Y$ are positively correlated when conditioned on $\bZ$.
 When we choose  $N_Y=(1-N_X)$ instead, $X$ and $Y$ are negatively correlated  when conditioned on $\bZ$.
 Let $\cov^{\rm MaxEnt}(X,Y)$, $\cov^{\rm pos}(X,Y)$, and $\cov^{\rm neg}(X,Y)$ denote the covariances of $X,Y$ with respect to
 the three different choices of the dependences of $N_X,N_Y$. We then have
 \[
 \cov^{\rm pos}(X,Y) > \cov^{\rm MaxEnt}(X,Y) > \cov^{\rm neg}(X,Y).
 \]
 This follows because 
 \[
 \cov^{\rm pos}(X,Y|\bZ=\bz) > \cov^{\rm MaxEnt}(X,Y|\bZ=\bz) > \cov^{\rm neg}(X,Y|\bZ=\bz), 
 \]
 holds for any $\bz$ from the law of total covariance:
 \[
 \cov(X,Y) = \Exp[\cov(X,Y|\bZ)] + \cov( \Exp[X|\bZ], \Exp[Y|\bZ] ),
 \] 
 since  the conditional expectations $\Exp[X|\bZ]$ the  $\Exp[Y|\bZ]$ are both functions of $\bZ$, which only depend on the respective marginal distribution and are therefore unaffected by the dependences of the
 noise variables.  $\square$

\subsection{Proof of Lemma \ref{lem:nolinks_ADMG}}
Going slightly beyond the statement in the lemma, we show the following criteria for excluding each possible edge type.
\begin{enumerate}[1)]
\item We can exclude the existence of a direct edge $X \to Y$ in $G$ if
\begin{enumerate}[a)]
\item $X$ occurs causally after $Y$ in the sense that in $G_X$ there exists an ancestor of $X$ that is at the same time a descendant of $Y$ in $G_Y$; or
\item at least one of the following implications is violated
\begin{enumerate}[i)]
    \item $P \to X \in G_X \implies P \to Y \in G_Y$,
    \item $X \to C \in G_X \implies Y \to C \in G_Y$ or $Y \leftrightarrow C \in G_Y$, 
    \item $X \leftrightarrow S \in G_X \implies Y \leftrightarrow S \in G_Y$, 
    \item $Y \to C \in G_Y \implies X \to C \in G_X.$
\end{enumerate}
\end{enumerate}
\item Likewise, the arrow $Y\rightarrow X$ can be excluded by swapping the roles of $X$ and $Y$. 
    \item The bidirected edge $X \leftrightarrow Y$ can be excluded if one of the following implications is violated
    \begin{enumerate}[a)]
        \item $X \to C \in G_X \implies Y \to C \in G_Y$ or $Y \leftrightarrow C \in G_Y$, 
        \item $Y \to C \in G_Y \implies X \to C$ or $X \leftrightarrow C \in G_X$.
    \end{enumerate}
\end{enumerate}
Since the criteria in point 3) already entail the criteria in 1) and 2), all three points taken together yield the lemma. 

We prove statement 1), and the rest works similarly. Condition
a) excludes $X\rightarrow Y$ because we had a causal cycle otherwise. We show b) by showing its contrapositive, that is, if $X \to Y \in G$, then all four implications hold.\\
i) If $P \rightarrow X \in G_X$, then $P \rightarrow X \in G$ since all directed edges in $G_X\setminus G$ are of the form $P \to C$ for $P \in \pa(Y)$, $C \in \ch(Y)$ but $X$ is not a child of $Y$. Combined with $X \to Y$ this yields  $P \to Y \in E_Y$.\\
ii) If $X \rightarrow C \in G_X$, then $X \rightarrow Y \rightarrow C \in G$ or $X \rightarrow C \in G$.  In the first case, it directly follows that $Y \to C \in G_Y$. In the second case, $Y \leftrightarrow C$ is added in the marginalization $G_Y$ since $X \to Y \in G$.\\
iii) If $X \leftrightarrow S \in G_X$, then $X \leftrightarrow S \in G$, which combined with $X \to Y \in G$ implies $Y \leftrightarrow S \in G_Y$.\\
iv) If $Y \to C \in G_Y$, then $X \to C \in G$ or $Y \to C \in G$, which both imply $X \to C \in G_X$ since $X \to Y \in G.$ $\square$
\subsection{Proof of Lemma \ref{lem:determine_edge_type_DAG}}
The criterion for $X$ having at least two children or not, as well as condition (1), can be derived from the marginalization rule that $C_1 \leftarrow X \to C_2 \in G$ turns into an edge $C_1 \leftrightarrow C_2$ in $G_Y$ and this is the only way that bidirected edges can arise. 

Conditions (2) and (3a) follow from rule 1b)iv) specified in the proof of Lemma \ref{lem:nolinks_ADMG}, and (3b) follows from rule 1b)i).

Turning to conditions (3c) and (3d), if $X$ and $Y$ have the same child $C$ in $G_X$,  $G_Y$, the structure of $X, Y, C \in G$ can in principle be either of the following:
\begin{equation*}\label{eq:possible_structures_comon_child}
    i)\ X \to Y \to C, \quad ii)\ Y \to X \to C, \quad\text{or}\quad iii)\ X \to C \leftarrow Y.
\end{equation*} 
To differentiate between them, we include the parents in our consideration.
Denoting $P = \pa^G(X) \cap \bZ, Q = \pa^G(Y)\cap \bZ$ and $R = \pa^G(C)\cap \bZ$, we obtain the following differences in the marginal graphs.
\renewcommand{\arraystretch}{1.2}
\begin{center}
    \begin{tabular}{|c|c|c|c|}
    \hline
     & i)  & ii)  & iii) \\
    \hline
    $G_X$ & 
    \begin{tikzpicture}[baseline={(0,0.4)}]
        \node (X) at (-0.6, 0) {$X$};
        \node (C) at (0.6, 0) {$C$};
        \node (P) at (-0.8, 0.8) {$P$};
        \node (Q) at (0.4, 0.8) {$Q$};
        \node (R) at (0.8, 0.8) {$R$};
        \draw[->] (X) -- (C);
        \draw[->] (P) -- (X);
        \draw[->] (Q) -- (C);
        \draw[->] (R) -- (C);
    \end{tikzpicture} & 
    \begin{tikzpicture}[baseline={(0,0.4)}]
        \node (X) at (-0.6, 0) {$X$};
        \node (C) at (0.6, 0) {$C$};
        \node (P) at (-0.8, 0.8) {$P$};
        \node (Q) at (-0.4, 0.8) {$Q$};
        \node (R) at (0.8, 0.8) {$R$};
        \draw[->] (X) -- (C);
        \draw[->] (P) -- (X);
        \draw[->] (Q) -- (X);
        \draw[->] (R) -- (C);
    \end{tikzpicture} & 
    \begin{tikzpicture}[baseline={(0,0.4)}]
        \node (X) at (-0.6, 0) {$X$};
        \node (C) at (0.6, 0) {$C$};
        \node (P) at (-0.8, 0.8) {$P$};
        \node (Q) at (0.4, 0.8) {$Q$};
        \node (R) at (0.8, 0.8) {$R$};
        \draw[->] (X) -- (C);
        \draw[->] (P) -- (X);
        \draw[->] (Q) -- (C);
        \draw[->] (R) -- (C);
    \end{tikzpicture} \\
    \hline
    $G_Y$ & 
    \begin{tikzpicture}[baseline={(0,0.4)}]
        \node (Y) at (-0.6, 0) {$Y$};
        \node (C) at (0.6, 0) {$C$};
        \node (P) at (-0.8, 0.8) {$P$};
        \node (Q) at (-0.4, 0.8) {$Q$};
        \node (R) at (0.8, 0.8) {$R$};
        \draw[->] (Y) -- (C);
        \draw[->] (P) -- (Y);
        \draw[->] (Q) -- (Y);
        \draw[->] (R) -- (C);
    \end{tikzpicture}  & 
    \begin{tikzpicture}[baseline={(0,0.4)}]
        \node (Y) at (-0.6, 0) {$Y$};
        \node (C) at (0.6, 0) {$C$};
        \node (P) at (0.4, 0.8) {$P$};
        \node (Q) at (-0.8, 0.8) {$Q$};
        \node (R) at (0.8, 0.8) {$R$};
        \draw[->] (Y) -- (C);
        \draw[->] (Q) -- (Y);
        \draw[->] (P) -- (C);
        \draw[->] (R) -- (C);
    \end{tikzpicture} & 
    \begin{tikzpicture}[baseline={(0,0.4)}]
        \node (Y) at (-0.6, 0) {$Y$};
        \node (C) at (0.6, 0) {$C$};
        \node (P) at (0.4, 0.8) {$P$};
        \node (Q) at (-0.8, 0.8) {$Q$};
        \node (R) at (0.8, 0.8) {$R$};
        \draw[->] (Y) -- (C);
        \draw[->] (Q) -- (Y);
        \draw[->] (P) -- (C);
        \draw[->] (R) -- (C);
    \end{tikzpicture} \\
    \hline
\end{tabular}
\end{center}
\renewcommand{\arraystretch}{1}
These differences further imply different relations for the sets of parents in the marginal graphs:
\begin{enumerate}[i)]
    \item $\pa^{G_X}(X) \subseteq \pa^{G_Y}(Y) \subseteq \pa^{G_X}(X) \cup \pa^{G_X}(C)$,
    \item $\pa^{G_Y}(Y)\subseteq \pa^{G_X}(X) \subseteq  \pa^{G_Y}(Y)\cup \pa^{G_Y}(C)$,
    \item  $\pa^{G_X}(X) \subseteq \pa^{G_Y}(C),\ \pa^{G_Y}(Y)\subseteq \pa^{G_X}(C)$.
\end{enumerate} Combining them yields the conditions in the lemma.

Finally, we prove that if neither of the conditions in the lemma apply, we can not identify the edge type. First, note that if no condition in the lemma is satisfied, then either $\ch^{G_X}(X)=\ch^{G_Y}(Y)=\{C\}$ and two of the conditions i) - iii) apply at the same time, or both have no child in the marginal graphs and $\pa^{G_X}(X)\subseteq \pa^{G_Y}(Y)$ or vice versa.
For all these cases,  we need to find two DAGs $G_1$, $G_2$ on the entire set of nodes with different edge types between $X$ and $Y$ but with the same marginalizations. In the case of one common child and i) as well as ii), we can define $G_1, G_2$ by
$$\pa^{G_i}(X) = \pa^{G_X}(X), \pa^{G_i}(W) = \pa^{G_Y}(W) \text{ for all } W \in \bW \setminus \{X\}, i=1,2.$$
Additionally, in $G_1$ we include $X \to Y \to C$, and in $G_2$ we add  $Y \to X \rightarrow C$. If i) and iii) hold, we define $G_1, G_2$ by
$$\pa^{G_i}(X) = \pa^{G_X}(X), \pa^{G_i}(W) = \pa^{G_Y}(W) \text{ for all } W \in \bW \setminus \{X\}, i=1,2,$$
 and include $X \to Y \to C$ in $G_1$ , as well as $X \to C \leftarrow Y$ in $G_2$. All other cases work similarly. $\square$
\subsection{Proof of Theorem \ref{thm:LOVO_via_lingam}}
(1) For a matrix $M \in \R^{l \times m}$, and an index $i \in [l]$, $M_{i:,:}$ denotes the submatrix of all rows starting from the $i$th one. Similarly, for a subset $A \subseteq [l]$, $M_{A,:}$ is the submatrix consisting of all rows with indices in $A$, and $M_{\widehat{A},:}$ the submatrix that arises by omitting all $A$ rows. Throughout the proof we assume that $\bW$ is enumerated as $\bW = (X, Y, Z_1, \dots, Z_k)$. 
Following \cite{Salehkaleybar2020}, we can rewrite \eqref{eq:lingam} as
$\bW = M\bN $
with the "mixing matrix"
$ M := (I-\Lambda)^{-1}$
which linearly combines the independent "sources" $N_1, \dots, N_{k+2}$. Note that $I -\Lambda$ is invertible since $\Lambda$ is strictly lower triangular after applying simultaneous row and column permutations, and the entries of $M$ coincide with the total causal effects defined via \eqref{eq:total_causal_effects}. By observing only the variables $(X, \bZ)$, we have a 
(slightly) over-complete ICA with $k + 1$ observed variables and $k + 2$ sources, namely
\begin{equation}\label{eq:overcomplete_ICA_X}
    \begin{pmatrix}
    X \\ \bZ
\end{pmatrix} = M_{\widehat{2}, :} N 
\end{equation}
where $M_{\widehat{2}, :}$ is the submatrix with the row for $W_2 = Y$ missing, and, similarly
\begin{equation}\label{eq:overcomplete_ICA_Y}
    \begin{pmatrix}
    Y \\ \bZ
\end{pmatrix} = M_{\widehat{1}, :} N.
\end{equation}
The main idea of the proof is to identify $M_{\widehat{2}, :}$ from $P(X, \bZ)$, as well as $M_{\widehat{1}, :}$ from $(Y, \bZ)$ and then combine them to reconstruct $M$. 
From Theorem 15 in \cite{Salehkaleybar2020},  if $Y$ has at least two children, then, 
\begin{enumerate}[a)]
    \item if $Y$ has a unique (with respect to the topological order) oldest child $W_j$, then $M_{\widehat{2}, :}$ can be identified up to swapping the columns corresponding to $Y$ and $W_j$ and up to rescaling of the column corresponding to $Y$. 
    \item Otherwise, $M_{\widehat{2}, :}$ can be identified uniquely up to rescaling the $Y$ column.
\end{enumerate}
If $Y$ has exactly one child $W_j$, then the column in $M_{\widehat{2}, :}$ corresponding to $Y$ is a multiple of the column for $W_j$, in formulas, $$M_{\widehat{2}, 2}= \lambda_{j,2}M_{\widehat{2}, j}.$$ If $Y$ has no children at all,  $M_{\widehat{2}, 2} = 0$. In both cases, obtaining $\bN'$ from $\bN$ by omitting $N_2$, and, in the case of one child, additionally replacing $N_j$ by $N_j' = N_j + \lambda_{j,2}N_{2}$, the vector $(X, \bZ)$
fulfills the complete, and therefore identifiable, ICA model
$$\begin{pmatrix}
    X \\ \bZ
\end{pmatrix} = M_{\widehat{2}, \widehat{2}}N'.$$ Thus,
\begin{enumerate}
    \item[c)] if $Y$ has at most one child, the submatrix $M_{\widehat{2}, \widehat{2}}$ can be identified uniquely. 
\end{enumerate}
Relating back to $M$, in all cases, $P(X, \bZ)$ uniquely determines $M_{\widehat{2}, \widehat{2}}$ or $M_{\widehat{2}, \widehat{\{2, j\}}}$, with $W_j$ being the oldest child of $Y$. In scenario a), additionally, we know two candidate columns $A, B$, where either $M_{\widehat{2}, 2}, M_{\widehat{2}, j} = A, B$ up to rescaling or vice versa. To find the correct assignment, we use the information obtained from $P(Y, \bZ)$; that is, we identified all columns of  $M_{\widehat{1}, :}$ except for column $1$, and at most one other column. In particular, we determined $M_{3:, 2}$ or $M_{3:, j}$. Therefore, comparing whether $A_{2:}$ or $B_{2:}$ coincides with $M_{3:, 2}$ or $M_{3:, j}$, yields correct assignment as well as correct scale. However, this fails in one exceptional case, specifically when $A_{2:}=B_{2:}$ up to scaling. Writing out the entries in $M_{2:, 1}$ $M_{2:,j}$ in terms of the $\lambda_{ij}$, and using faithfulness,
we obtain that this can occur only if $\text{ch}(Y)=\{X, Z_j\}$ and $\text{ch}(X)=\{Z_j\}$, which is excluded in the assumptions of the theorem.

In case b), the only ambiguity in $M_{\widehat{2},:}$ concerns the scaling of the $Y$ column, which again can derived from the information we have on $ M_{\widehat{1}, :}$.

The same holds for reversed roles. So, if for both \eqref{eq:overcomplete_ICA_X}, and \eqref{eq:overcomplete_ICA_Y} identifiability cases a) or b) apply, we can infer $M_{\widehat{2}, :}, $ and $ M_{\widehat{1}, :}$ and combine them to $M$.

If case c) applies in one of the ICAs, assume the one related to $(X, \bZ)$, we are still missing the value of $m_{21}$. Since $Y$ has multiple, and $X$ at most one child, according to Lemma \ref{lem:determine_edge_type_DAG}, $X \to Y \in G$ if and only if $X$ has multiple children in $G_X$. If so, we can choose one of these children $C$. Employing that in the joint model, all directed paths from $X$ to $C$ go through $Y$, we obtain
$$m_{21} = m_{C1}/m_{C2}.$$ If $X \centernot{\to} Y$, $\lambda_{21}=0$,  which determines $m_{21}$  via $\Lambda = I - M^{-1}$.

If case c) applies in both ICAs, then both $X, Y$ have at most one child, which corresponds to conditions (3a)-(3d) in Lemma \ref{lem:determine_edge_type_DAG}. In cases (3a)-(3c), we know 
$X \centernot{-} Y$, and therefore $\lambda_{21}=\lambda_{12}=0$, which gives $m_{12}, m_{21}$  via $\Lambda = I - M^{-1}$. In case (3d), we know that $X$ and $Y$ have the same child $C$ in the marginal graphs, whereas in the joint graph, $X \to Y \to C$ without a direct connection between $X$ and $C$, since this would contradict the fact that $X$ has only one child. Therefore, as above
$$m_{21} = m_{C1}/m_{C2}.$$ Moreover, we that find $m_{12}=0$ due to acyclicity. Again, the same holds for reversed roles.
Finally, we can compute $\Lambda$ as $\Lambda = I - M^{-1}$.

(2) To prove the identifiability of $P(X, Y, \bZ)$, we use the fact that once the projected mixing matrices $M_{\widehat{1},:}$, $M_{\widehat{2},:}$ in \eqref{eq:overcomplete_ICA_X}, \eqref{eq:overcomplete_ICA_Y} are known, under the genericity assumption on the moments, all cumulants of the exogenous sources $\bN$ can be identified \citep[Lemma 5]{Schkoda2024}
These cumulants uniquely determine $P(\bN)$, which, combined with the overall mixing matrix $M$, yields $P(X, Y, \boldsymbol{Z})$. $\square$

\section{Additional experiments}
\subsection{How often do Lemmas \ref{lem:nolinks_ADMG} - \ref{lem:determine_edge_type_DAG} succeed in excluding edges?}\label{subsec:add_details_performance_lemmas}
\begin{figure}
\centering
\includegraphics[width=0.32\textwidth]{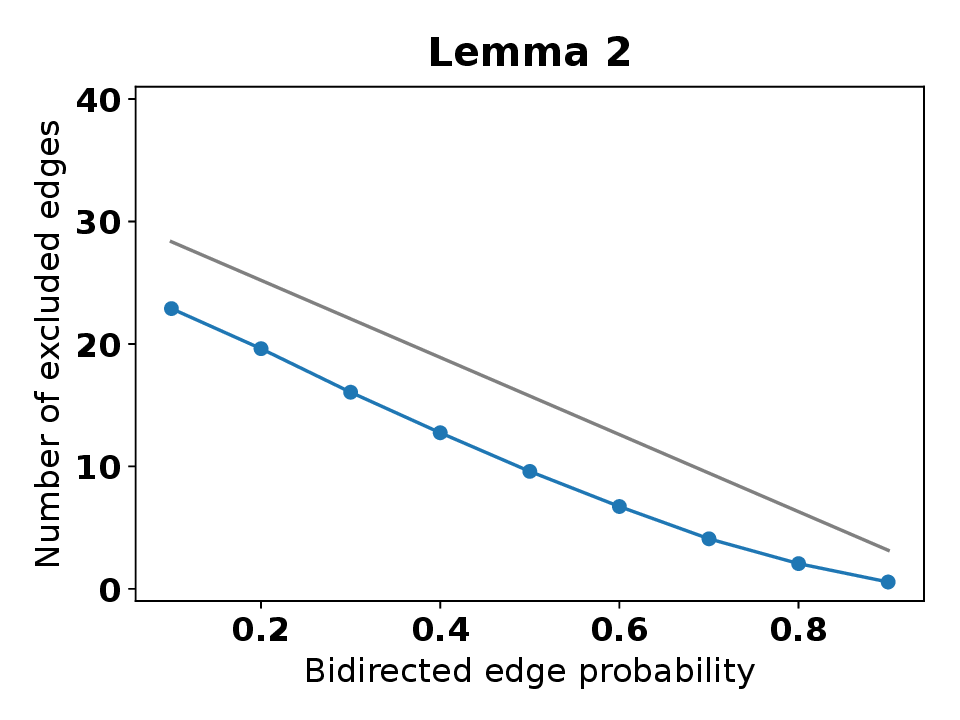}
\includegraphics[width=0.32\textwidth]{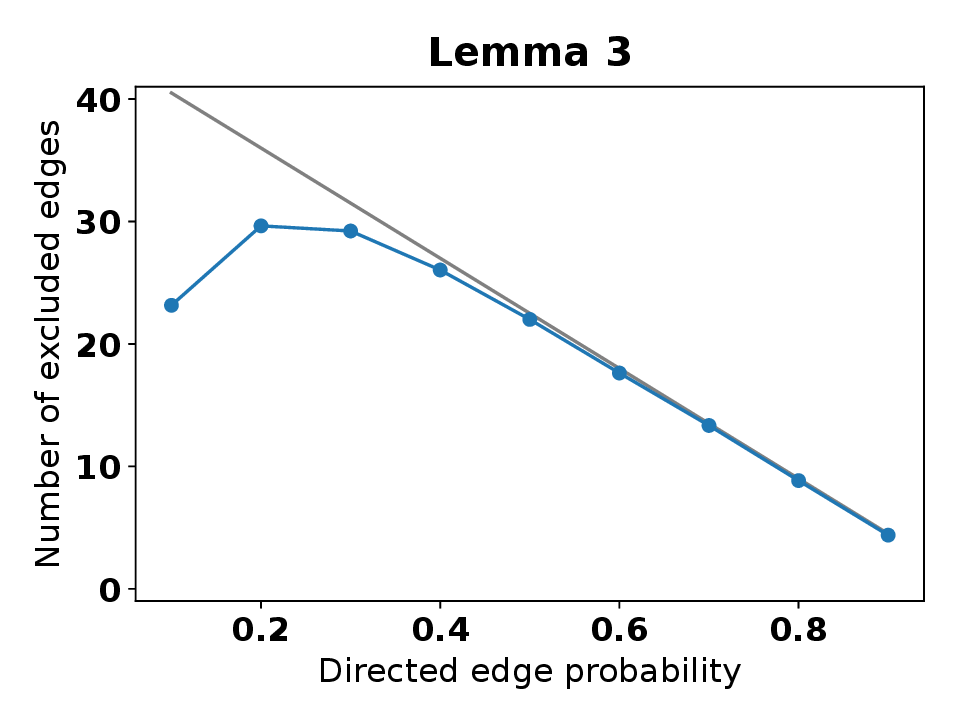}
\caption{For Lemma \ref{lem:determine_edge_type_DAG} and $p\geq 0.3$, the average number of detected absent edges (blue) is close to the true number of absent edges (grey), whereas Lemma \ref{lem:nolinks_ADMG} does not find all absent edges. }\label{fig:mean_nr_excluded_edges}
\end{figure}

While the measurements depicted in Figure \ref{fig:no_excluded_edges} give insight into how often the Lemmas find at least one pair without edge per graph, which is the crucial factor for the realizability of LOVO, another interesting question is what proportion of unlinked pairs are recognized by the lemmas as such. To address this, Figure \ref{fig:mean_nr_excluded_edges} compares the average number of detected unlinked pairs (blue) to the number of pairs in the graph that are actually not connected, which is expected to be $(1-p)\cdot \binom{10}{2}$ in an Erdős–Rényi DAG with edge probability $p$, and $(1-p)(1-q)\cdot \binom{10}{2}$  for ADMGs with directed edge probability $p$ and bidirected edge probability $q$ (grey).

\subsection{LOVO applied to DirectLiNGAM and RCD with varying sample size}\label{subsec:varying_n}
\begin{figure}
    \centering
    \includegraphics[width=0.24\linewidth]{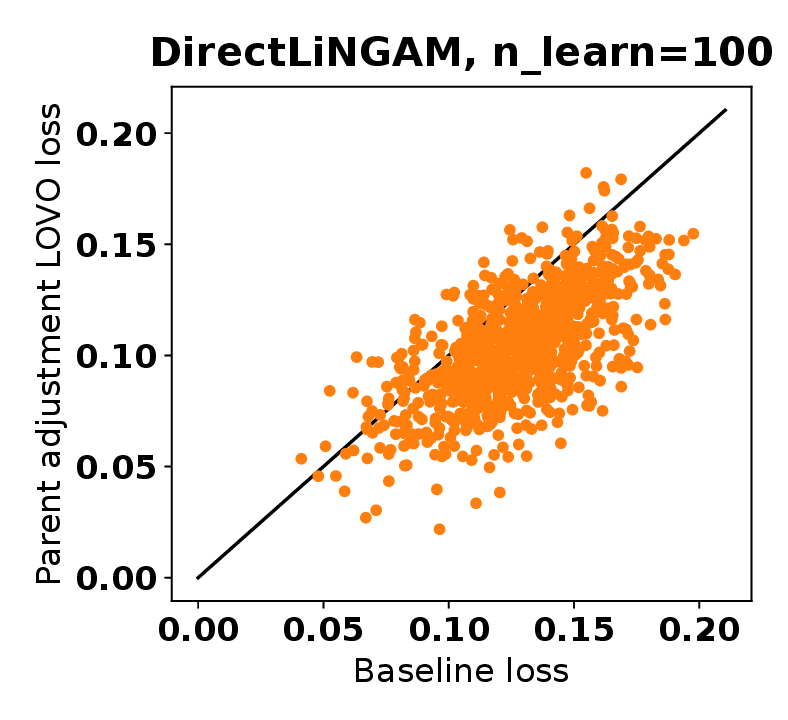}
    \includegraphics[width=0.24\linewidth]{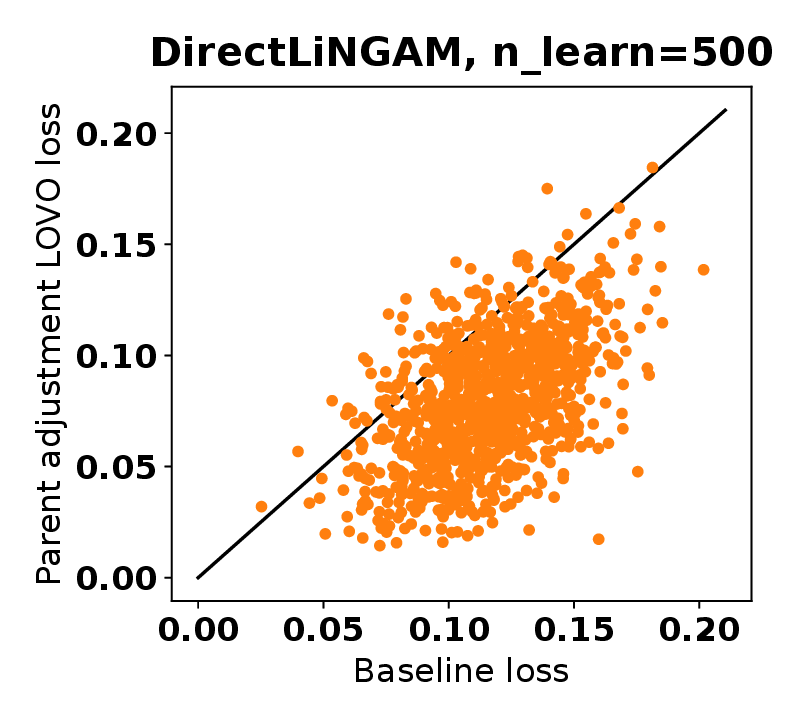}
    \includegraphics[width=0.24\linewidth]{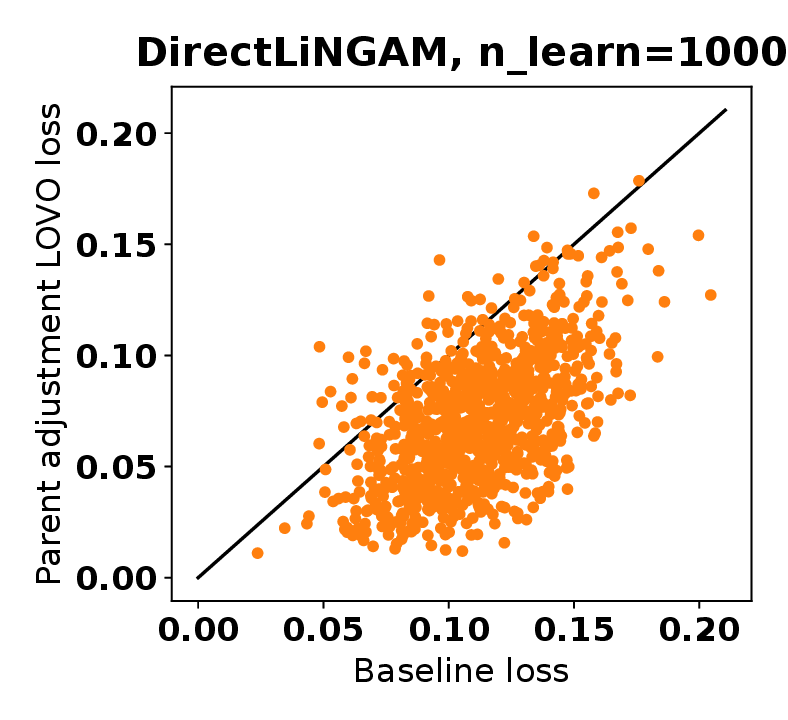}
    \includegraphics[width=0.24\linewidth]{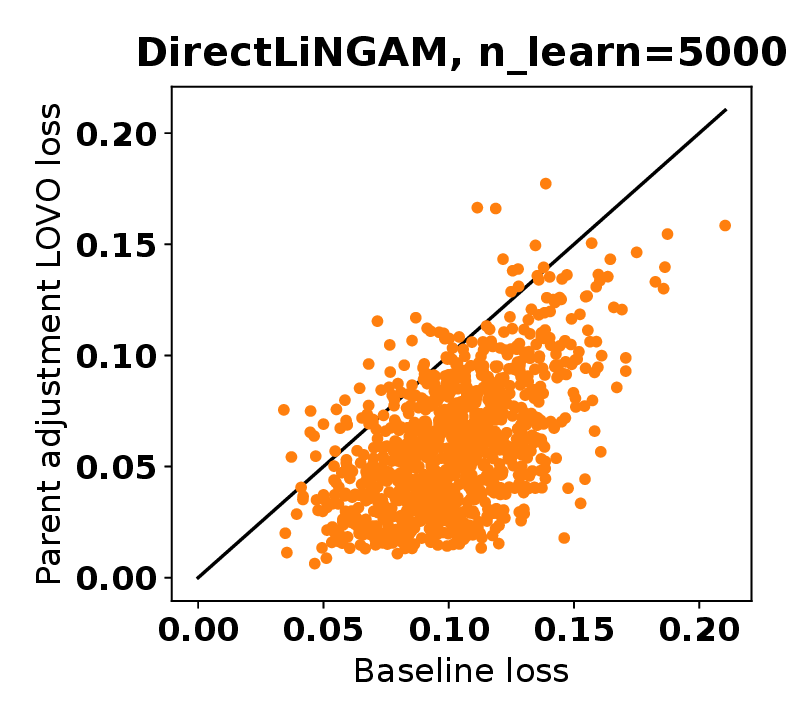}
    \includegraphics[width=0.24\linewidth]{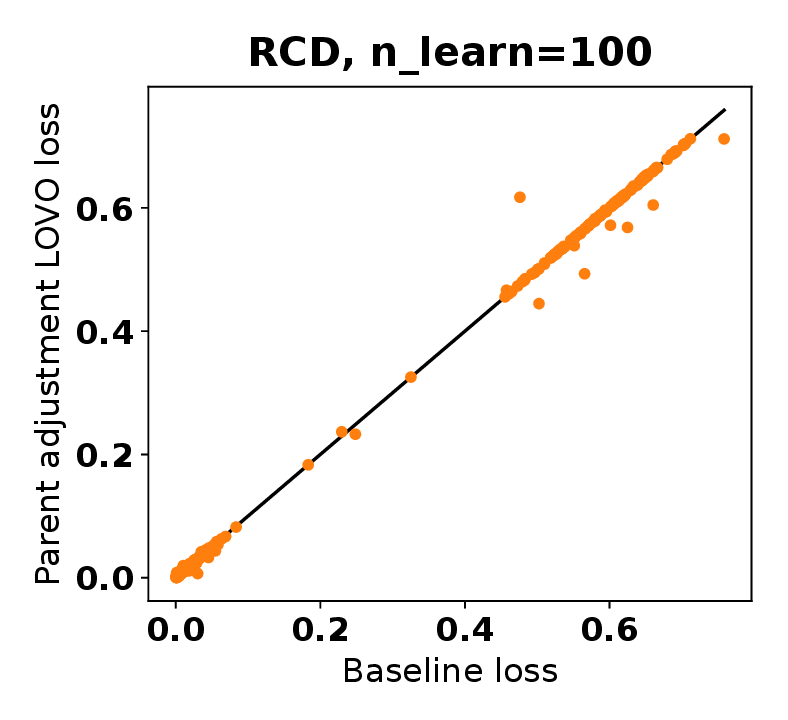}
    \includegraphics[width=0.24\linewidth]{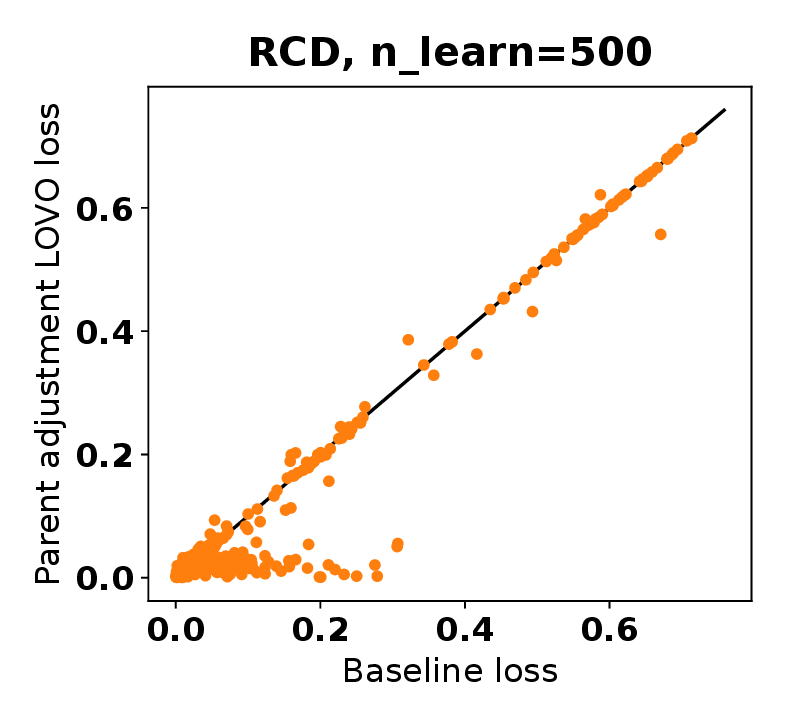}
    \includegraphics[width=0.24\linewidth]{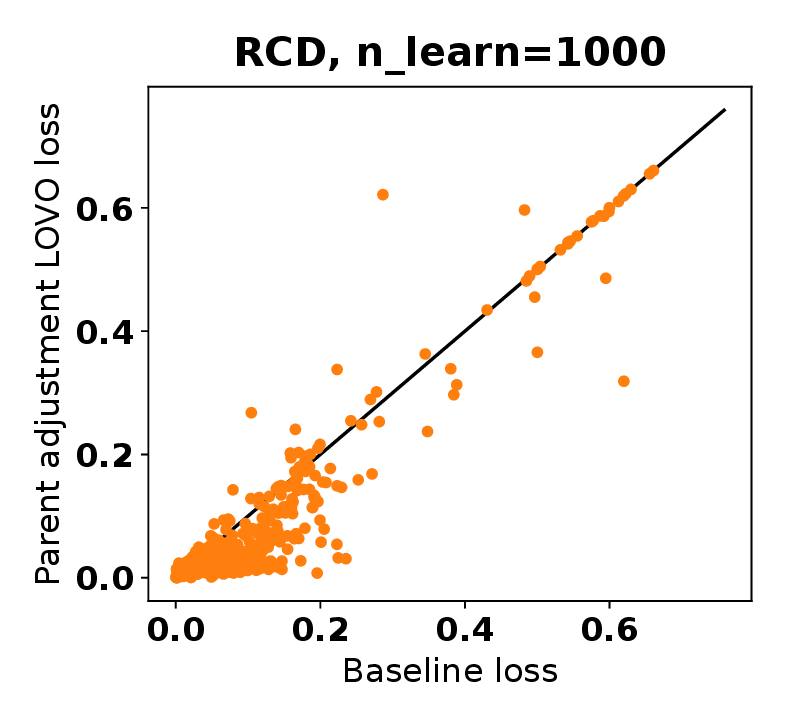}
    \includegraphics[width=0.24\linewidth]{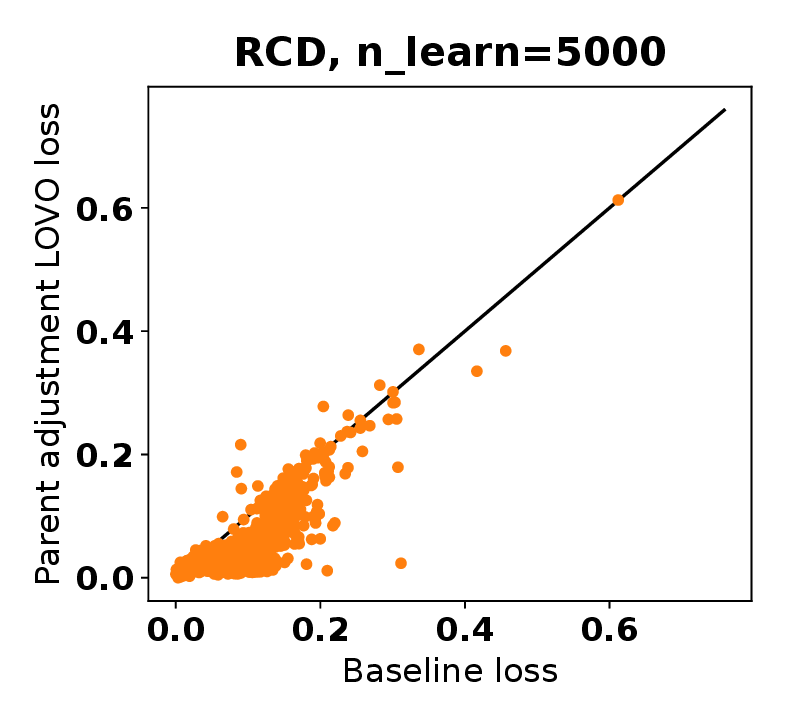}
    \caption{Evolution of LOVO prediction loss when learning sample sizes increases, and by that, the accuracy of the learned graphs increases.}
    \label{fig:lovo_varying_n_learn}
\end{figure}
 This section provides more details on the experiment described in \ref{subsec:experiment_RCD_DirectLiNGAM}. Specifically, we analyze the LOVO prediction loss for varying sample sizes, which is expected to relate to the accuracy of the learned graphs. Indeed, the mean SHD of a marginal graph learned with DirectLiNGAM is $12.3, 6.9, 6.5, 6.5$ for $n_{\text{learn}}=100, 500, 1000, 5000$, respectively, and $4.0, 2.4, 1.5, 0.3$ for RCD.
Figure \ref{fig:lovo_varying_n_learn} shows that the LOVO prediction loss tends to decrease with increased learning sample size. A notable observation is that, for RCD with smaller sample sizes, the LOVO loss is very close to the baseline loss and often abstains from making predictions, doing so in about $52\%$ of the replications.  This is due to RCD almost always only learning bidirected and no directed edges, meaning it does not commit to any causal directions, which makes it  harder to challenge its output; compare Section A11 in \citet{Faller2024}. In the cases where predictions are made, the learned union of parents, which is  the adjustment set in LOVO, is almost always empty. Consequently, $\hat{\rho}^{\text{LOVO}}=0$, and both prediction errors are close to the absolute value of true correlation $\rho_{XY}$.
Moreover, the scatter plot reveals two clusters cluster characterized by error values below and above $0.3$.
The cluster with lower errors corresponds to pairs correctly identified as unlinked, while the other cluster contains pairs where an edge exists. Note that we can relate the points in the scatter plot to pairs, even though each point represents the cross-validation error averaged over all pairs, since the cross-validation error was often only computed using one pair and the LOVO predictor abstained for all other pairs. The cluster related to higher values gets smaller for increased sample size and eventually disappears. Also the number of how often LOVO abstains decreases; to $37\%, 29\%, 3\%$ for $n_{\text{learn}}=500, 1000, 5000$. In contrast, for DirectLiNGAM, LOVO abstained more rarely for lower sample sizes, specifically in $0.3\%, 2\%, 1\%, 23\%$ of the replications for  $n_{\text{learn}}=100, 500, 1000, 5000$.

\subsection{Comparison to Self-Compatibility}
{\color{new} To compare with the self-compatibility score from \cite{Faller2024}, we repeat the same experiment. Recall that the self-compatibility score runs the algorithm in question on various subsets and counts the incompatibilities the resulting graphs entail. Figure \ref{fig:self-comp} plots the LOVO loss against the self-compatibility score evaluated on RCD and DirectLiNGAM. For the data-generation, the same parameters as in the preceding subsection are used, and we collect the results for all the four different values of $n_{\text{learn}}$ into one plot.
\begin{figure}
    \centering
    \includegraphics[width=0.33\linewidth]{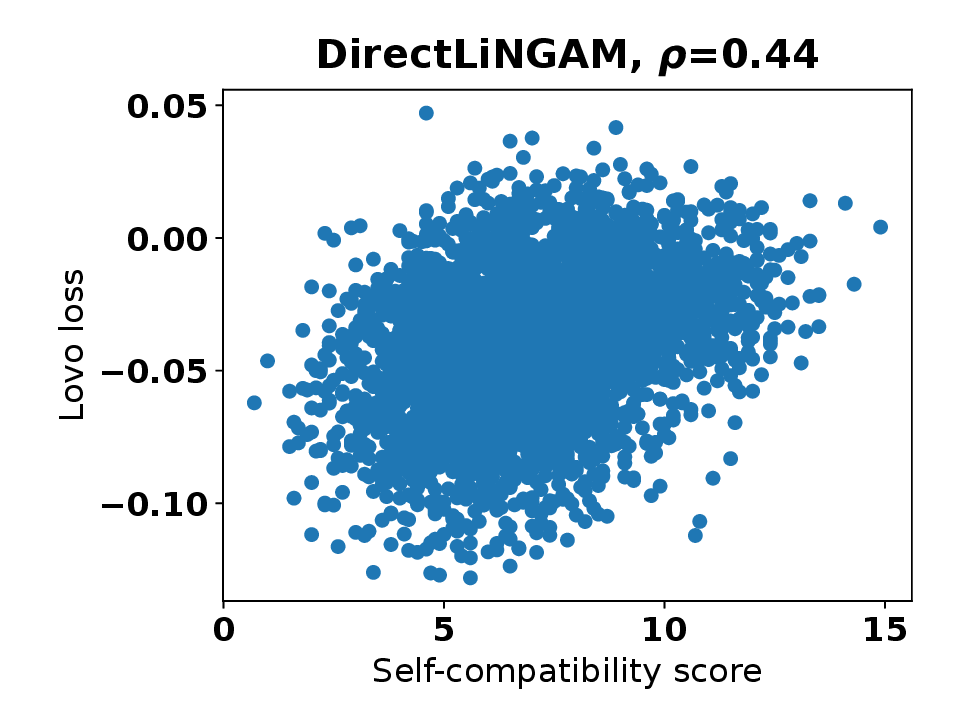}
    \includegraphics[width=0.33\linewidth]{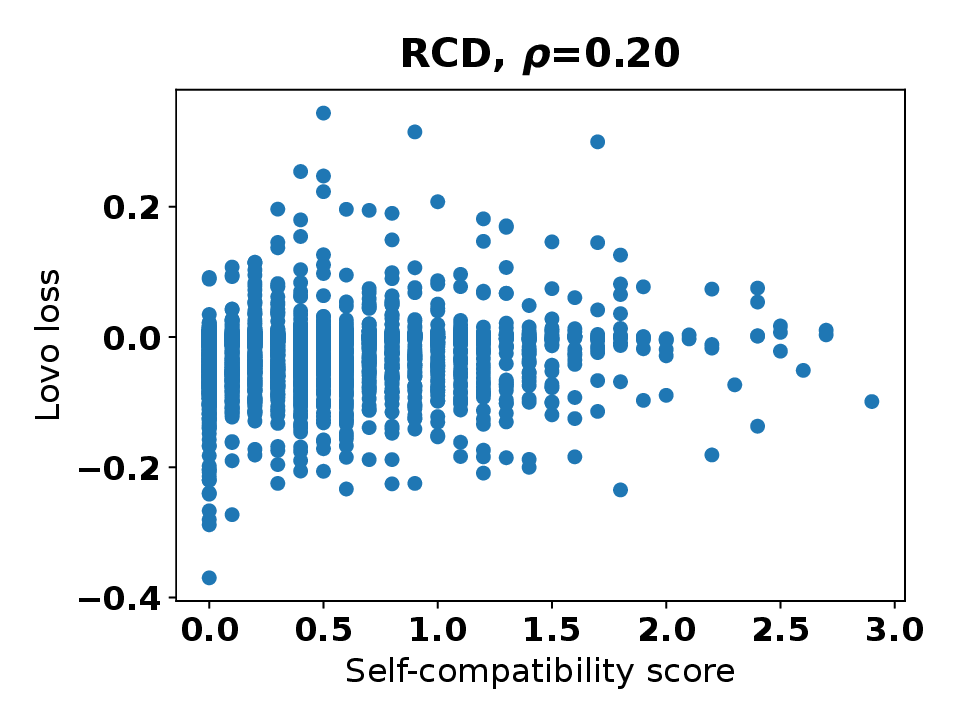}
    \caption{\color{new}While the self-compatibility score and LOVO loss are significantly correlated, the two metrics show distinct patterns, reflecting their different objectives of incompatibility counting and LOVO prediction.}
    \label{fig:self-comp}
\end{figure}
}

{\subsection{\color{new}LOVO applied to DirectLiNGAM and RCD on real-world data}\label{subsec:real-world}
\color{new}
\begin{figure}
    \centering
    \begin{minipage}[c]{0.33\linewidth}
        \includegraphics[width=\linewidth]{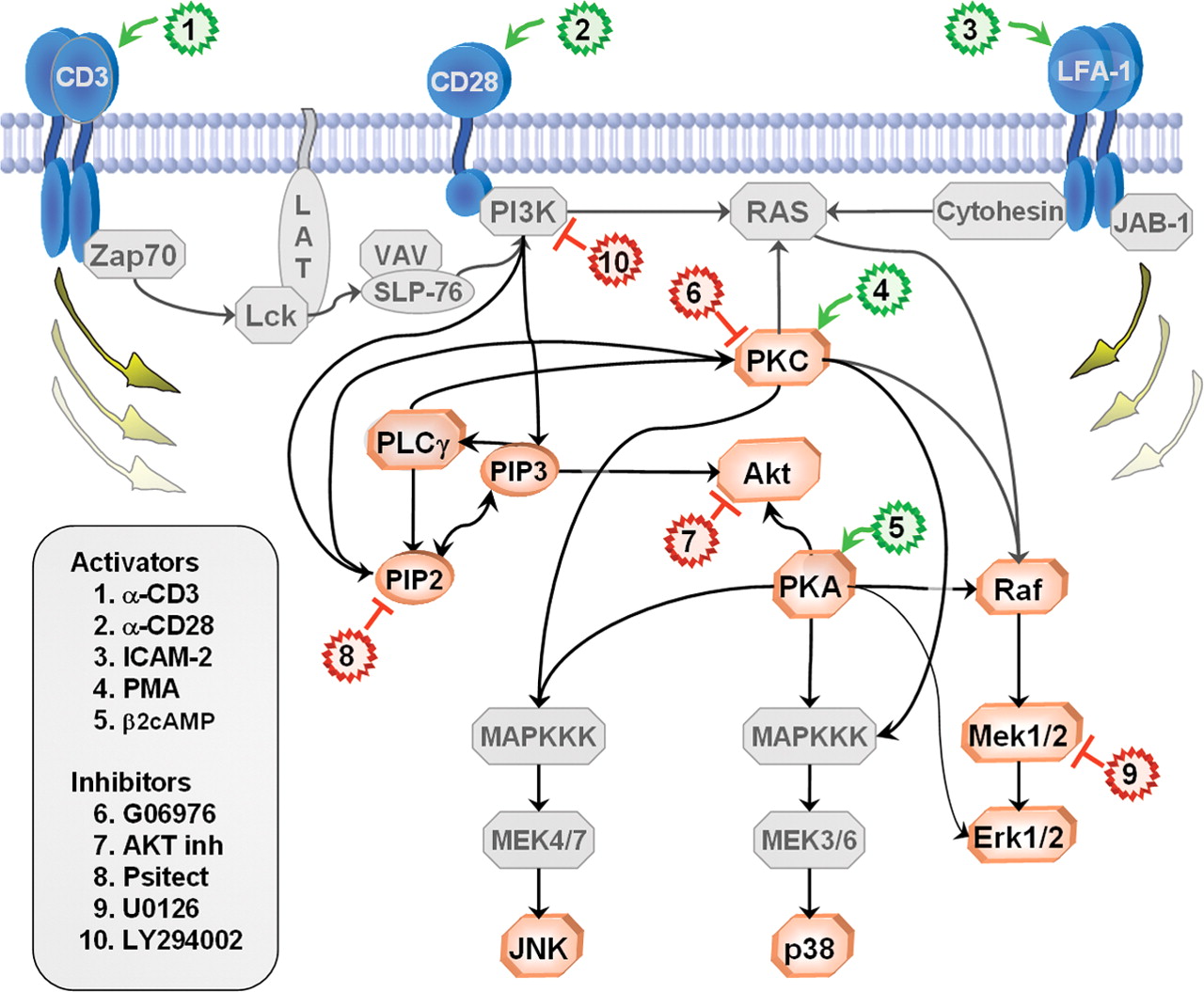}
    \end{minipage}
    \hspace*{0.8cm}
    \begin{minipage}[c]{0.5\linewidth}
        \includegraphics[width=\linewidth]{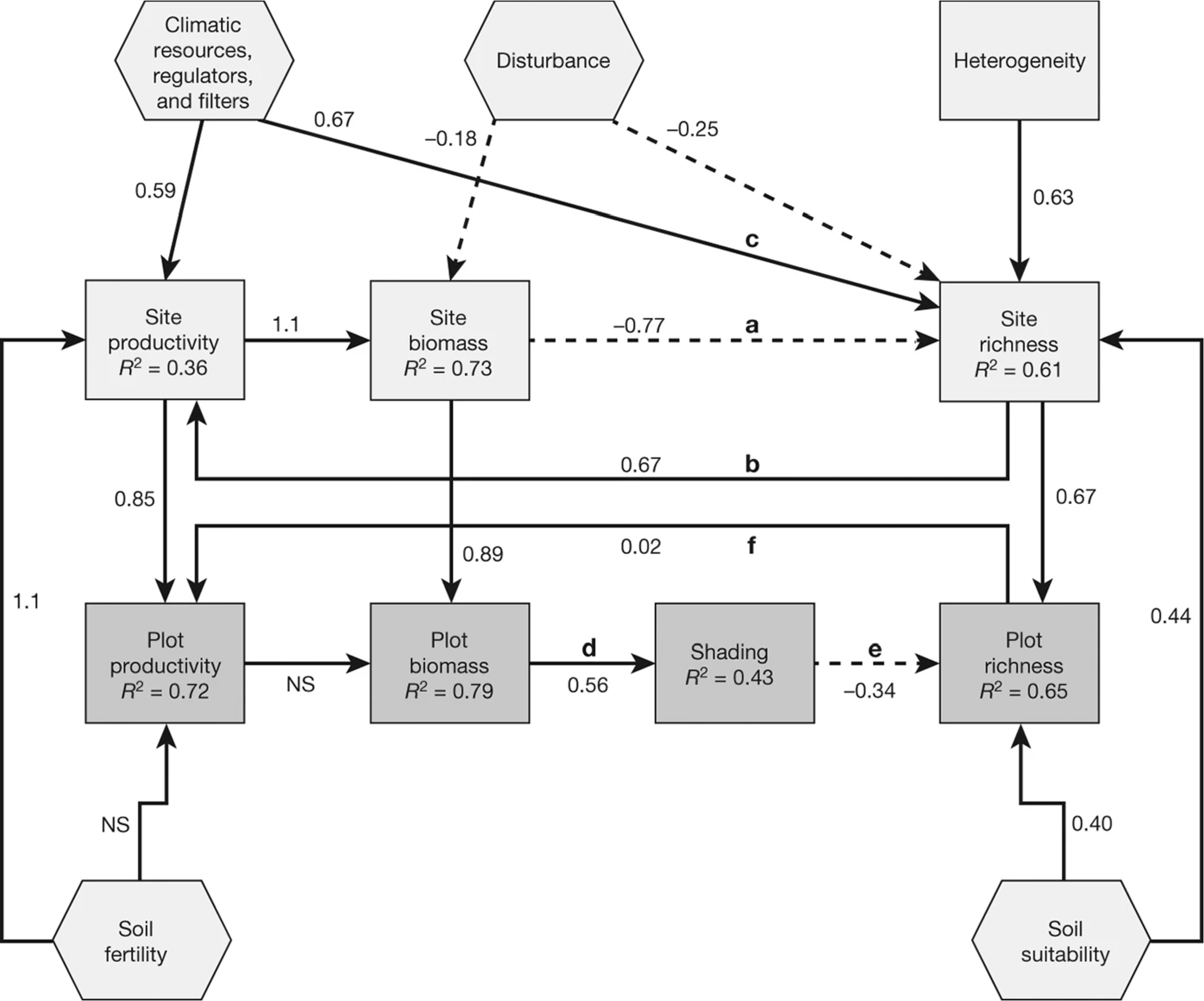}
    \end{minipage}
    \caption{\color{new}For both data sets, the original papers postulate ground-truth causal graphs, which are visualized here. The left plot shows the Sachs data; all the new nodes refer to features in the data set, and the remaining ones are unobserved variables.}
    \label{fig:real-world}
\end{figure}
\color{new}

We evaluate RCD and DirectLiNGAM on two real-world data sets: a protein-signaling data set from \cite{sachs2005causal} with 7466 samples across 11 features and a biodiversity and productivity data set from \cite{Grace_2016} comprising 1126 samples and 12 variables (Figure \ref{fig:real-world}). For the second data set, the original study estimated effects using two separate models: one for the upper seven variables and another for the remaining variables in the lower three rows, except \textit{soil fertility}. Given the larger sample size in the second model, where the variables are measured for each of the 1126 \textit{plots} versus the coarser-scale \textit{sites} used for the upper variables, we focus our analysis on the second model.
\color{new}

On both datasets, RCD very rarely commits to a non-trivial output in the sense that it almost always returns a complete ADMG, which indicates that the datasets do not satisfy the method's assumptions. Given these outputs not entailing any statements, LOVO always abstains. In contrast, DirectLiNGAM always returns a DAG, and we can compute the LOVO cross-validation error, which is worse than the baseline in both cases; see Table \ref{tab:real-world}. As a comparison, we also apply LOVO to the hypothesized ground truth graphs depicted in Figure \ref{fig:real-world}. The required marginal graphs $G_X, G_Y$ are derived via the marginalization rules. For the protein dataset, the LOVO prediction error is comparable to the baseline error, while for the ecology dataset, the LOVO error is smaller.
}

\begin{table}
\centering
\caption{\color{new}The LOVO and baseline errors for the real-world data sets suggest that neither DirectLiNGAM nor RCD is a suitable algorithm for these data.}
\label{tab:real-world}
\begin{tabular}{|l|l|c|c|c|}
\hline
  &   & DirectLiNGAM & RCD & ``Ground truth''  \\
  \hline
\multirow{2}{*}{\cite{sachs2005causal}} &  LOVO error  & 0.152 & - & 0.187   \\
\cline{2-5}
  &  Baseline error  & 0.146 & - & 0.187  \\
\hline
\multirow{2}{*}{\cite{Grace_2016}} &  LOVO error  & 0.403 & - & 0.354 \\
\cline{2-5}
  &  Baseline error  & 0.376 & - & 0.436  \\
\hline
\end{tabular}
\end{table}

\subsection{Further details for Subsection \ref{subsec:dl}}\label{sec:additional_details}

\begin{figure}
		\centering
		\includegraphics[width=0.33\textwidth]{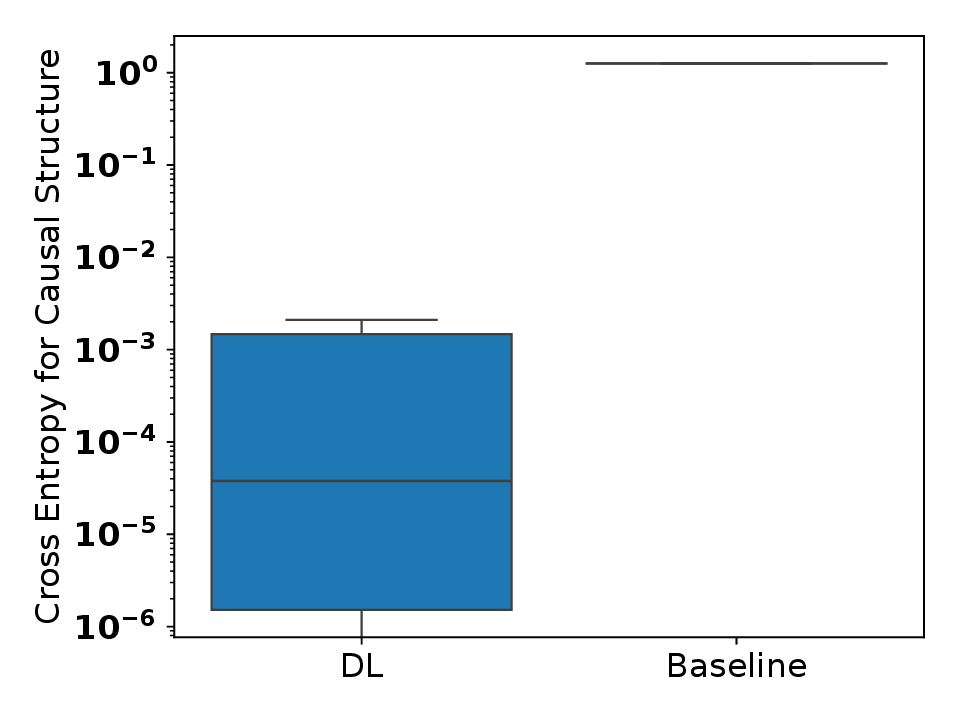}
		\caption{\label{fig:structure_cross_entropy} 
			Prediction error for recovering the causal structure from the learned representation of the deep learning model on unseen examples (as described in Section \ref{subsec:dl}). Our prediction model has a lower loss than the baseline in all examples.} 
	\end{figure}
 \paragraph{Architecture}
 The main component of the deep learning model used in Section \ref{subsec:dl} is the encoder from \cite{ke2022learning}.
 This encoder gets as input a data matrix $\bW\in \R^{n\times d}$, where $n\in \N$ is the number of samples and $d\in \N$ is the number of variables (in our case $d=2$).
 Initially, each entry of this matrix is embedded into $\R^{h/2}$ with a linear transformation.
 We also embed the column index (i.e. the node identity) of every entry into $\R^{h/2}$.
 Unlike \cite{ke2022learning}, we use another linear transformation for this.
 We concatenate this input embedding and identity embedding to get an $h$-dimensional representation of every entry.
 Eventually, we add a row of zeros to the initial data matrix $\bW$, which will be used later for the encoder summary.  
 This gives us an initial embedding $e^0 \in \R^{n + 1\times d\times h}$.
 
\cite{ke2022learning} propose to alternatingly use attention blocks that calculate attention weights between samples for every node and between nodes for every sample.
At attention layer $i$ we expect an input tensor $e^{i-1}\in \R^{n+1\times d\times h}$ and start by applying a classical self-attention  block to all matrices $e^{i-1}_j\in \R^{d\times h}, j=1,\dots, n$ that result from indexing the sample dimension in $e^{i-1}$.
Implementation-wise, this amounts to passing $e^{i-1}$ to a standard attention layer and considering the first dimension as batch dimension.
We then apply a feed-forward layer.
This results in a tensor $\hat e^i \in \R^{n+1 \times d\times h}$.
The second attention block is then applied to the matrices $\hat e^i_{:, j}, j=1, \dots, d$, that result from indexing the nodes, i.e., we reshape the tensor to have the second dimension as batch-dimension (and reshape it back after the attention block).
After every attention block, we add a feed-forward layer, and we add a pre-layer norm and a residual connection to every attention block and feed-forward layer.

The final encoder summary is obtained by another attention block, where we consider the column dimension the batch dimension again, and we only use the $(n+1)$-th row as key and all other rows as queries.
This gives us a final embedding $e\in \R^{d\times  h}$, which we flatten to be in $\R^{d\cdot h}$.

Instead of the decoder proposed by \cite{ke2022learning}, we  add another feed-forward layer that receives the concatenated embeddings of each marginal dataset as input and outputs a scalar.

For the second model that predicts the causal structure from the hidden representations  $e_{X, Z}$ of the first model, we simply used a feed forwards layer with four output dimensions and a softmax layer to encode the four possible causal structures $\{\to, \gets, \leftrightarrow, \text{no edge}\}$ as categories.

\paragraph{Training}
\begin{table}[t]
	\centering
	\caption{Hyperparameters for training the deep learning models.}
	\begin{tabular}{lc}
		\toprule
		Hyperparameter & Value\\
		\midrule
			Batch size & 1\\
			Learning rate & 1e-4\\
			Gradient clipping value & 10\\
			Epochs & 2\\
			Encoder layers & 3\\
			Feed forward hidden layers & 1\\
			Feed forward widening & 4\\
			Activation & GELU\\
			Attention heads & 8 \\
			Hidden dimension $h$ & 64\\
			Samples per dataset & 3000\\

			Test examples & 100\\
			Loss Transitive Prediction & MSE\\
			Training examples Transitive Prediction& 100000\\
			Loss Structure from Embedding & Cross-entropy\\
			Training examples Structure from Embedding& 1000\\
			\bottomrule
	\end{tabular}
	\label{tab:hyperparams}
\end{table}
Note that \citet{ke2022learning} propose to train the model on a dataset, where each "datapoint" consists of a synthetically generated adjacency matrix as target with a matrix containing multiple samples drawn from this graph as input features. 
For every such datapoint, we  generate  one of the 12 DAGs consisting of two nodes and three edges (compare Table \ref{tab:dags}) with
equal probability, and data with sample size $3n$ as in Section \ref{subsec:experiment_parents}. 
Again, we split the data into three equal-sized subsets.
Let $m\in\N$ be the number of datapoints of the training set and
denote the sample matrices for the $j$-th adjacency matrix with
\begin{align*}
    M^j_{X, Z} &:= \{x^j_i, z^j_i\}_{i=1, \dots, n}\\
    M^j_{Y, Z} &:= \{y^j_i, z^j_i\}_{i=n+1, \dots, 2n}\\
    M^j_{X, Y} &:= \{x^j_i, y^j_i\}_{i=2n +1, \dots, 3n}.
\end{align*}
We then solve the minimization problem
\begin{displaymath}
	\arg\min_{f\in \cF} \sum_{j=1}^m \left(f(M^j_{X, Z}, M^j_{Y, Z}) - \hat \rho^j_{X,Y}\right)^2,
\end{displaymath}
 where  $\cF$ is the function class defined by the model architecture and $\hat \rho^j_{X,Y}$ is the correlation coefficient computed from the third part of the samples $M^j_{X, Y}$. 
The parameter settings of the main model training are summarized in Table \ref{tab:hyperparams}.
Most notably, we generated 100000 pairs of marginal data matrices and respective ground truth correlation coefficient $\rho_{X,Y}$.

The second model (that is trained to predict the causal structure) is simply a feed-forward network with a single hidden layer.
To train it, we generate $k\in \N$ more marginal samples $\{x^j_i, z^j_i\}_{i=1, \dots, M}$ and apply the encoder from the pre-trained model above to get a dataset 
\begin{displaymath}
\{(e^j_{X, Z}, s_j)\}_{j=1, \dots, k},    
\end{displaymath}
 where $s_j\in \{\to, \gets, \leftrightarrow,  \centernot{-}\}$  denotes the true underlying causal structure that generated the $j$-datapoint.
 We use a cross-entropy loss to train this model.
The second model is trained on 1000 pairs of embeddings $e_{X, Z}$ and (one-hot encoded) underlying structure.
We trained the LOVO prediction model using a squared loss and the second model using the cross-entropy loss.
Unless stated otherwise, we used the same parameters for the main model and the second model.

\paragraph{Computational resources}
The main deep learning model from Section \ref{subsec:dl} was trained on an AWS EC2 instance of type \texttt{p3.2xlarge}.
These machines contain  Intel Xeon E5-2686-v4 processors with 2.3 GHz and 8 virtual cores as well as 61 GB RAM.
The training ran less than an hour.
All inference and further experiments were run on a MacBook Pro with Apple M1 processor and 32 GB RAM and can be run in less than an hour.

\new{\subsection{Evaluating the LOVO variants for DAGs, CPDAGs, and PAGs}
This section sheds light on our LOVO version for DAGs, CPDAGs, and PAGs as proposed in Appendix \ref{sec:LOVO_further_graph_types}. As in Section \ref{subsec:experiment_parents}, we first explore how often Lemmas \ref{lem:exclude_links_directed_part} for DAGs and CPDAGs and Lemma \ref{lem:nolinks_ADMG} using the limited information entailed by PAGs succeed in finding unlinked pairs. Again, we report the mean number of excluded edges in each run (Figure \ref{fig:mean_nr_excluded_edges_further_graph_types}) as well as the number of runs where no single unlinked pair can be found (Figure \ref{fig:no_excluded_edges_further_graph_types}). Next, we employ the LOVO prediction based on the true marginal graphs and simulated data and plot the LOVO loss in comparison to the baseline loss in Figure \ref{fig:LOVO_further_graph_types}. For DAGs, we use parent adjustment as before, whereas for CPDAGs and PAGs, we adjust using the set of all nodes on paths between $X$ and $Y$ excluding colliders as described in Section \ref{subsec:CPDAGs_PAGs}.
\begin{figure}
\centering
\includegraphics[width=0.32\textwidth]{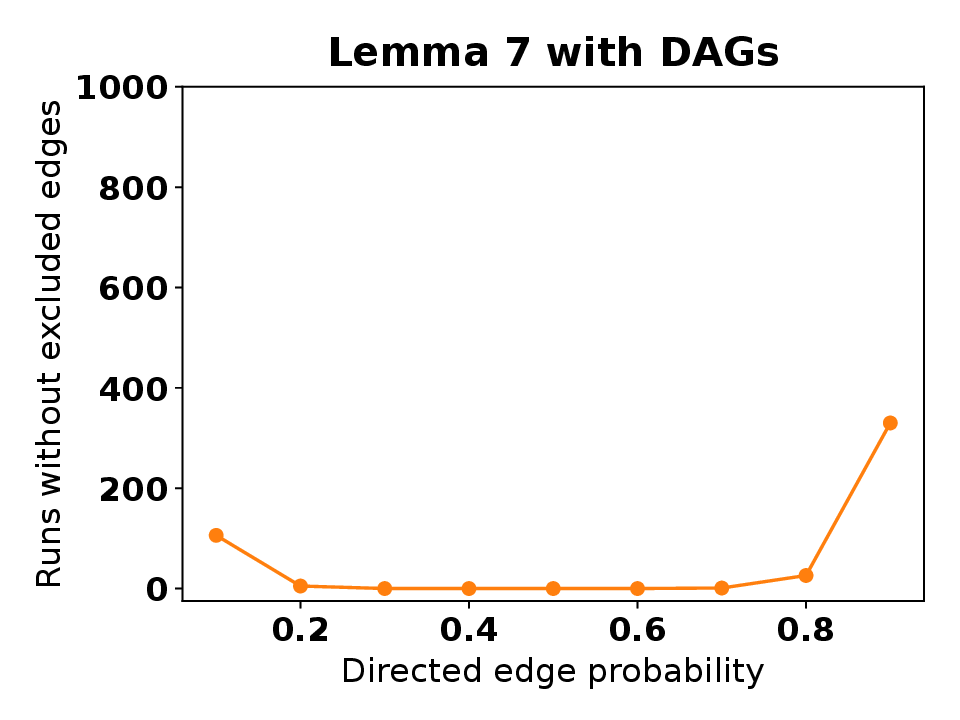}
\includegraphics[width=0.32\textwidth]{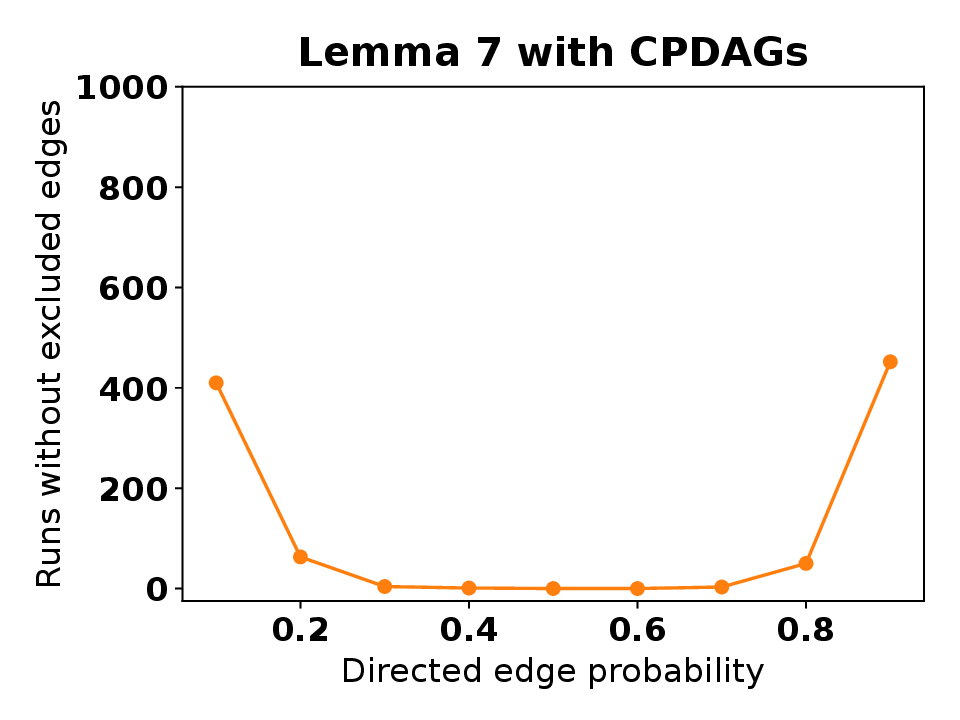}
\includegraphics[width=0.32\textwidth]{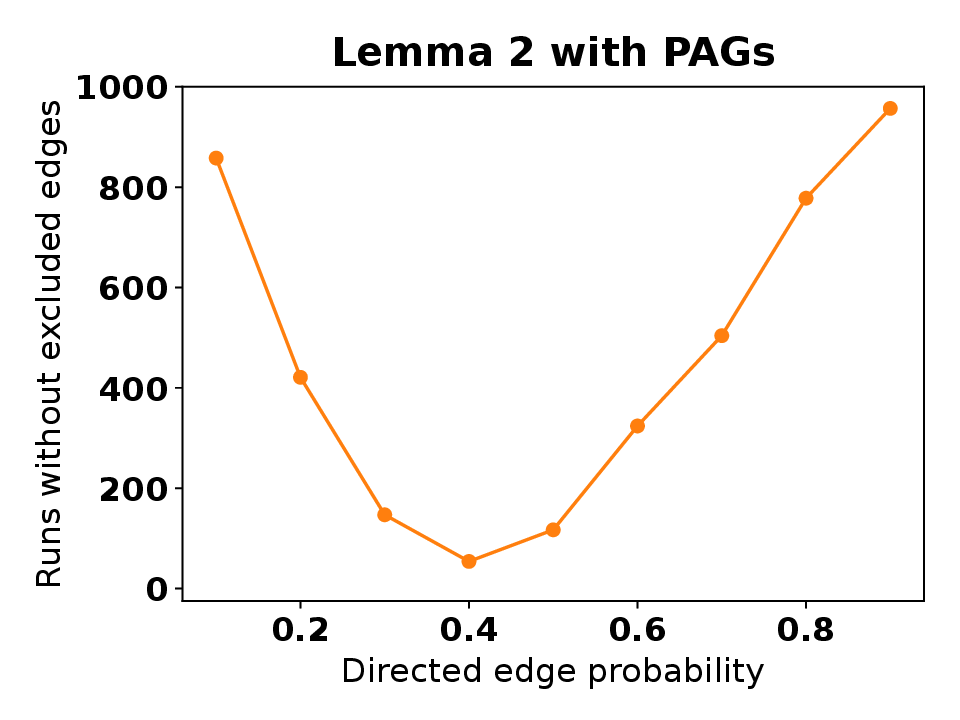}
\caption{\color{new}
Lemma \ref{lem:exclude_links_directed_part} detects at least one feasible pair in almost all runs for $p \in [0.2,0.8]$ and DAGs as well as CPDAGs. Lemma \ref{lem:nolinks_ADMG} employed with PAGs finds feasible pairs in most runs when $p \in [0.3,0.5]$, but often does not find any when $p$ is close to $0$ or $1$.
}\label{fig:no_excluded_edges_further_graph_types}
\end{figure}

\begin{figure}
\centering
\includegraphics[width=0.32\textwidth]{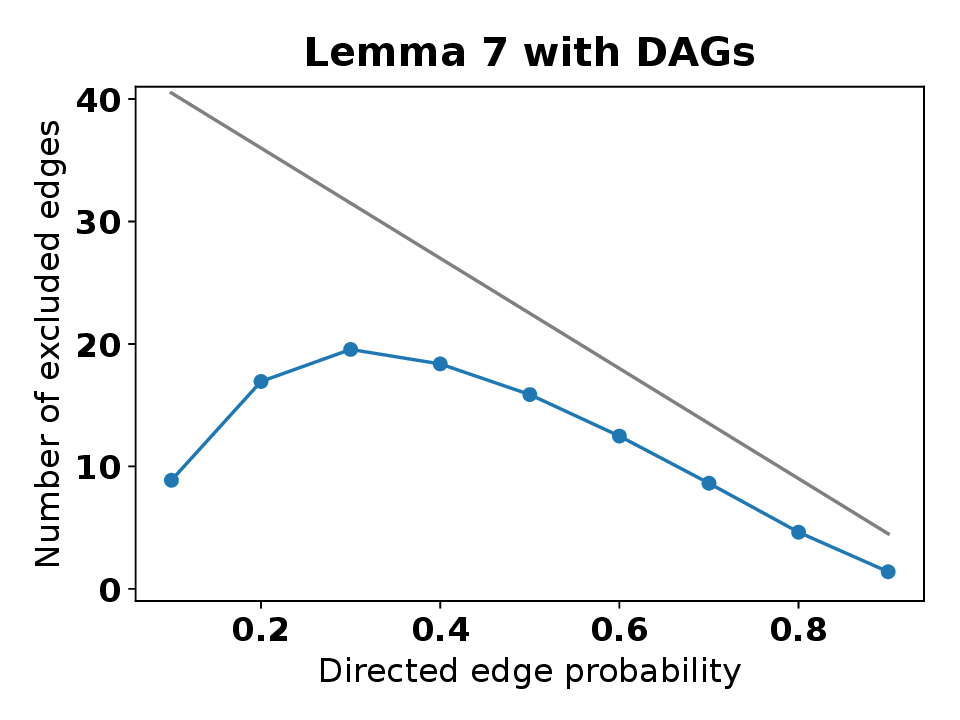}
\includegraphics[width=0.32\textwidth]{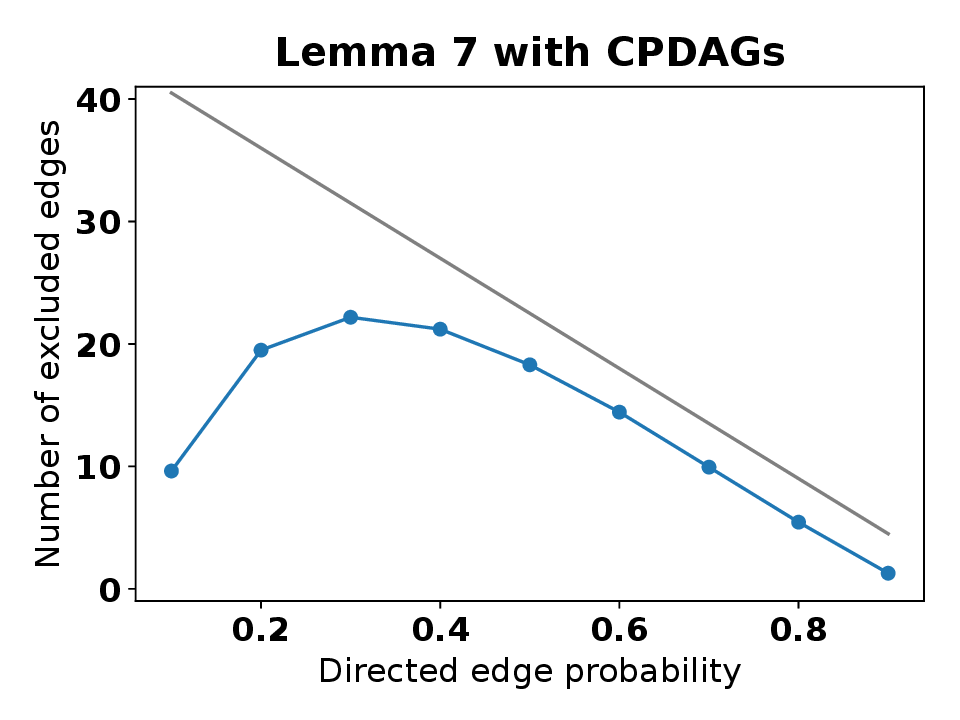}
\includegraphics[width=0.32\textwidth]{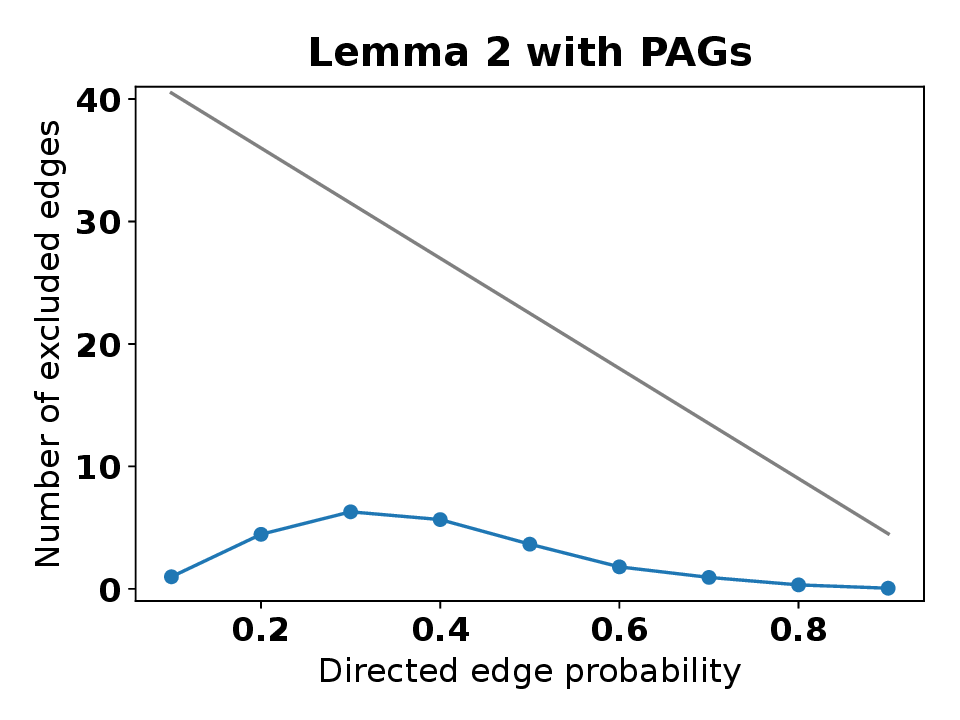}
\caption{\color{new}
For Lemma \ref{lem:exclude_links_directed_part} and $p\geq 0.3$, the average number of detected absent edges (blue) is close
to the true number of absent edges (grey), whereas Lemma \ref{lem:nolinks_ADMG} does not find all absent edges.}\label{fig:mean_nr_excluded_edges_further_graph_types}
\end{figure}

\begin{figure}
\centering
\includegraphics[width=0.32\textwidth]{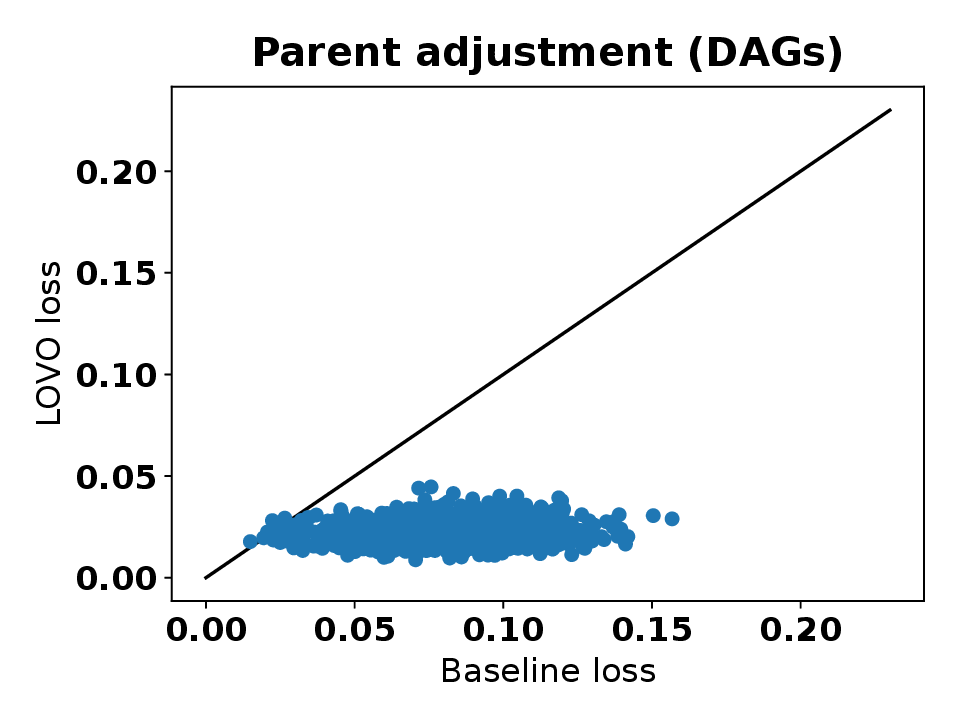}
\includegraphics[width=0.32\textwidth]{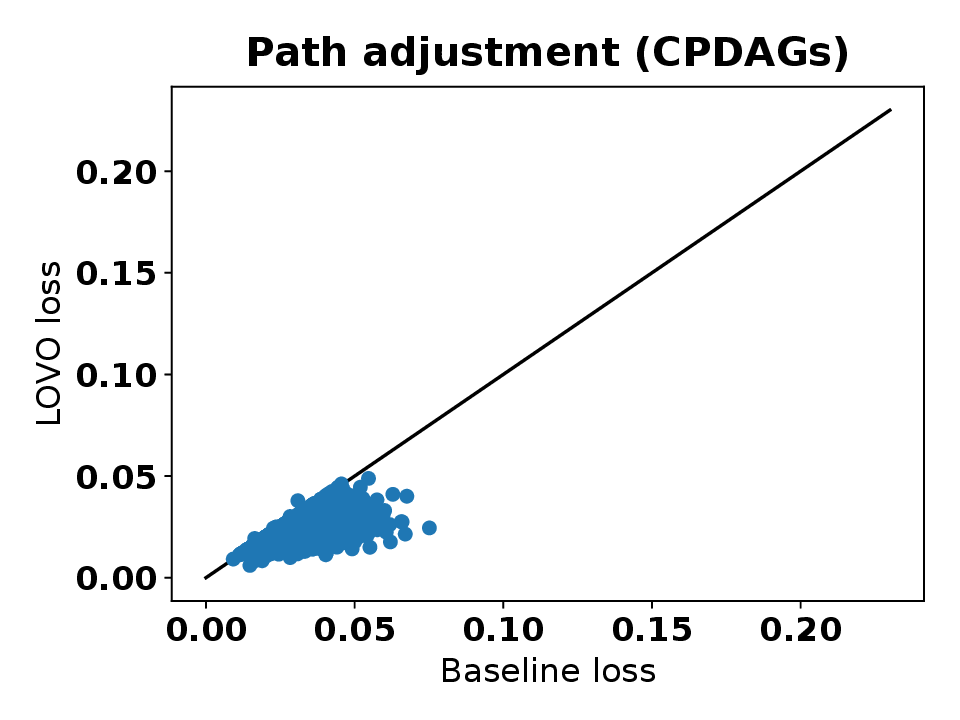}
\includegraphics[width=0.32\textwidth]{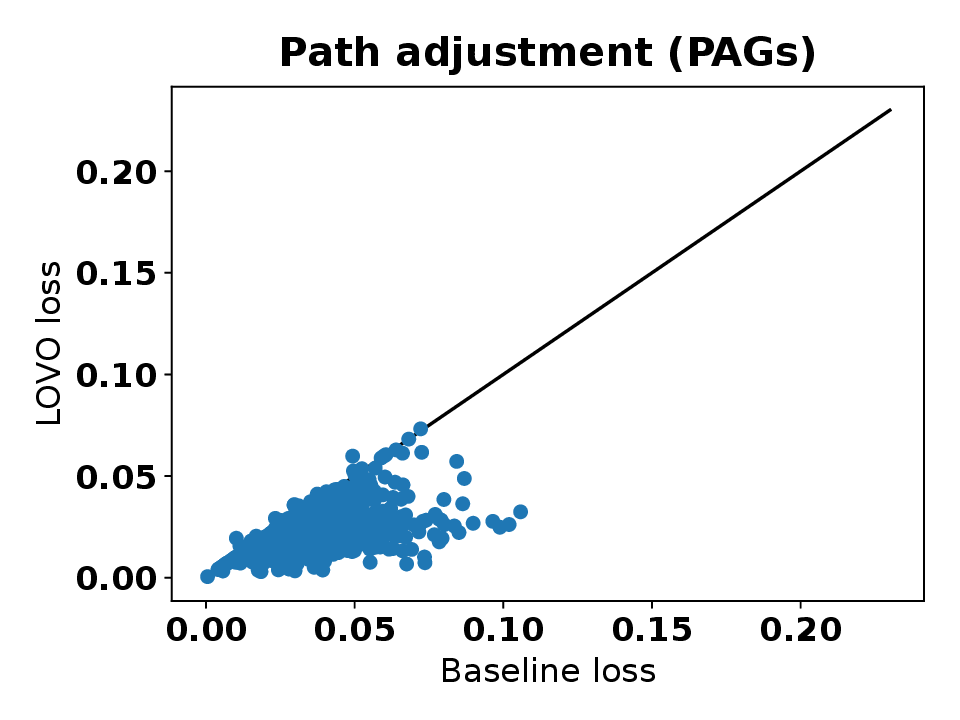}
\caption{\color{new}In all three cases, the LOVO loss rarely surpasses the baseline loss. For DAGs, we observe that the LOVO error is considerably smaller in most cases, while for CPDAGs and PAGs, the errors are often identical. This can be attributed to the same phenomenon encountered when applying LOVO to RCD in \ref{subsec:experiment_RCD_DirectLiNGAM}: we frequently need to abstain, resulting in only one or a few pairs contributing to the cross-validation average. If all these pairs have the empty set as their adjustment set, the baseline loss aligns with the LOVO loss.}\label{fig:LOVO_further_graph_types}
\end{figure}}
\end{document}